\documentclass[manuscript,screen]{acmart}
\usepackage{booktabs}
\usepackage{multirow}
\usepackage{forest}
\usepackage{subfigure}
\usepackage{comment}
\usepackage{color}
\usetikzlibrary{arrows.meta,shapes,positioning,shadows,trees}

\newtheorem{definition}{Definition}[section]

\AtBeginDocument{%
  }

\setcopyright{acmlicensed}
\copyrightyear{2024}
\acmYear{2024}
\acmDOI{XXXXXXX.XXXXXXX}

\acmISBN{978-1-4503-XXXX-X/18/06}

\begin{document}

\title{Advances in Set Function Learning: A Survey of Techniques and Applications}

\author{Jiahao Xie}
\email{jiahaox@udel.edu}
\orcid{https://orcid.org/0009-0002-8716-6566}

\affiliation{%
  \institution{University of Delaware}
  \city{Newark}
  \state{Delaware}
  \country{USA}
}

\author{Guangmo Tong}
\email{amotong@udel.edu}
\orcid{https://orcid.org/0000-0003-3247-4019}

\affiliation{%
  \institution{University of Delaware}
  \city{Newark}
  \state{Delaware}
  \country{USA}
}

\renewcommand{\shortauthors}{Xie et al.}

\begin{abstract}
  Set function learning has emerged as a crucial area in machine learning, addressing the challenge of modeling functions that take sets as inputs. Unlike traditional machine learning that involves fixed-size input vectors where the order of features matters, set function learning demands methods that are invariant to permutations of the input set, presenting a unique and complex problem. This survey provides a comprehensive overview of the current development in set function learning, covering foundational theories, key methodologies, and diverse applications. We categorize and discuss existing approaches, focusing on deep learning approaches, such as DeepSets and Set Transformer based methods, as well as other notable alternative methods beyond deep learning, offering a complete view of current models. We also introduce various applications and relevant datasets, such as point cloud processing and multi-label classification, highlighting the significant progress achieved by set function learning methods in these domains. Finally, we conclude by summarizing the current state of set function learning approaches and identifying promising future research directions, aiming to guide and inspire further advancements in this promising field.
\end{abstract}



\received{6 September 2024}
\received[revised]{2 January 2025}
\received[accepted]{18 January 2025}

\maketitle

\section{Introduction}
Set function learning is an emerging and rapidly developing field within machine learning \cite{wagstaff2019limitations}, focusing on learning functions defined on set-structured data. In contrast to conventional learning paradigms where the order of input data significantly affects the learning process, set function learning methods are characterized by their invariance to permutations of input elements \cite{lee2019set,zaheer2017deep}. This fundamental property makes them particularly effective for tasks involving unordered data. 
Conventional models, such as convolutional neural networks (CNNs) \cite{krizhevsky2012imagenet} and recurrent neural networks (RNNs) \cite{pascanu2013difficulty}, have achieved significant success in tasks such as time series analysis and natural language processing \cite{dennis2019shallow,sbrana2020n,sharma2020adaptation,thomas2022integrating}, where preserving the order of input data is essential for capturing the underlying structure. 
However, many real-world applications involve learning from inherently unordered sets  \cite{qi2017pointnet,aittala2018burst}, where conventional methods struggle because they rely heavily on input order. For example, in point clouds analysis for 3D object recognition and reconstruction, the individual points representing an object’s surface are inherently unordered \cite{qi2017pointnet++,xu2018spidercnn,zhou2018voxelnet}. Traditional methods often require extensive preprocessing, resulting in inefficiencies or data structure distortion. Similarly, in multi-label classification, where a single instance is associated with multiple labels \cite{yazici2020orderless,yeh2017learning,ridnik2021asymmetric}, treating these labels as a set is more appropriate than using traditional approaches like binary relevance. Binary relevance treats each label independently, failing to capture complex interdependencies between labels, while  
set function learning models can effectively capture the underlying structure of unordered data, leading to more accurate and robust predictions.

There is increasing literature proposing novel methods that are capable of handling set-structured data, opening new avenues for machine learning and set-based learning problems \cite{murphy2018janossy,wagstaff2019limitations}. For instance, DeepSets \cite{zaheer2017deep} introduces a framework for learning permutation-invariant functions, ensuring that the outputs remain unchanged regardless of the order of input elements. PointNet \cite{qi2017pointnet} revolutionizes point clouds processing by directly dealing with raw point sets without requiring voxelization or other preprocessing steps, simplifying the workflow and preserving data fidelity. Set Transformer \cite{lee2019set} leverages attention mechanisms to capture complex dependencies among set elements, enhancing the model’s ability to understand intricate relationships within the data. These pioneering works have shown the significant potential of neural networks to effectively model and learn from set-structured data, leading to substantial advancements in various domains such as point cloud processing \cite{rezatofighi2021learn,maron2020learning} and recommendation systems \cite{li2020reciptor,gim2021recipebowl}.

Despite the promising advancements, set function learning faces several unique challenges. A fundamental challenge is ensuring permutation-invariance \cite{wagstaff2022universal}, as the output of a set function learning model should remain unchanged regardless of the order of set elements. Another critical challenge is scalability \cite{ou2022learning}, as set function learning models should be capable of handling inputs ranging from small to large sets, often with varying sizes across different instances. This variability demands models that are flexible enough to adapt to sets with arbitrary cardinality while maintaining consistent performance. Additionally, the combinatorial nature of sets leads to significant computational challenges. As the size of the ground set increases, the number of possible subsets grows exponentially, making normal approaches infeasible for large-scale problems. The highly non-linear and interdependent relationships between set elements further complicate the learning process \cite{lee2019set}, requiring models that are expressive enough to capture these complex dependencies without becoming computationally intractable. It remains a central challenge to balance these requirements and effectively learn from limited data. Indeed, addressing these interconnected challenges demands specialized model architectures and innovative learning algorithms.

Given the rapidly growing interest in set function learning, 
we present a comprehensive survey of this promising area, providing researchers with insights into the state-of-the-art advancements. We review breakthrough papers and recent advancements, covering both theoretical foundations and practical implementations. 
While Kimura et al. \cite{kimura2024permutation} conduct literature review for permutation-invariant neural networks, their work only focuses on some typical methods and lacks the discussion of various applications. In contrast, as one of the very first surveys on set function learning, our work serves as a reference for anyone seeking to understand, apply, or advance this field, making several significant contributions. 
It provides a unified framework for understanding and categorizing diverse approaches to set function learning. The introduction of foundational theories allows interested readers to quickly capture basic concepts and engage in this area. Additionally, the systematical view of the strengths and limitations of different methodologies helps readers select the most appropriate approaches for specific tasks.
The extensive discussion of applications across multiple domains underscores the broad impact and potential of set function learning methods, encouraging their adoption in new areas. Furthermore, the introduction of various datasets serves as a valuable resource for set function learning research. Finally, by identifying the challenges and future directions, we offer valuable insights for the research community, potentially inspiring new research ideas and accelerating progress in this significant area. 

The rest of this survey is organized as follows: In Section \ref{section of Preliminaries}, we formally introduce the problem of set function learning and related basic concepts. Section \ref{section of Deep learning methods} discusses various deep learning methods for solving set function learning problems, while other approaches are mentioned in Section \ref{section of Other methods}. The reviewed set function learning methods are summarized in Table \ref{tab:countrydata}. Section \ref{section of Applications} describes various applications of set function learning models across different domains and introduces relevant datasets.
Finally, we make a conclusion and discuss future directions in Section \ref{section of Conclusion and future directions}.

\begin{table}[t]\footnotesize
    \centering
    \caption{A summary of reviewed work}
    \begin{tabular}{llll}
        \toprule [2pt]
        \multicolumn{2}{c}{\textbf{Category}} & \textbf{Strategy} & \textbf{Reference} \\
        \midrule[0.5pt]\midrule[0.5pt]
        \multirow{25}{*}{\shortstack{\textbf{Deep} \\ \textbf{learning} \\ \textbf{methods}}} 
        &\multirow{2}{*}{CNN based methods} &Extend convolution to sets & \cite{wendler2019powerset,xu2018spidercnn}   \\
        
        &&Integrate convolution with symmetry aggregation & \cite{aittala2018burst,zhong2018compact} \\
        \cmidrule[1pt]{2-4}
        &\multirow{1}{*}{RNN based methods} &Optimize set probability & \cite{li2023set,qin2019adapting}   \\
        \cmidrule[1pt]{2-4}
         &\multirow{2}{*}{FNN based methods} &Define discrete distributions for set variables & \cite{rezatofighi2021learn}   \\
        &&New permutation-invariant and equivariant-functions  & \cite{yu2023predicting} \\
        \cmidrule[1pt]{2-4}
         &\multirow{4}{*}{DeepSets based methods} &Generalization of DeepSets & \cite{maron2020learning,murphy2018janossy}   \\
         &&Expressive power analysis of DeepSets & \cite{wagstaff2019limitations,zweig2022exponential,wang2023polynomial,zaheer2017deep}  \\
        &&Improve DeepSets through new aggregating approaches & \cite{abedin2019sparsesense,bartunov2022equilibrium,horn2020set}  \\
        
        &&Extend DeepSets to specific scenarios & \cite{ou2022learning,wang2023deep,yi2021cytoset}  \\
        \cmidrule[1pt]{2-4}
         &\multirow{2}{*}{PointNet based methods} 
         &Expressive power analysis of PointNet & \cite{bueno2021representation}  \\
         &&Improve capabilities of PointNet  & \cite{prokudin2019efficient,qi2017pointnet,qi2017pointnet++}  \\
        \cmidrule[1pt]{2-4}
         &\multirow{4}{*}{Set Transformer based methods}
         &Improve Set Transformer under specific condition & \cite{lee2019set,zhang2022set}  \\
         &&Employ Set Transformer as encoder & \cite{gim2021recipebowl,jurewicz2022set,li2020reciptor,zhang2022relational}   \\
        && Extend Set Transformer to meta-learning & \cite{lee2022set}  \\
        &&Other methods based on attention mechanisms & \cite{girgis2021latent,zhao2021point}  \\
        \cmidrule[1pt]{2-4}
        &\multirow{2}{*}{DSPN based methods}
        &Replace pooling in DSPN with FSPool & \cite{zhang2019fspool}  \\
        &&Improve DSPN under specific conditions & \cite{kosiorek2020conditional,zhang2020set,zhang2019deep,zhang2021multiset}  \\
        
         \cmidrule[1pt]{2-4}
         &\multirow{2}{*}{DSF based methods} &Improve DSF under specific conditions & \cite{dolhansky2016deep,ghadimi2020deep,manupriya2022improving}\\
         &&Address the limitation of DSF & \cite{de2022neural}\\
        \cmidrule[1pt]{2-4}
        &\multirow{7}{*}{Other deep learning methods} & RepSet for handling sets of vectors  & \cite{skianis2020rep}   \\
        
        &&Exchangeable Neural ODE (ExNODE)  & \cite{li2020exchangeable}  \\
        && GaitSet for recognizing individual gaits & \cite{chao2019gaitset}  \\
        &&Deep Message Passing on Sets fot relational learning & \cite{shi2020deep} \\
        && Prototype-oriented optimal transport (POT)  & \cite{dan2021learning} \\
        &&Permutation-Optimisation (PO) module & \cite{zhang2018learning}  \\
        &&  Mini-Batch Consistency (MBC) and UMBC framework  & \cite{bruno2021mini,willette2023scalable} \\
        \midrule[0.5pt]\midrule[0.5pt]
        \multirow{5}{*}{\shortstack{\textbf{Other} \\ \textbf{methods}}} 
        & &Kernel methods & \cite{buathong2020kernels,nikolentzos2017matching}  \\
        &&Sparse Set Function Fourier Transform (SSFT) & \cite{wendler2021learning}  \\
        &&Set Locality Sensitive Hashing (SLoSH) & \cite{lu2024slosh}  \\
        &&Decision tree for learning submodular functions & \cite{feldman2013representation}  \\
        &&PAC learning for submodular function & \cite{raskhodnikova2013learning}  \\

        \midrule[0.5pt]\midrule[0.5pt]
        \multirow{11}{*}{\textbf{Applications}} 
        &Point cloud processing &Point Transformer, Basis Point Sets, Dumlp-pin, etc & \cite{fei2022dumlp,prokudin2019efficient,qi2017pointnet,qi2017pointnet++,wang2019dynamic,zhao2021point}  \\
        &Set anomaly detection & DeepSets, Set Transformer, EquiVSet, etc& \cite{jung2015exploration,zaheer2017deep}  \\
        &Recommendation systems &Recipebowl, Reciptor& \cite{gim2021recipebowl,li2020reciptor}  \\
        &Set expansion and set retrieval &Equilibrium Aggregation, Set Transformer, PointNet, etc & \cite{zaheer2017deep,bartunov2022equilibrium,lee2019set,qi2017pointnet}  \\
        &Time series prediction &SparseSense, SeFT, Deep Temporal Sets, SFCNTSP, etc& \cite{horn2020set,yu2023predicting,abedin2019sparsesense,wang2023deep}  \\
        &Multi-label classification &FLEXSUBNET, DSF, SEA-NN, set-RNN, etc & \cite{qin2019adapting,rezatofighi2021learn,skianis2020rep,zhang2019fspool,ghadimi2020deep,manupriya2022improving,de2022neural}  \\
        &Molecular property classification &Equilibrium Aggregation, Meta-Interpolation, EquiVSet, etc & \cite{lee2022set,bartunov2022equilibrium,zhang2022composition,ou2022learning}  \\
        &Amortized Inference &Set Transformer, Deep Amortized Clustering, NCP, etc  & \cite{lee2019set,lee2019deep,pakman2020neural,wang2020amortized,jha2022neural,mullertransformers}  \\
        &Human activity recognition &Gaitset & \cite{chao2019gaitset}  \\
        &Predict clinical outcomes & CytoSet & \cite{yi2021cytoset}  \\
        &Cancer detection & Universal MBC & \cite{willette2023scalable}  \\
        \bottomrule[2pt]
    \end{tabular}
    \label{tab:countrydata}
\end{table}

\section{Preliminaries}
\label{section of Preliminaries}

Set function is a type of function defined on set and particularly relevant in machine learning that deals with set structural data, such as point clouds \cite{qi2017pointnet,qi2017pointnet++,xu2018spidercnn}, molecular structures \cite{duvenaud2015convolutional,faber2016machine}, and any other unordered collection of elements \cite{gilmer2017neural,rezatofighi2017deepsetnet,zhang2020set}. We begin this section by introducing the definition of set function.
\begin{definition}[Set function]
\label{definition of set function}
    For two sets $X,Y$, a set function is defined as a mapping from $2^{X}$ to $Y$, i.e., $f:2^{X}\to Y$, where $Y$ is the response range and can be any set, such as the set of scalars, vectors, sets, and more complex structures.
\end{definition}
There are many common set functions and we provide some examples as follows:

\textbf{Example 2.1} (Point set function \cite{qi2017pointnet,zhao2021point}). In point cloud classification, a point cloud is a set of points in 3D space, where each point is represented by its coordinates $(x,y,z) \in \mathbb{R}^3$. The objective is to learn a point set function that can predict the label associated with the input point cloud. Formally, this point set function can be formulated as $h\big(\{(x_i,y_i,z_i)\}_{i=1}^n\big)=l$, where $\{(x_i,y_i,z_i)\}_{i=1}^n$ denotes the set of coordinates representing the point cloud and $l$ is the corresponding label. 

\textbf{Example 2.2} (Product cost summarizing function \cite{lee2019set}). In a recommendation system, the objective is to learn a product cost summarizing function to recommend cost-effective products to users. Formally, this product cost summarizing function can be formulated as
$c\big(\{\boldsymbol{f}_i\}_{i=1}^n\big)=\sum_{i=1}^n\big(f_{i}^1\cdot f_{i}^2\big)$, where $\{\boldsymbol{f}_i\}_{i=1}^n \subseteq\mathbb{R}^d$ is the set of products, $\boldsymbol{f}_i=\big(f_{i}^1,f_{i}^2,\dots,f_{i}^d\big)$ is a $d$-dimensional feature vector of product $i$, $f_{i}^1$ and $f_{i}^2$ represent price and quantity respectively.

\textbf{Example 2.3} (Square corner prediction function \cite{zhang2019fspool}). For object detection in traffic scenes, the objective is to learn a square corner prediction function to predict bounding boxes around objects such as cars. Formally, this square corner prediction function can be formulated as $g\big(\{(x_i,y_i)\}_{i=1}^4\big)=\{(x_i^\prime,y_i^\prime)\}_{i=1}^4$, where $\{(x_i,y_i)\}_{i=1}^4\subseteq\mathbb{R}^2$ represents the vertices of a square and $\{(x_i^\prime,y_i^\prime)\}_{i=1}^4\subseteq\mathbb{R}^2$ represents four corners of the rotated square.

Having shown some examples of set functions, we introduce supervised set function learning, a branch of set function learning, aiming to learn set functions from labeled training data. The supervised set function learning problem is defined as follows. 

\begin{definition}[Supervised set function learning]
\label{definition of Supervised set function learning}
    For two sets $X$ and $Y$, suppose that $\mathcal{D}$ is an unknown underlying probability distribution over $2^X\times Y$, from which the training set $D$ is assumed to be sampled, i.e. $D = \{(x_i,y_i)\}_{i=1}^n$, where each $x_i\in 2^X$ is an input set and $y_i\in Y$ is the corresponding target output label. We begin by choosing a hypothesis space $\mathcal{H}\subseteq\big\{h:2^X \rightarrow Y\big\}$. The goal is to find a function $h\in \mathcal{H}$ that maps input sets $x_i$ to outputs $y_i$, such that the loss function $L_\mathcal{D}(h)\overset{\underset{\mathrm{def}}{}}{=}\mathop{\mathbb{P}}\limits_{(x,y)\sim\mathcal{D}}[\mathcal{\ell}(h(x),y)]$ can be minimized, where $\mathcal{\ell}:\mathcal{H}\times Z\rightarrow\mathbb{R}^+$ measures the difference between the prediction $h(x)$ and the ground truth $y$, and $Z=2^X\times Y$.
\end{definition}

\begin{figure}[t]  
    \centering      
    \subfigure[ Sphere point cloud ] 
    {
        \begin{minipage}[t]{0.2\linewidth}
            \centering           \includegraphics[width=0.9\columnwidth]{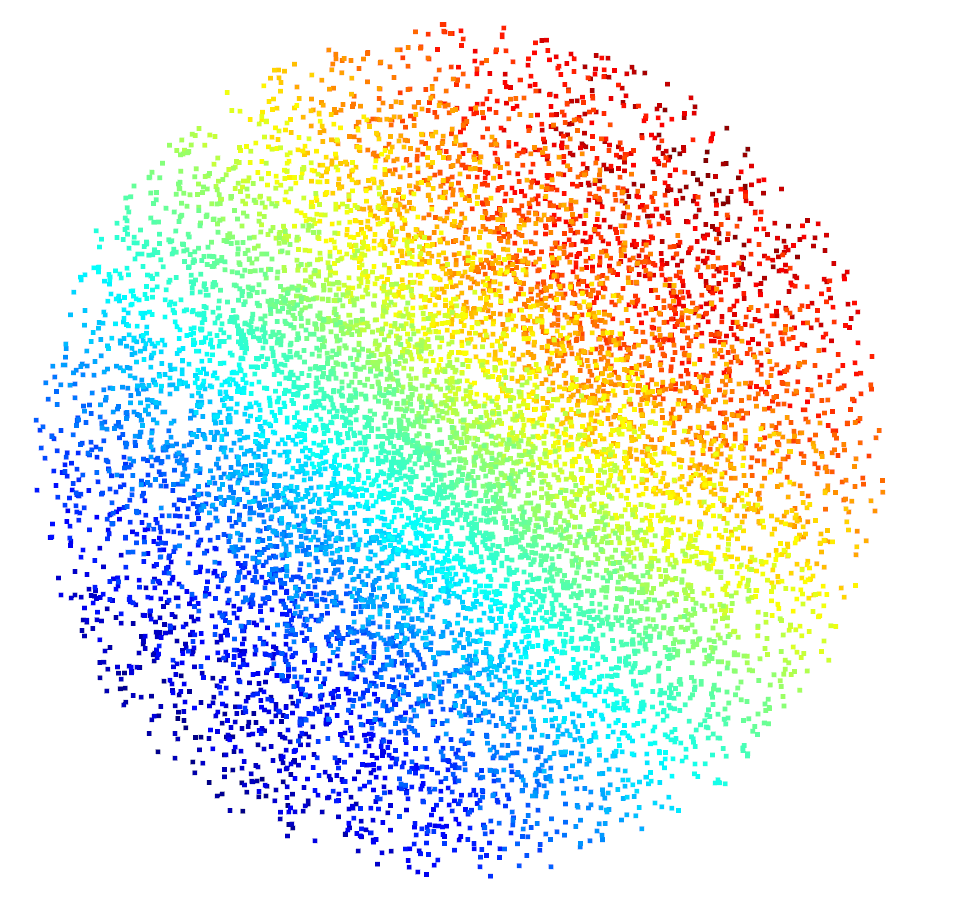}   
        \end{minipage}%
    }
    \subfigure[Cube point cloud] 
    {
        \begin{minipage}[t]{0.2\linewidth}
            \centering      \includegraphics[width=0.9\columnwidth]{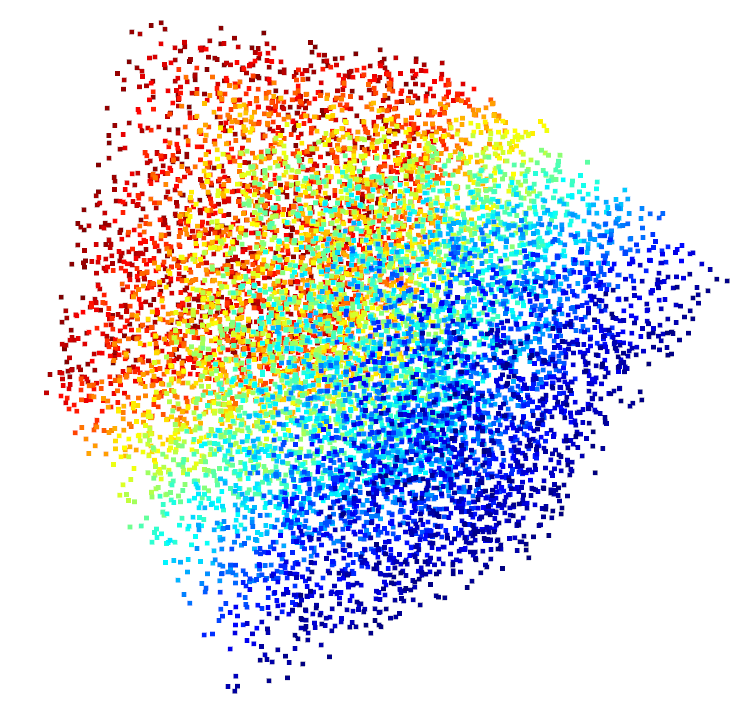}   
        \end{minipage}
    }%
    \subfigure[Products price prediction] 
    {
        \begin{minipage}[t]{0.2\textwidth}
            \centering      \includegraphics[width=0.9\columnwidth]{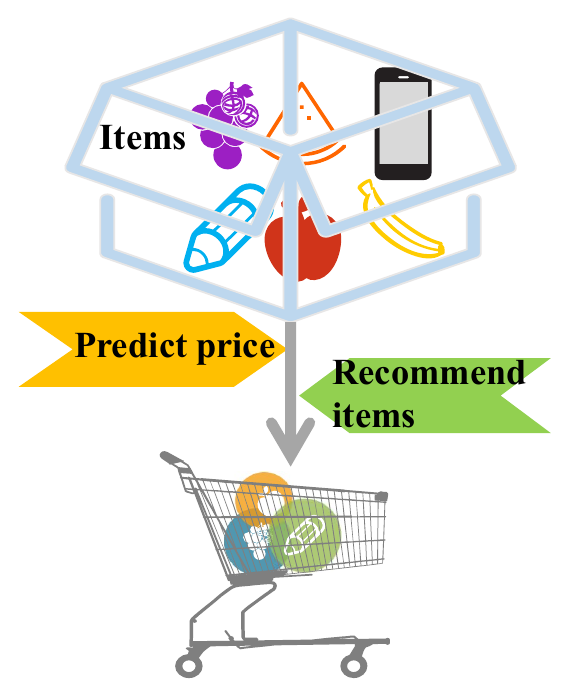}   
        \end{minipage}
    }%
    \subfigure[Conors prediction] 
    {
        \begin{minipage}[t]{0.2\textwidth}
            \centering      \includegraphics[width=0.9\columnwidth]{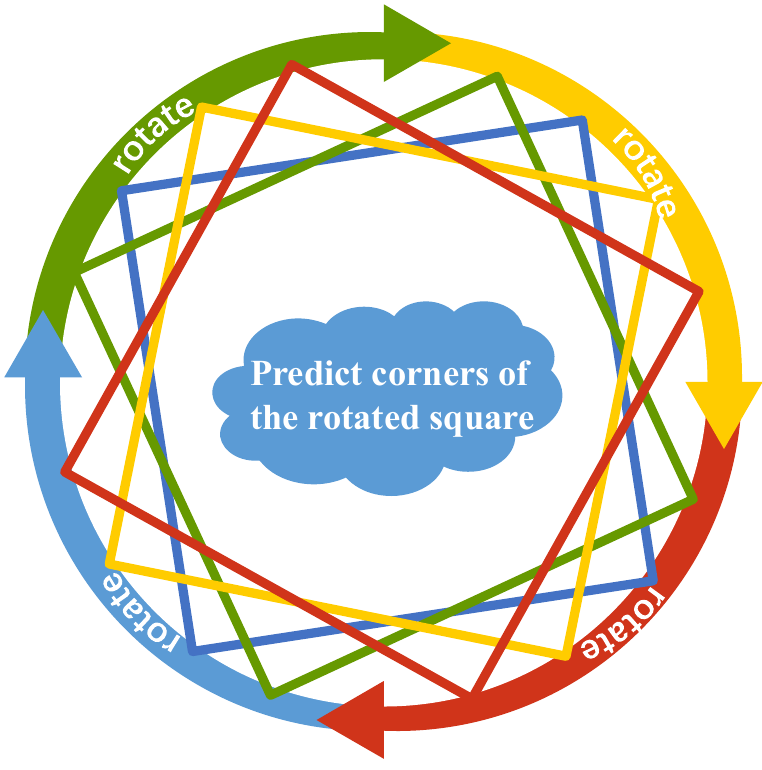}   
        \end{minipage}
    }%
    \caption{Visualization of examples} 
    \label{figure of Visualization of examples} 
    \Description{figure of point cloud}
\end{figure}

Supervised set function learning has applications in various domains such as computer vision (e.g., object detection \cite{zhang2019deep} and scene understanding \cite{qi2017pointnet}), natural language processing (e.g., text summarization \cite{bai2023permutation} and relation extraction \cite{liu-lapata-2019-hierarchical}) and bioinformatics (e.g., protein prediction \cite{jiang2023pharmacophoric} and drug discovery \cite{jumper2021highly}). There are some example tasks of supervised set function learning as follows. The visualizations of these examples are illustrated in Figure \ref{figure of Visualization of examples}.

\textbf{Example 2.4} (Point cloud classification \cite{qi2017pointnet,qi2017pointnet++}). 
Let $X$ be the set of all 3D points in the space and $Y$ be the set of object categories such as sphere and cube. Each input set $x_i \in 2^X$ represents a point cloud and the corresponding label $y_i \in Y$ represents the object category of the point cloud. The goal is to find a function $h \in \mathcal{H}$ that classifies each point cloud $x_i$ into its correct category $y_i$. 

\textbf{Example 2.5} (Predicting total cost of a set of products \cite{wagstaff2019limitations}). Let $X$ be the set of all possible products and $Y$ be the set of possible total costs. Each input set $x_i \in 2^X$ represents a set of products, where each product is characterized by features such as price, weight, and quantity. The corresponding label $y_i \in Y$ represents the total cost of the set of products. The goal is to find a function $h \in \mathcal{H}$ that accurately predicts the total cost $y_i$ for each set of products $x_i$.

\textbf{Example 2.6} (Predicting corners of the rotated square \cite{zhang2019deep,zhang2019fspool}). Let $X$ be the set of all 2D points in a plane and $Y$ be the set of possible sets of four points. Each input set $x_i \in 2^X$ represents the vertices of a square rotated by an angle $\theta$ around the origin. The corresponding label $y_i \in Y$ represents the four corners of the square after rotation. The goal is to find a function $h \in \mathcal{H}$ that predicts the four corners $y_i$ for each set of vertices $x_i$ given the rotation angle $\theta$.

With a growing literature focusing on designing novel set function learning methods, we summarize three issues that should be taken into account when designing new set function learning methods, including permutation-invariance, theoretical expressive power, and scalability.

1) \textbf{Permutation-invariance:} Permutation invariance is a fundamental requirement \cite{zhang2020set} for set function learning. In tasks such as point cloud processing \cite{qi2017pointnet} and molecular property prediction  \cite{jiang2023pharmacophoric}, the output of models should remain consistent regardless of the order of the input set elements \cite{jumper2021highly,qi2017pointnet++,rezatofighi2017deepsetnet}. This property, known as permutation invariance, is essential for any learning method that handles set-structured data. To show the significance of permutation-invariance, consider the scenario of using conventional CNNs for point cloud classification. In this case, the input point set should be transformed into an ordered vector before being fed into the network. However, if the points are permuted differently, the extracted features after convolution and pooling will reflect the new order of points, potentially resulting in different classification outcomes. This variability is undesirable since the output label should be invariant for classifying the same set of points regardless of their order. This example underscores the necessity of a permutation-invariant hypothesis space in set function learning methods.
To define permutation-invariance for functions on matrices, we first introduce the permutation matrix. The permutation matrix is a square binary matrix with exactly one entry of 1 in each row and each column, and 0 elsewhere, representing a permutation of set elements. For example, the permutation matrix $[(0,1,0),(0,0,1),(1,0,0)]$ represents the permutation where the first element goes to the second position, the second element to the third position, and the third element to the first position.
Given an input $\mathcal{X} \in \mathbb{R}^{N \times m} $ consisting of $N$ $m$-dimensional vectors, a permutation matrix $\Pi$ belongs to the set of all permutation matrices $\Pi_N$. Using $n$ to represent the dimension of the output vectors, we can formally define permutation-invariance as:
\begin{definition}[Permutation-invariance]
\label{definition of Permutation-invariance}
    For each $\mathcal{X}\in \mathbb{R}^{N \times m}$ and $\Pi \in \Pi_N$, if $f(\Pi \mathcal{X})=f(\mathcal{X})$ always holds, the function $f:\mathbb{R}^{N \times m} \rightarrow \mathbb{R}^{N \times n}$ is permutation-invariant.
\end{definition}
To keep permutation-invariance, various techniques are employed and we summarize three key strategies as follows. 
\begin{itemize}
\item \textbf{Sorting} is a straightforward technique employed in various learning models \cite{wagstaff2022universal}, where input set elements are sorted into a canonical ordering before being fed into the model. This mechanism essentially restricts the hypothesis space to inherently permutation-invariant functions, i.e. $f(\mathrm{sort}(X))$, where $X$ is the input set. However, this restriction may exclude some complex relationships that depend on the original data ordering or structure, biasing the hypothesis space towards functions that work well with particular sorting and limiting the generalization ability of models.

\item \textbf{Augmenting the training data} with various permutations of the input sets is a commonly used technique \cite{vinyals2015order}. The basic idea is to create multiple reordered versions of each input set and include all these versions in the training data. While this strategy keeps the original hypothesis space unchanged, it encourages the model to find approximately permutation-invariant functions within this space. This approach is more flexible than sorting and can be easily combined with existing learning methods such as RNNs \cite{bahdanau2015neural}. 
However, augmenting significantly increases the data size and it is computationally infeasible to generate all possible permutations. 

\item \textbf{Aggregating features of set elements} through symmetric functions is an effective technique \cite{bartunov2022equilibrium}. The key idea is to employ a permutation-invariant function to aggregate feature vectors from all elements in the input set into a unified set-level representation. This approach explicitly builds permutation-invariance into the model architecture and restricts hypothesis space to permutation-invariant functions of the form $f(g(\phi(x_1), \phi(x_2),..., \phi(x_n)))$, where $\{x_i\}_{i=1}^n$ is the input set, $\phi$ is the encoder, $g$ is a symmetric aggregation function such as sum (DeepSets \cite{zaheer2017deep}) and max (PoinNet \cite{qi2017pointnet}), and $f$ is a task-specific function. In fact, more complex encoders and aggregators, such as attention mechanisms \cite{wang2023deep}, can expand the hypothesis space, potentially capturing higher-order element relationships.
\end{itemize}

2) \textbf{Theoretical expressive power:} Expressive power in set function learning refers to the capacity of models to represent and approximate set functions. Set function learning methods should have sufficient expressive power with theoratical guarantee to capture complex relationships between set elements and set-level features \cite{wagstaff2019limitations,wagstaff2022universal}. For example, point cloud processing often requires capturing high-level geometric structures and patterns \cite{qi2017pointnet}. Insufficient expressive power can lead to underfitting and poor performance in complex tasks.

3) \textbf{Scalability:} It is vital for set function learning methods to handle input sets of varying sizes and run in polynomial time \cite{lee2019set}, ensuring their practical applicability \cite{ou2022learning}. For example, in point cloud processing \cite{qi2017pointnet++}, the number of points representing an object can vary with resolution and sampling method, sometimes reaching millions. Scalable methods can also adapt to dynamic applications with growing data over time \cite{rafiey2020fast,wang2019dynamic}, allowing for incremental learning and updating the model with newly available data instead of retraining from scratch. 
 
\section{Deep learning methods}
\label{section of Deep learning methods}

Deep learning methods have become pivotal in addressing various learning problems \cite{lecun2015deep}. The fundamental neural networks, such as CNNs \cite{krizhevsky2012imagenet} and RNNs \cite{van2016pixel}, have achieved remarkable success in multiple tasks, such as image segmentation \cite{ronneberger2015u,chen2017deeplab}, object detection \cite{ren2016faster,redmon2016you} and speech synthesis \cite{oh2019speech2face,tan2024naturalspeech}. However, these models implicitly incorporate regularity assumptions in their neural structures, making them less adaptable to irregular data domains such as sets, which lack a fixed permutation \cite{zaheer2017deep}. To handle set function learning tasks, several works extend CNNs and RNNs to set function learning, while there is also growing literature developing novel deep learning methods specialized for processing set structural data. In this section, we introduce various deep learning methods designed for set function learning problems, categorizing them into multiple groups: CNN based methods (Section \ref{section of CNN based methods}), RNN based methods (Section \ref{section of RNN based methods}), DeepSets based methods (Section \ref{section of DeepSets based methods}), PointNet based methods (Section \ref{section of PointNet based methods}), Set Transformer based methods (Section \ref{section of Set Transformer based methods}), Deep Set Prediction Network based methods (Section \ref{section of Deep set prediction networks based methods}), Deep Submodular Function based methods (Section \ref{section of Deep Submodular Function based methods}) and other deep learning methods (Section \ref{section of Other deep learning methods}) that cannot be classified into above groups.

\begin{figure}[t]
    \centering
    \includegraphics[width=0.9\linewidth]{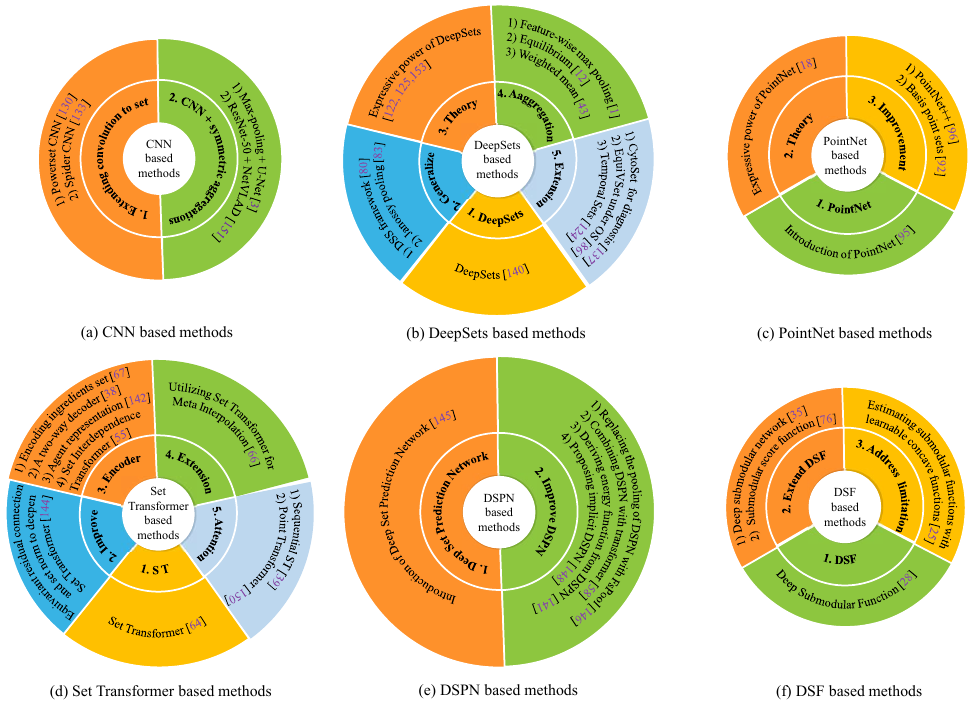}
    \caption{Structure figures. \ref{structure figures}\textcolor[RGB]{109,35,130}{(a)} shows the structure of Section \ref{section of CNN based methods}. \ref{structure figures}\textcolor[RGB]{109,35,130}{(b)} shows the structure of Section \ref{section of DeepSets based methods}. \ref{structure figures}\textcolor[RGB]{109,35,130}{(c)} shows the structure of Section \ref{section of PointNet based methods}. \ref{structure figures}\textcolor[RGB]{109,35,130}{(d)} shows the structure of Section \ref{section of Set Transformer based methods}. \ref{structure figures}\textcolor[RGB]{109,35,130}{(e)} shows the structure of Section \ref{section of Deep set prediction networks based methods}. \ref{structure figures}\textcolor[RGB]{109,35,130}{(f)} shows the structure of Section \ref{section of Deep Submodular Function based methods}.}
    \label{structure figures}
\end{figure}

\subsection{CNN Based Methods} 
\label{section of CNN based methods}
CNNs are highly efficient architectures due to their ability to leverage local connectivity and shared weights \cite{hua2018pointwise}, leading to breakthroughs in a wide variety of tasks such as image processing \cite{tan2019efficientnet}. To make full use of such advantages, some research extends CNNs to set-based learning problems. In this section, we introduce several CNN-based set function learning methods and divide them into two categories. One is extending convolution to set structural data and the other is integrating CNN with symmetric aggregations. The structure of this section is illustrated in Figure \ref{structure figures}\textcolor[RGB]{109,35,130}{(a)}.

\subsubsection{Extending Convolution Operation to Set} \
\newline The first strategy for CNNs to learn set function is extending convolution operation to set. 
Wendler et al. \cite{wendler2019powerset} propose a novel class of CNNs for set functions by proposing powerset convolution and pooling. Powerset convolution is designed to be shift-equivariant, meaning it commutes with specified shifts on the powerset domain. The shift is defined as modifying a set function $s(A)$ by removing a subset $Q$ from argument $A$, denoted as $T_Qs=(s_{A\setminus Q})_{A\subseteq N}$, where $N$ is the ground set. The convolution is given by \cite{puschel2018discrete} as $(h\ast s)_A=\sum_{Q\subseteq N}h_Qs_{A\setminus Q}$, where the filter $h$ is a set function. 
The powerset convolution layer is constructed by conducting powerset convolutions on multiple channels, summarizing the feature maps as in \cite{bronstein2017geometric}, with both input and output being sets of set functions. Each output set function is derived by convolving the input set functions with corresponding filters, summing the results, and applying the non-linear transformation. Powerset pooling layer can reduce the complexity by aggregating elements into a smaller ground set, mapping the original set function to a new one on this reduced set, and can be implemented in various ways, such as combining elements as in \cite{scheibler2015fast} and using a simple max rule. Powerset CNNs consist of multiple powerset convolution and pooling layers, the number of which can be adjusted according to specific tasks. However, the complexity analysis \cite{lu2016practical} shows that powerset CNNs are impractical for large ground sets as the number of set functions grows exponentially with the size of the ground set. 
Xu et al. \cite{xu2018spidercnn} develop SpiderCNN, a novel CNN designed specifically for processing point clouds. The core component of SpiderCNN is the SpiderConv layer, which replaces conventional convolutional layers to enable convolution on point sets. Given a function $F$ defined on a point set $P\subseteq \mathbb{R}^n$ and a filter $g:\mathbb{R}^n\rightarrow \mathbb{R}$ within a sphere centered at the origin with radius $r\in\mathbb{R}$, the SpiderConv can be formulated as 
\begin{equation}
 F\ast g(p)=\sum_{q\in P, \|q-p\|\leq r}F(q)g(p-q),
 \nonumber
\end{equation}
where $p,q$ are points. Conventional convolution becomes a special case of SipderConv when $P=\mathbb{Z}^2$ is a regular grid. In SpiderConv, the filter $g$ belongs to a parameterized filter family $\{g_w\}$, which is piece-wise differentiable for $w$ and can be efficiently optimized using stochastic gradient descent (SGD). $\{g_w\}$ is defined as the product of a step function and a Taylor polynomial, enabling capturing local geodesic information and ensuring expressiveness. SpiderCNN inherits the advantages of CNNs, making it effective to extract deep features and achieve good performance in segmentation tasks.

\subsubsection{Combining CNN with Symmetric Aggregation}\
\newline The second strategy is combining CNN with symmetric aggregation like mean and max pooling, where the final aggregated feature depends only on the set contents and not their order. To deal with burst image deblurring, Aittla et al. \cite{aittala2018burst} propose a U-Net-inspired \cite{ronneberger2015u} framework with symmetric pooling operations. In this framework, each image of the input set is processed individually through identical neural networks with tied weights, producing feature vectors. These feature vectors are computed through symmetry operations such as mean and max pooling. Eventually, the pooled features are processed through further neural network layers, outputting an estimate of the sharp image. In addition, the intermediate pooling layer is introduced, followed by $1\times1$ convolutions that fuse global features into local ones. This layer allows the concatenation of the pooled global state back to the local features, enabling information exchange between the set entities. Zhong et al. \cite{zhong2018compact} design a CNN based architecture called SetNet, which aggregates face descriptors into a compact descriptor. This framework is developed to enhance the efficiency and accuracy of retrieving a set of images that match a given query containing the descriptions of images for multiple identities. SetNet utilizes ResNet-50 \cite{he2016deep} to extract features from each image, generating individual descriptors. These descriptors are aggregated into a fixed-length set-level vector using NetVLAD \cite{arandjelovic2016netvlad}, which also helps reduce memory usage and runtime.

\subsection{RNN Based Methods}
\label{section of RNN based methods}
RNNs are efficient in recognizing patterns when handling sequential structure data, such as time series \cite{sbrana2020n,dennis2019shallow}, speech \cite{gelly2017optimization,pandey2022self} and text \cite{thomas2022integrating,sharma2020adaptation}. The connections in RNNs form directed cycles, enabling them to maintain a hidden state that captures temporal dependencies in input sequences \cite{zhang2021sbo}. This characteristic makes RNNs particularly suitable for sequential tasks. In this section, we introduce some works that extend RNNs to set function learning.

Qin et al. \cite{qin2019adapting} present set-RNN, an adaptation of RNN for dealing with multi-label text classification, where the target output is a label set. Previous approaches tackling such tasks either transform the set into a predefined sequence or connect sequence probability with set probability, but these methods lack solid theoretical foundations and perform poorly in practice. The authors propose a novel training objective that maximizes set probability defined as the sum of probabilities across all sequence permutations of the set. During training, a variant of beam search is employed to approximate set probability by identifying the top $K$ highest probability sequences. The same approximation technique is used during prediction to find the label set with the highest probability. This novel objective enhances the flexibility of set-RNN to search the best label orders, enabling it to efficiently tackle multi-label classification tasks. Inspired by \cite{qin2019adapting}, Li et al. \cite{li2023set} develop a new approach called set learning, which optimizes the set probability by considering multiple permutations of structured objects. This method is applied to generative information extraction (IE) tasks, where the input is a text $X=[x_1,x_2,...]$ and the output is a set of structured objects $S=\{s_1,s_2,...\}$, with each structured object consisting of several spans from $X$. Set learning introduces a new method to calculate the set probability, formulated as:
\begin{equation}
    p(S|X=p(Y|X))=\sum_{\pi_z(Y)\in \Pi(Y)}P(\pi_z(Y)),
\label{equation of set probability}
\end{equation}
where $\Pi(Y)$ denotes all possible permutations of $Y$ and $\pi_z(Y)$ is a specific permutation in $\Pi(Y)$. The set $Y$ has the same size as $S$, containing all elements of $S$ flattened into sub-sequences. Based on Seq2Seq learning \cite{sutskever2014sequence}, the set learning optimizes the set probability through Equation(\ref{equation of set probability}) and reduces the calculation cost through permutation sampling, achieving good performance on IE tasks. However, with the size of training data increasing, the benefits of permutation sampling diminish, while the runtime significantly increases.

\subsection{FNN Based Methods}
\label{section of FNN based methods}
Feedforward neural network (FNN), also known as MLP, is a fundamental architecture where information flows in one direction, from the input layer through hidden layers to the output layer \cite{bebis1994feed}. It is widely used for various machine learning tasks such as classification \cite{huang2000classification}, 
and pattern recognition \cite{rajput2014back}. In this section, we discuss several methods handling set structured data based on FNNs. 

Rezatofighi et al. \cite{rezatofighi2021learn} introduce an innovative deep FNN based approach, Deep Perm-Set Network (DPSN), to address set prediction problems, where the outputs are sets with arbitrary permutation and cardinality. DPSN models the set distributions by defining discrete distributions for set cardinality and permutation variables, as well as a joint distribution over set elements given a fixed cardinality. In the scenario where permutation is fixed during training, the permutation of the output set elements is consistently ordered during training, suitable for tasks such as multi-label classification. The network predicts the cardinality and the state (i.e., existence scores) of each set element by optimizing a loss function that combines cardinality loss (e.g., the negative logarithm of categorical distribution) with state loss (e.g., binary cross-entropy). The fixed permutation simplifies the learning process by eliminating the need to handle varying orderings of set elements. In the scenario of learning the distribution over permutations, the model address tasks where elements order varies during training, such as object detection. DPSN approximates the marginalization over all possible permutations, sampling significant permutations and dynamically determining the best assignment (permutation) for each training instance using the Hungarian algorithm. The learning process optimizes the posterior distribution over the network parameters by jointly considering cardinality, permutation, and state losses. In the scenario where order does not matter, the permutation of output set elements is assumed to be uniformly distributed, applicable to tasks where the specific order of set elements is not important. The network optimizes the cardinality and state losses without considering element order, dynamically determining the assignment between network outputs and ground truth annotations during each SGD iteration. While DPSN is proved to be effective in experiments, its scalability is limited due to the exponential increase in possible permutations with the set size growing.

Yu et al. \cite{yu2023predicting} design a simple framework only with Simplified Fully Connected Networks (SFCNs) for temporal set prediction of user behaviors. This framework firstly adopts an element embedding layer to learn the representations of set elements. Subsequently, the set representation is computed through the newly designed permutation-invariant functions and an SFCN is applied to capture temporal dependencies among sets. To enable interactions between elements within each set, the newly designed permutation-equivariant function is employed to establish relationships between elements. Following this, another SFCN is used to uncover implicit correlations across multiple embedding channels. Finally, the user representations are aggregated adaptively by average-pooling and the probability of this behavior's occurrence in the next period-set is calculated using an adaptive fusing module. Notably, this work is the first to show that a simple architecture can effectively deal with temporal set prediction tasks.

\subsection{DeepSets Based Methods}
\label{section of DeepSets based methods}

Designing novel deep learning methods to tackle set function learning problems has been an active research topic since Zaheer et al. \cite{zaheer2017deep} propose the foundational framework known as DeepSets. This pioneering work establishes key design principles for deep permutation-invariant neural networks and outlines the essential components, such as permutation-equivariant feature extraction and permutation-invariant set pooling. In this section, we introduce the basic concepts of DeepSets and the subsequent advancements. This section is organized as in Figure \ref{structure figures}\textcolor[RGB]{109,35,130}{(b)}.

\subsubsection{DeepSets}
\
\newline In order to handle learning tasks over set structured data, Zaheer et al. \cite{zaheer2017deep} construct a permutation-invariant model. For a countable set $X$ and a set $Y$, the function $f:X\rightarrow Y$ is a valid set function, i.e., invariant to the permutation of elements in $X$, if and only if it can be decomposed into the form: $\rho\big(\sum_{x\in X}\phi(x)\big)$, where $\phi$ and $\rho$ are appropriate transformations. As for an uncountable set $X$ with fixed size $M$, any continuous function $f$ defined on $X$, i.e., $f:\mathbb{R}^{d\times M}\rightarrow Y$, is permutation-invariant if and only if $f$ can be approximated arbitrarily closely by a function of the form $\rho(\sum_{x\in X}\phi(x))$. Therefore, any set function $f$ can be represented in this formulation:
\begin{equation}
\label{DeepSets invariant equation}
f(X)=\rho\Big(\sum_{x\in X}\phi(x)\Big).
\end{equation}
Based on this analysis, DeepSets is developed, capable of approximating any permutation-invariant function on the ground set $X$ by using universal function approximators, such as neural networks, for the transformations $\phi$ and $\rho$. The model contains two main operations: 1) Each element $x_m$ of the ground set $X$ is transformed into a representation $\phi(x_m)$ through the neural network $\phi$. 2) The representations $\phi(x_m)$ are summed to produce a single vector, which is then processed through network $\rho$. The key idea is to aggregate all representations via summation and then apply nonlinear transformations through networks. In particular, the intermediate layers within DeepSets, such as $\phi$, often exhibit permutation-equivariance, meaning that the processing of each element is independent of the order in which the elements are presented. This property ensures that the order of the elements does not affect their individual processing. Similar to Definition \ref{definition of Permutation-invariance}, we formally define permutation-equivariance as follows.
\begin{definition}[Permutation-equivariance]
\label{definition of Permutation-equivariance}
    For each $\mathcal{X} \in \mathbb{R}^{N \times m}$ and $\Pi \in \Pi_N$, if $g(\Pi \mathcal{X})=\Pi g(\mathcal{X})$ always holds, the function $g:\mathbb{R}^{N \times m} \rightarrow \mathbb{R}^{N \times n}$ is permutation-equivariant.
\end{definition}
The authors propose a novel formulation of permutation-equivariant functions, which can be represented as a neural network layer whose standard form is $g_{\Theta}(\boldsymbol{x})=\sigma(\Theta \boldsymbol{x})$, where $\Theta\in \mathbb{R}^{M\times M}$ is the weight vector and $\sigma:\mathbb{R}\rightarrow \mathbb{R}$ is a nonlinear function such as sigmoid function. It is proved that $g_{\Theta}:\mathbb{R}^M\rightarrow \mathbb{R}^M$ is permutation-equivariant if and only if all diagonal elements of $\Theta$ are equal and all off-diagonal elements are tied together, i.e., 
\begin{equation}
\label{DeepSets equivariant weight}
\Theta=\lambda\boldsymbol{I}+\gamma(\boldsymbol{11}^\top)
\end{equation}
where $\lambda,\gamma\in \mathbb{R}$, $\boldsymbol{1}=[1,...,1]^\top\in\mathbb{R}^M$ and $\boldsymbol{I}\in\mathbb{R}^{M\times M}$ is a identity matrix. Therefore, the neural network $g_{\Theta}(\boldsymbol{x})=\sigma(\Theta \boldsymbol{x})$ is permutation-equivariant if $\Theta=\lambda\boldsymbol{I}+\gamma(\boldsymbol{11}^\top)$, i.e., $g(\boldsymbol{x})=\sigma((\lambda\boldsymbol{I}+\gamma(\boldsymbol{11}^\top)) \boldsymbol{x})$.
The layer has several other variations when specifying the operations and parameters. In summary, the permutation-equivariant property of the intermediate layers in DeepSets ensures that each element is treated consistently regardless of its position in the set. The final symmetric aggregation combines these equivariant features in a permutation-invariant manner, ensuring that the order of elements does not affect the output.

\subsubsection{Generalization of DeepSets}
\
\newline There are some works generalizing DeepSets. Margon et al. \cite{maron2020learning} focus on a principled approach to learning from unordered set elements, particularly when the elements themselves exhibit inherent symmetries. The authors propose the Deep Sets for Symmetric elements (DSS) framework, which generalizes the DeepSets to accommodate additional symmetries of elements. The core innovation is the introduction of DSS layers, incorporating multiple linear layers $L$ that are equivariant to the permutations of the set and the inherent symmetries of the set elements, such as translational symmetry in images and rotational symmetry in 3D shapes. This symmetry is represented by a group $H$ which operates on the elements. Concretely, a DSS layer applies a transformation to each set element while also considering the aggregated information from the entire set. Based on Equation(\ref{DeepSets equivariant weight}), the DSS layer for a set $\{x_1,...,x_n\}\subseteq \mathbb{R}^d$ with symmetry group $H$ and feature dimension $d$ is defined by:
\begin{equation}
    L(X)_i=L_1^H(x_i)+L_2^H\Big(\sum_{j\neq i}^nx_j\Big),
\nonumber
\end{equation}
which generalizes DeepSets by applying linear $H$-equivariant functions $L_1^H, L_2^H$. The authors prove that DSS networks are universal approximators because the individual element-wise networks are universal for the symmetry group $H$, addressing the issue that restricting a network to be invariant or equivariant may reduce the expressive power \cite{maron2019universality}. Consequently, DSS layers can represent any function that respects the symmetries of the set elements and the set itself. In summary, the DSS framework extends DeepSets to problems involving symmetric elements, providing a comprehensive and theoretically grounded approach for learning from sets with intrinsic symmetries. Murphy et al. \cite{murphy2018janossy} propose Janossy pooling, a novel model for constructing permutation-invariant functions. Janossy pooling provides a universal method by representing a permutation-invariant function as the average of a permutation-sensitive function applied to all possible reorderings of the input sequence. However, the computational cost of summarizing all permutations and backpropagating gradients is pretty high. To solve this issue, the authors develop three approximation methods to trade off complexity and generalization: 1) Canonical orderings: Elements of the input sequence are reordered according to a predefined criterion, reducing the computational cost by avoiding the need to consider all permutations; 2) $k$-ary dependencies: The permutation-sensitive function is restricted to depend only on subsets of k elements at a time, reducing the number of permutations considered while still capturing important interactions; 3) Permutation sampling: During training, permutations are randomly sampled, leading to fewer permutations. These strategies enable Janossy Pooling to unify and generalize existing methods, achieving competitive performance on various tasks compared to state-of-the-art techniques. Notably, DeepSets can be seen as a special case of Janossy Pooling with 1-ary dependencies, where the function depends on individual elements without considering interactions beyond simple aggregation. In contrast, Janossy Pooling allows for k-ary dependencies, capable of capturing higher-order interactions within the data.

\subsubsection{Theoretical Analysis of DeepSets}
\
\begin{table}[t]
\caption{The comparison among research on expressiveness analysis with latent space dimension $L$}
\begin{tabular}{@{}lllll@{}}
\toprule
Research        & DeepSets \cite{zaheer2017deep}   & Wagstaff et al. \cite{wagstaff2019limitations} & Zweig et al. \cite{zweig2022exponential} & Wang et al. \cite{wang2023polynomial} \\ \midrule
   $L$     & $N+1$ & $N$ & $\exp(\min\{\sqrt{N},D\})$ & $\mathrm{poly}(N,D)$ \\
 $D$ & $D=1$    & $D=1$ &  $D>1$ & $D>1$    \\ \bottomrule
\label{The comparison among all research on expressiveness analysis}
\end{tabular}
\end{table}
\newline There have been several works conducted to analyze the theoretical properties of DeepSets as it is a fundamental set function learning approach. Wagstaff et al. \cite{wagstaff2019limitations} refer to permutation-invariant function $f$ represented by the formulation of Equation(\ref{DeepSets invariant equation}) as sum-decomposition, where the combination $(\rho,\phi)$ of function $\phi:\mathbb{R}\rightarrow Z$ and function $\rho:Z\rightarrow\mathbb{R}$ is a sum-decomposition with latent space $Z$ for function $f$, namely, the function $f$ is sum-decomposable via $Z$. They analyze the limitations of enforcing permutation-invariance using sum-pooling and derive a necessary condition that a sum-decomposition based model with universal function representation should satisfy. It is demonstrated that only if the dimension $L$ of latent space where the summation is located is no less than the set size $N$, the sum-decomposition based models can represent arbitrary continuous functions defined on a set with size $N$. To resolve the open question regarding the representation capabilities of high-dimensional DeepSets posed in \cite{wagstaff2019limitations}, Zweig et al. \cite{zweig2022exponential} conduct expressive power analysis of DeepSets. They indicate that DeepSets require an exponentially large width to approximate certain symmetric functions, implying that the dimension $L$ of latent space grows exponentially with the size $N$ and dimension $D$ of the input set. This analysis demonstrates that DeepSets may be inherently inefficient for representing certain high-dimensional symmetric functions unless it is enhanced with mechanisms enabling interactions between set elements. Wang et al. \cite{wang2023polynomial} further reveal the relationship between the latent space dimension $L$ and the expressive power of DeepSets \cite{zaheer2017deep}, overcoming the limitations of previous works that focus solely on one-dimensional features or complex analytic activations, which are impractical due to the exponential growth of $L$ with $N$ and $D$. Considering high dimensional features, i.e. $D>1$, the bounds of minimal latent space dimension $L$ are proved to be divided into two categories according to the encoding network $\phi$: 1) If $\phi$ applies a linear layer with power mapping, we can get $N(D+1)\leq L<N^5D^2$. 2) If $\phi$ applies a linear layer and an exponential activation function, we can get a tighter bound, $ND\leq L\leq N^4D^2$. The proposed bounds imply that it is sufficient to model the latent space of DeepSets with $L$ being $\mathrm{poly}(N,D)$ for the universal approximation of set functions. It is also demonstrated that continuous mappings $\phi$ and $\rho$ are crucial for ensuring universal approximation of DeepSets. Table \ref{The comparison among all research on expressiveness analysis} compares the lower bounds of different research. 

\subsubsection{Proposing Novel Aggregating Methods}
\
\newline There are some works trying to propose novel aggregating methods for set function learning. Aggregating inputs into a single representation is a common mechanism in set function learning, such as DeepSets which utilizes sum-pooling to aggregate element-wise embeddings. Inspired by DeepSets, Abedin et al. \cite{abedin2019sparsesense} employ a set-based deep learning approach called SparseSense to handle the sparse data from passive sensors in human activity recognition (HAR) tasks. Unlike traditional methods that require dense data streams or rely on interpolation to estimate missing data points, SparseSense processes the sparse data directly, mitigating large estimation errors and long recognition delays. This method regards sparse sensor data as sets, allowing the model to focus on extracting discriminative features of all activity categories without relying on temporal correlations. The key idea is to apply a shared embedding network to project each set element into a higher-dimensional space, followed by feature-wise maximum pooling to aggregate these embeddings into a fixed-size global representation for activity classification. SparseSense extends DeepSets to HAR, demonstrating that set-based neural networks can effectively handle irregular data points and tolerate missing information. Bartunov et al. \cite{bartunov2022equilibrium} introduce an optimization-based aggregation method named Equilibrium Aggregation. This method generalizes existing pooling-based approaches, overcoming the limitations of existing techniques such as sum-pooling, which are constrained by their representational power. The Equilibrium Aggregation models the potential function $F_\theta(x,y)$, which quantifies the discrepancy between each set element $x$ and aggregation result $y$, as a learnable neural network with parameter $\theta$. This layer architecture can be integrated into another multi-layer neural network to aggregate sets. The energy-minimization of Equilibrium Aggregation can be formulated as:
\begin{equation}
\phi_\theta(X)=\arg\min\limits_{y}\,\Big(R_\theta(y)+\sum_{i=1}^NF_\theta(x_i,y)\Big),
\label{energy-minimization of Equilibrium Aggregation}
\end{equation}
where $R_\theta(y)$ is regularization. The aggregation result $y$ is computed through solving Equation(\ref{energy-minimization of Equilibrium Aggregation}) with numerical methods such as gradient descent. The neural network framework with Equilibrium Aggregation can be formulated as $\rho(\phi_\theta(X))$ where $\rho$ is a neural network. It is theoretically proved that this framework can universally approximate any continuous permutation-invariant functions if the output of Equation(\ref{energy-minimization of Equilibrium Aggregation}) has the same size as the input set. Equilibrium Aggregation provides a more flexible framework than DeepSets by utilizing a learnable potential function, potentially achieving better performance in tasks that require more detailed data representation. Horn et al. \cite{horn2020set} propose a novel framework called Set Functions for Time Series (SeFT) for classifying irregularly sampled time series. SeFT regards time series data as a set of observations, addressing the issues of irregular sampling and unaligned measurements without requiring imputation. This model employs a set function $f:S\rightarrow{\mathbb{R}^C}$ derived from Equation(\ref{DeepSets invariant equation}). Denoting $s_j$ as a single observation of the time series $S$, the function $f$ can be formulated as:
\begin{equation}
f(S)=g\Big(\frac{1}{|S|}\sum_{s_j\in S}h(s_j)\Big),
\nonumber
\end{equation}
where $h:\Omega\rightarrow\mathbb{R}^d$ and $g:\mathbb{R}^d\rightarrow\mathbb{R}^c$ are both neural networks, with $h$ mapping observations from the domain $\Omega$ to a $d$-dimensional latent space, and $g$ further mapping this latent representation to the final $c$-dimensional classification space. The variant of positional encoding \cite{vaswani2017attention} is used for time encoding, employing multiple trigonometric functions at different frequencies to convert 1-dimensional time $t$ of each observation into a multi-dimensional input. To handle large observation sets and highlight the most relevant data points, a weighted mean aggregating approach based on scaled dot-product attention with multiple heads is designed to weigh different observations. This aggregating method independently calculates each element's embedding, achieving a runtime and memory complexity of $\mathcal{O}(n)$. The SeFT extends the representation of DeepSets specifically to the time series with irregular sampling, where the order of observations is not fixed and might not follow a regular interval. 

\subsubsection{Extending DeepSets to Specific Scenarios}\
\newline There are multiple works proposing methods that build on the foundational concepts of DeepSets, extending these concepts to specific scenarios. Yi et al. \cite{yi2021cytoset} introduce CytoSet designed to deal with sets of cells and predict the clinical outcome of patients. Since the order of cells' profile has no biological relevance in flow and mass cytometry experiments, CytoSet regards the cytometry data as a set and extracts information through a permutation-invariant neural network based on DeepSets. This approach predicts the clinical outcome from the patient sample represented as a set of cells, with each cell characterized by a vector of protein measurements. In the proposed model, several permutation-equivariant blocks, as described in \cite{zaheer2017deep}, are stacked to transform the representation of each set element. The output of these blocks is processed by max-pooling, which measures the presence of high response cells and produces an embedding vector for the set. This vector is then passed through fully connected layers to predict the clinical outcome. CytoSet extends the concept of DeepSets to clinical cytometry data analysis, generalizing CellCNN \cite{arvaniti2017sensitive} and CytoDx \cite{hu2019robust}, and achieving better experiments performance compared to them. Ou et al. \cite{ou2022learning} present equivariant variational inference for set function learning (EquiVSet) to predict set-valued outputs (subsets) that optimize a certain utility function over a given ground set under the optimal subset (OS) supervision oracle, where the optimal subset provides the maximum utility. They combine energy based method with DeepSets to construct an appropriate set mass function that increases monotonically with a set utility function. To enable training models on varying ground sets and overcome the instability caused by the high dimension of sets when directly optimizing likelihood, a scalable training and inference algorithm is proposed by utilizing the maximum likelihood principle in conjunction with mean-field inference as a surrogate. The EquiVSet improves DeepSets to model the utility function explicitly and handle more complex tasks involving OS oracles. Wang et al. \cite{wang2023deep} develop an effective model termed as DTS-ERA, which combines the proposed Deep Temporal Sets (DTS) with Evidential Reinforced Attentions (ERA) to uncover the signature behavioral patterns (SBPs) of multimodal data in behavior analysis of children with Autism spectrum disorder. DTS-ERA is implemented in the manner of few-shot learning, enabling it to effectively handle situations with limited data. DTS is a multimodal version of DeepSets, capable of capturing complex temporal and spatial relationships in multimodal data. It is composed of a temporal encoder and a spatial encoder, which generate feature representations that maintain temporal dependencies and spatial locality. These feature representations are then concatenated and aggregated through an average-pooling to obtain the deep-set encoding. In ERA, DTS is combined with reinforcement learning agent where an evidential reward function is designed to learn an epistemic policy, which selects representative embeddings as attention signatures. ERA incorporates evidential learning to estimate uncertainty, allowing the model to distinguish between known and unknown regions effectively, thereby improving the reliability of the predictions. 

\subsection{PointNet Based Methods}
\label{section of PointNet based methods} 

PointNet \cite{qi2017pointnet} is another important and pioneering set function learning approach, particularly designed to deal with point clouds, taking the point sets as input and outputting labels. In this section, we introduce the fundamental concepts of PointNet and discuss several relevant works based on it. The structure of this section is outlined in Figure \ref{structure figures}\textcolor[RGB]{109,35,130}{(c)}.

\subsubsection{PointNet}\
\newline PointNet can directly process point clouds without converting them into regular data structures such as 3D voxel grids, maintaining the inherent properties of point clouds. The components of PointNet are similar to DeepSets while only replacing the sum-pooling with max-pooling. For a finite point set $X$ and its element $x$, the set function $f:2^X\rightarrow Y$ whose value corresponds to the semantic label of point set can be approximated by the PointNet formulated as
\begin{equation}
f(X)=\rho\Big(\mathop{\max}_{x\in X}\phi(x)\Big),
\nonumber
\end{equation}
where $\phi$ captures features of each point in $X$. These features are aggregated through max-pooling and then passed to $\rho$ to obtain the output. Both $\rho$ and $\phi$ are neural networks or other parameterized models with learnable parameters. This framework is permutation-invariant because max-pooling can process arbitrary orders of points in the point set and obtain the same output. It is theoretically proved that PointNet is capable of approximating any continuous set function if the max-pooling layer contains enough neurons.

\subsubsection{Theoretical Analysis of PointNet}\
\newline Chrisian et al. \cite{bueno2021representation} explore the expressive power of neural networks that use set pooling mechanisms. The authors introduce and analyze a variety of set pooling architectures, such as sum-pooling (DeepSets), max-pooling (PointNet), and average-pooling (normalized-DeepSets). The theoretical analysis reveals that PointNet cannot generally approximate averages of continuous functions over sets (e.g., center-of-mass), and DeepSets is strictly more expressive than PointNet in the constant cardinality setting. This finding implies that the choice of set pooling function has a dramatic impact on the expressiveness of these networks. Unexpectedly, it is also proved that any function which can be uniformly approximated by both PointNet and normalized-DeepSets should be constant under the unbounded cardinality setting. 

\subsubsection{Improving Capabilities of PointNet}\
\newline There are several works that enhance PointNet, extending its applicability to more complex scenarios. To overcome the limitation that PointNet cannot learn the local structures at various scales, Qi et al. \cite{qi2017pointnet++} develop a hierarchical neural network termed PointNet++, which applies PointNet recursively to nested partitions of the input point set. PointNet++ contains multiple set abstraction levels, including sampling layer, grouping layer, and PointNet layer. The sampling layer chooses points to define local regions' centroids, around which the grouping layer explores neighboring points to build local regions. Then the PointNet layer encodes local region patterns into feature vectors. In particular, the grouping layer has two implementations: multi-scale grouping and multi-resolution grouping. These methods are capable of adaptively aggregating multi-scale features with respect to corresponding point densities, thereby eliminating the impact of varying point set densities on different regions. Generally, PointNet++ begins by extracting local features that capture fine geometric structures within small neighborhoods through PointNet. These local features are then grouped into larger units and further processed to generate higher-level features. Such hierarchical process is repeated iteratively until the comprehensive features of the entire point set are obtained, realizing both robustness and detail capture. While PointNet uses a global max-pooling operation to aggregate features from the entire point set, PointNet++ enhances it by introducing a multi-scale hierarchical learning process, expanding its capabilities to capture detailed local structures and handle varying point densities. In contrast to PointNet that directly processes point cloud by considering each point independently and utilizing max-pooling to aggregate global features, Prokudin et al. \cite{prokudin2019efficient} design a type of residual representation termed as basis point sets (BPS), which can encode a point cloud into a fixed-length vector, enabling the use of standard machine learning techniques. To construct BPS, the point clouds are normalized to fit a unit ball with radius $r\in\mathbb{R}$, where we randomly sample $k\in\mathbb{R}$ points from a uniform distribution to obtain basis point set. By calculating the minimal distance from each basis point to the nearest point in the point cloud, we obtain feature vectors for every point cloud. These feature vectors can be taken as inputs of the learning algorithms. The point cloud classification experiments demonstrate that the framework combining MLP with BPS achieves performance comparable to PointNet, while significantly reducing the number of parameters and computational complexity.

\subsection{Set Transformer Based Methods}
\label{section of Set Transformer based methods}
In this section, we introduce Set Transformer \cite{lee2019set}, a powerful neural network designed for learning functions on sets, and discuss several set function learning methods built upon it. This section is organized as in Figure \ref{structure figures}\textcolor[RGB]{109,35,130}{(d)}.

\subsubsection{Set Transformer}\
\newline Set Transformer \cite{lee2019set} is an attention-based neural network method capable of modeling interactions across input set elements, which are often overlooked by set pooling methods such as DeepSets \cite{zaheer2017deep} and PointNet \cite{qi2017pointnet}. Based on Transformer \cite{vaswani2017attention}, Set Transformer employs permutation-invariant self-attention to capture pairwise and higher-order interactions between elements. Suppose that $Q,R\in\mathbb{R}^{n\times d}$ are query set and value set respectively, consisting of $n$ $d$-dimensional vectors. To construct the Set Attention Block (SAB), the authors employ the Multihead Attention Block (MAB), which is a variant of the Transformer’s encoder, with positional encoding and dropout removed. Given matrices $X,Y\in\mathbb{R}^{n\times d}$, the MAB with parameter $\omega$ is defined as follows:
\begin{equation}
\label{MAB in Set Transformer}
\mathrm{MAB}(Q,V)=\mathrm{LN}\big(H+\mathrm{rF}(h)\big), \,\, h=\mathrm{LN}\big(X+\mathrm{Multihead}(X,Y,Y;\omega)\big),
\end{equation}
where $\mathrm{LN}$ is layer normalization \cite{ba2016layer} and $\mathrm{rF}$ is an arbitrary row-wise feed-forward layer. The SAB can be formulated as $\mathrm{SAB}(X)=\mathrm{MAB}(X,X)$.
The higher-order interactions of elements can be captured by stacking multiple SABs. In order to reduce the high computational cost associated with self-attention, the Induced Set Attention Block (ISAB) is designed based on SAB and inspired by inducing point methods used in sparse Gaussian processes. The ISAB containing $m$ inducing points $I$, i.e., $m$ trainable $d$-dimensional vectors $I\in\mathbb{R}^{m\times d}$, can be formulated as:
\begin{equation}
\label{ISAB in Set Transformer}
\begin{split}
    &\mathrm{ISAB}_m(X)=\mathrm{MAB}(X,h)  \in\mathbb{R}^{n\times d},\, \,h=\mathrm{MAB}(I,X) \in\mathbb{R}^{m\times d},
\end{split}
\end{equation}
where $h$ is permutation-equivariant to $X$ and $\mathrm{ISAB}_m(X)$ is permutation-invariant to $X$. In this way, the computational time is reduced from $O(n^2)$ in SAB to $O(mn)$ in ISAB. The Pooing by Multihead Attention (PMA) with $k$ seed vectors $S\in\mathbb{R}^{k\times d}$, i.e., $\mathrm{PMA}_k(Z)=\mathrm{MAB}\big(S,\mathrm{rF}(Z)\big)$, is developed to aggregate encoding features set $Z\in\mathbb{R}^{n\times d}$. This mechanism allows the model to adaptively weigh the importance of different elements in the set, which is particularly useful in scenarios requiring multiple correlated outputs, such as clustering tasks. Generally speaking, in Set Transformer, the input set is encoded by the stacking of SABs or ISABs, followed by aggregation using PMA. The aggregated representation is then passed through a feed-forward network to produce the output. It is theoretically proved that Set Transformer is capable of universally approximating any set function.

\subsubsection{Improving Set Transformer}\
\newline Through gradient analysis, Zhang et al. \cite{zhang2022set} indicate that DeepSets and Set Transformer probably suffer from vanishing and exploding gradients when stacking more layers. They also observe that layer normalization discourages performance because the invariance of layer norm decreases the representation power and drops potentially useful information for prediction. To tackle such issues and make these set neural networks deeper, DeepSets++ (DS++) and Set Transformer++ (ST++) are developed by proposing equivariant residual connections (ERC) and set norm. ERC is a refined residual connection approach adhering to the clean path principle, capable of avoiding potential gradient issues by maintaining a clean path from input to output. The set norm is a novel normalization layer, which standardizes each set over the minimal number of dimensions and transforms features individually. This mechanism preserves most of the mean and variance information, avoiding the invariance issues associated with layer normalization. By integrating ERC and set norm into the encoders of DeepSets and Set Transformer respectively, the enhanced models DS++ and ST++ are constructed. These improvements enable the models to achieve greater depth and comparable performance, effectively addressing the instability issues in the original versions.

\subsubsection{Employing Set Transformer as Encoder}\
\newline There are several works employing the Set Transformer as encoders to construct new models for set function learning. Li et al. \cite{li2020reciptor} propose an effective recipe representation learning model named Reciptor, which jointly processes ingredients and cooking instructions. The ingredient set is encoded by Set Transformer, enhancing the model’s ability to capture interdependence among elements. A pretrained skip-instruction model is employed to encode the cooking instructions, generating initial embeddings and providing a context-aware representation of the entire cooking process. These initial embeddings are subsequently processed by a forward LSTM to produce the final instruction embeddings. To further optimize the learned embeddings, the authors utilize a novel knowledge graph-based triplet sampling loss \cite{chechik2010large}, ensuring that semantically related recipes are closer in the latent space. The embeddings are refined by combining a triplet loss with a cosine similarity loss between ingredient and instruction embeddings. The Reciptor outperforms baselines on two newly designed downstream classification tasks. Based on Set Transformer, Gim et al. \cite{gim2021recipebowl} design a set-based cooking recommender called RecipeBowl, which processes a given set of ingredients and cooking tags to output corresponding ingredient and recipe choices. Set Transformer is employed as the encoder to build a comprehensive representation of the ingredient set. A two-way decoder maps the representation into two distinct embedding spaces: one for predicting missing ingredients and the other for recommending relevant recipes. The model is trained using a combination of negative likelihood loss based on Euclidean distances and cosine embedding loss for the recipe prediction task, ensuring that the predicted ingredients and recipes are aligned with their actual counterparts in the embedding space. 

Zhang et al. \cite{zhang2022relational} present efficient algorithms to address the challenges of relational reasoning in cooperative Multi-Agent Reinforcement Learning (MARL) with permutation-invariant agents. They leverage Set Transformer to implement complex relational reasoning among agents in MARL. Two algorithms are proposed, including model-free and model-based offline MARL algorithms. The model-free approach employs transformers to estimate the action-value function, incorporating a pessimistic policy to handle distributional shifts in offline settings. The model-based approach estimates the system dynamics with transformers, also utilizing a pessimistic policy. The key contribution is deriving generalization error bounds for transformers in MARL, demonstrating that these bounds are independent of the number of agents and less sensitive to the depth of the network. Jurewicz et al. \cite{jurewicz2022set} develop Set Interdependence Transformer (SIT), an efficient set encoder to solve set-to-sequence tasks. This set-to-sequence model is established by combining the SIT with a permutation decoder. Set transformer serves as the basic set encoder, learning permutation-equivariant representations of individual elements and permutation-invariant representations of the entire set. SIT enhances these representations with an augmented attention mechanism to capture higher-order interdependencies. The permutation decoder uses an improved pointer attention mechanism to select elements, forming coherent output sequences. This approach effectively handles sets of varying cardinalities and generalizes well to unseen set sizes, as shown in experiments.

\subsubsection{Extending Set Transformer to Meta-learning}\
\newline Lee et al. \cite{lee2022set} propose Meta-Interpolation, a universal task augmentation method designed for few-task meta-learning. Meta-Interpolation utilizes Set Transformer to process the embeddings of support and query sets from different tasks and learn a parameterized set function, mapping sets of task embeddings to new embeddings that mix features from different tasks. This process creates new tasks that have unique features drawn from the tasks being interpolated. Bilevel optimization is employed to jointly optimize parameters of the meta-learner and the set function. The upper-level optimization aims to minimize the loss on meta-validation tasks, ensuring that this augmentation strategy improves generalization. The lower-level optimization adapts the meta-learner to augmented tasks, reducing the risk of overfitting to the limited meta-training set. This method theoretically regularizes the meta-learner by enforcing a distribution-dependent regularization, which decreases the Rademacher complexity and thus improves the generalization. 

\subsubsection{Other Methods Utilizing Attention Mechanisms}\
\newline There are multiple works utilizing attention mechanisms to learn set functions, similar to Set Transformer.
Girgis et al. \cite{girgis2021latent} develop an encoder-decoder framework, Latent Variable Sequential Set Transformers, termed as AutoBots, to deal with the challenging task of predicting the future trajectories of multiple interacting agents. The permutation-equivariant encoder processes sequences of sets representing the agents states over time, incorporating both temporal and social information through multiple Multi-Head Self-Attention (MHSA) modules. The decoder utilizes multiple matrices of learnable seed parameters, enabling the model to capture multi-modal nature of future trajectories. This process allows for the generation of diverse and socially consistent predictions across the entire scene in a single forward pass. The model achieves state-of-the-art performance, particularly in trajectory predictions that adhere to real-world constraints such as road layouts. Zhao et al. \cite{zhao2021point} propose Point Transformer, a novel architecture tailored for unordered 3D point sets. Point Transformer mainly contains SortNet and local-global attention module. SortNet is a novel neural network that learns to sort the input point cloud data into a specific order based on selected features. Once the points are sorted, the Point Transformer layer applies local attention to aggregate features of each point from its $k$ nearest neighbors, capturing fine-grained details and local geometric structures within small regions. Following the local attention, global attention aggregates features from the entire point cloud or larger regions, complementing the local information and enabling the network to understand the overall structure. The outputs from the local and global attention modules can be combined, either through concatenation or a weighted sum, to form a comprehensive feature representation for each point. This representation can be used in downstream tasks for learning the underlying shape. 

\subsection{Deep Set Prediction Network Based Methods}
\label{section of Deep set prediction networks based methods}
This section introduces Deep Set Prediction Network, an effective method to solve set prediction problems, and discusses relevant works that build on DSPN to enhance its capabilities. The structure of this section is illustrated in Figure \ref{structure figures}\textcolor[RGB]{109,35,130}{(e)}.

\subsubsection{Deep Set Prediction Network}\
\newline Deep Set Prediction Network (DSPN) \cite{zhang2019deep} is a model designed to predict sets from feature vectors, addressing the issue that previous methods such as RNNs result in discontinuous and inaccurate predictions due to lack of consideration on the unordered nature of sets. DSPN employs the same module for both encoding and decoding processes. Concretely, the encoder $g_{enc}$ maps the input set $X$ into the latent space $z$, i.e. $z=g_{enc}(X)$, obtaining the representation of $X$. The decoder $g_{dec}$ predicts a set from this representation, i.e., $\hat{X}=g_{dec}(z)$, applying gradient descent with a learnable initial guess to find a set whose latent representation matches the input set. This process can be regarded as a nested optimization. In the inner loop, the predicted set is refined iteratively to minimize the difference between its encoding and the target representation. Meanwhile, in the outer loop, the weights of the model are trained by minimizing the loss between the predicted set and the true set. Formally, the representation loss and decoder are defined as: 
\begin{equation}
\label{Deep Deep Set Prediction Networks}
\begin{split}
L_{repr}(\hat{X},z)=||g_{enc}(\hat{X})-z||^2, \,\,\,\,\,
g_{dec}(z)=\mathop{\arg\min}\limits_{\hat{X}}L_{repr}(\hat{X},z),
\end{split}
\end{equation}
where the permutation-invariant $L_{repr}$ compares $\hat{X}$ with the latent representation of $X$. Considering $g_{enc}$ is a neural network, the gradient descent is utilized for $T$ steps to solve the minimization of Equation(\ref{Deep Deep Set Prediction Networks}) from initial set $\hat{X}^{(0)}$. At the same time, the weights of $g_{enc}$ are trained to minimize set loss $L_{set}(\hat{X}^{(T)},Y)$, where the $L_{set}$ can represent Chamfer loss or pairwise loss, to obtain an appropriate representation $z$. In general set prediction, there is no set encoder since the input is usually a vector instead of a set, in which case, a term is added to the loss of outer loop to ensure $g_{enc}(X)\approx z$. The DSPN shows significant improvements over traditional methods, particularly in providing accurate set predictions without requiring complex post-processing. This work opens up new possibilities for set prediction problems.

\subsubsection{Improving DSPN}\
\newline There are several works that improve DSPN and extend its application to specific scenarios. Zhang et al. \cite{zhang2019fspool} design a differentiable set pooling method called FSPool, which consists of two operations, sorting and weighted summation. Instead of treating set elements as whole units, FSPool sorts each feature independently across the set elements. After sorting, a weighted sum is computed, with the weights determined by a learnable calibrator function. FSPool can handle sets of different sizes using a continuous representation of weights. In experiments involving both bounding box and state prediction, the authors combine DSPN and other models, such as MLP, with FSPool, max-pooling, and sum-pooling respectively. The results indicate that simply replacing the pooling function in an existing model with FSPool leads to better results and faster convergence.
By replacing the gradient descent updates of DSPN with transformer that provides more expressive and efficient updates, Kosiorek et al. \cite{kosiorek2020conditional} propose Transformer Set Prediction Network (TSPN), where an MLP is utilized to predict the number of points from the input embedding and decide the size of the initial predicted set. TSPN initializes the predicted set with a random set of points sampled from a learned distribution, enhancing flexibility. The transformer is employed to iteratively update set elements, leveraging self-attention mechanism to model dependencies between elements and output the predicted set. Compared to DSPN, TSPN achieves more expressiveness and less computational cost. Zhang et al. \cite{zhang2020set} develop a framework called Deep Energy-based Set Prediction (DESP), which treats set prediction as a problem of conditional density estimation rather than optimization with set-specific losses. This method utilizes deep energy-based models to capture the distribution of sets given some input features. The energy function $E_\theta(x,Y)$ assigns a scalar energy to a pair of input features $x$ and a set $Y$, where a lower energy indicates a higher likelihood of the set. Given the input, the probability of a set is $P_\theta(Y|x)=\frac{1}{Z(x;\theta)}\exp\big(-E_\theta(x,Y)\big)$,
where $Z(x;\theta)$ is a partition function. In the proposed framework, two permutation-invariant energy functions $E_{DS}(x,Y)$ and $E_{SE}(x,Y)$ are derived from DeepSets and DSPN respectively. These energy functions can be used to formulate deep energy based models, which can be trained by minimizing the negative log-likelihood, allowing for the approximation of the true data distribution without requiring explicit pairwise comparison between predicted and ground truth sets. DESP utilizes a stochastically augmented prediction algorithm, which helps explore multiple modes to generate diverse outputs. The final predicted set is determined as the set with the lowest energy among all captured sets. DESP extends the capabilities of DSPN to more effectively handle the inherent complexity and stochasticity of real-world tasks. 

Zhang et al. \cite{zhang2021multiset} define a relaxation of common set-equivariance, the multiset-equivariance, which does not require equal elements in a multiset to remain equal after transformation. This property is crucial for handling multisets with duplicate elements, enabling models to process them with different strategies. Additionally, the authors propose exclusive multiset-equivariance, which describes models that are multiset-equivariant but not set-equivariant, aiming to tackle the issue that set-equivariant functions cannot represent certain functions on multisets. It is proved that DSPN satisfies the exclusive multiset-equivariance when selecting the appropriate set encoder. To reduce memory and computational requirements, the implicit DSPN (iDSPN) is developed by employing approximate implicit differentiation to replace the gradient descent of DSPN. This method avoids storing intermediate gradient steps by directly computing the gradient at the optimal point, making the optimization process more efficient. iDSPN shows superior performance compared to traditional set-equivariant models, especially in handling multisets and large-scale set prediction tasks.

\subsection{Deep Submodular Function Based Methods}
\label{section of Deep Submodular Function based methods}
Submodular function is a significant subclass of set function, in which case, learning submodular function is an important area within set function learning. In this section, we introduce Deep Submodular Function (DSF) which focuses on learning submodular functions, and discuss related research that improves the capabilities of DSF. The structure of this section is outlined in Figure \ref{structure figures}\textcolor[RGB]{109,35,130}{(f)}.
We begin this section by introducing the definition of submodular function.
\begin{definition}[Submodular function]
\label{definition of submodular function}
    If for two sets $V,Y$ and a set function $f: 2^V \rightarrow Y$, given any two sets $A,B\subseteq V$, it holds that $f(A)+f(B) \geq f(A\cup B)+f(A\cap B)$, the set function $f$ is submodular. In particular, when it holds that $f(A) + f(B) = f(A\cup B) + f(A\cap B)$, then the set function $f$ is modular.
\end{definition}
Submodular functions are extensively used in machine learning \cite{dolhansky2016deep,iyer2021submodular,lee2019set,zaheer2017deep} and they have several important properties: 1) Diminishing returns: For any two sets $A,B$ such that $A \subseteq B \subseteq V$ and $s\notin B$, given a submodular function $f$, the diminishing returns can be denoted as $f(A\cup \{s\})-f(A) \geq f(B\cup \{s\})-f(B)$, which means the incremental gain of adding an element to a set decreases as the set becomes larger. 2) Natural concavity: Submodular functions are viewed as the discrete analog of concave functions because the property of diminishing returns is akin to the definition of concavity. 3) Modularity: This property implies additivity, meaning that for disjoint sets $A$ and $B$, it holds that $f(A\cup B)=f(A)+f(B)$. Modularity simplifies machine learning tasks by ensuring linearity \cite{blondel2022efficient,doerr2020optimization}, particularly in feature selection \cite{dolhansky2016deep} where it facilitates the computation of the relevance of feature sets. 4) Monotonicity: The value of a monotone submodular function does not decrease when additional elements are added to a set. 

\subsubsection{Deep Submodular Functions}\
\newline Deep Submodular Functions (DSFs) \cite{dolhansky2016deep} are a special class of submodular functions. which strictly generalizes many existing submodular functions and inherits some properties from them. For example, DSF can represent decomposable submodular functions, which can be expressed as sums of concave functions composed with modular functions. A notable subclass of these functions is the feature-based submodular function \cite{wei2014unsupervised}, which can be formulated as $f(X)=\sum_{u\in U}w_u\phi_u\big(m_u(X)\big)$, where $\phi_u$ denotes non-deceasing univariate normalized concave function, $m_u$ represents feature-specific modular function and $w_u$ is feature weight, all of which are non-negative. To overcome the limitation that features themselves cannot interact in feature-based submodular functions, an additional layer of nested concave functions is employed, i.e., $f(X)=\sum_{s\in S}\omega_s\phi_s\big(\sum_{s\in U}w_{s,u}\phi_u(m_u(X))\big)$, where $S$ denotes a set of meta-features, $\omega_s$ represents meta-feature weight and $\phi_s$ is a non-decreasing concave function. The term $w_{s,u}$ represents the feature weight corresponding to meta-feature $s$. By recursively applying the above layer, we can derive the DSF. Consider a series of disjoint sets $V^{(0)},V^{(1)},V^{(2)},...,V^{(K)}$, where $V^{(0)}$ is the ground set, $V^{(1)}$ is features set, $V^{(2)}$ is meta-features set, $V^{(3)}$ is meta-meta-features set and by analogy until $V^{(K)}$, with each set representing a layer. Denoting the size of $V^{(i)}$ as $d^i=|V^{(i)}|$, we can employ a matrix $w^{(i)}\in \mathbb{R}_+^{d^i\times d^{i-1}}, i\in \{1,2,...,K\}$ to connect two continuous layers. The element at row $v^i$ and column $v^{i-1}$ of matrix $w^{(i)}$  is denoted by $w_{v^i}^{i}$, where $w_{v^i}^{i}:V^{i-1}\rightarrow\mathbb{R}_+$ represents a modular function over set $V^{i-1}$. In this case, the matrix includes $d^i$ such modular functions. Moreover, given a non-negative non-decreasing concave function $\phi_{v^k}:\mathbb{R}_+\rightarrow\mathbb{R}_+$ and any set $A\subseteq V$, a $K$-layer DSF $f:2^V\rightarrow\mathbb{R}_+$ can be formulated as
\begin{equation}
f(A)=\phi_{v^K}\Big(\sum_{v^{K-1}\in V^{K-1}}w_{v^K}^{(K)}\big(v^{K-1}\big)\phi_{v^{K-1}}\Big(...\sum_{v^2\in V^{(2)}}w_{v^3}^{(3)}\big(v^2\big)\phi_{v^2}\Big(\sum_{v^1\in V^{(1)}}w_{v^2}^{(2)}\big(v^1\big)\phi_{v^1}\Big(\sum_{a\in A}w_{v^1}^{(1)}\big(a\big)\Big)\Big)\Big)\Big).
\nonumber
\end{equation}
Having shown the definition of DSF, it can be seen that DSF is composed of multiple layers, with each layer taking a positive linear combination of the previous layer's outputs followed by the application of a concave function. This hierarchical structure enables DSF to capture complex interactions within the data. The layered structured DSF shares similarities with deep neural networks (DNN), allowing for extending DNN learning techniques to DSF. The authors utilize a max-margin learning approach that is tailored to maintain the submodularity for training DSFs. This learning process involves adjusting the parameters of the DSF to maximize a margin-based objective function, ensuring that the learned function assigns high values to desired subsets while penalizing undesired ones.

\subsubsection{Extending DSF to Specific Scenarios}\
\newline There are some works that extend DSF to specific scenarios and improve its capabilities. Ghadimi et al. \cite{ghadimi2020deep} propose a novel model called deep submodular network (DSN), which combines the principles of deep learning with submodular optimization for multi-document summarization. DSN is similar to DSF, with the key difference being that DSN employs both modular and submodular functions to construct network blocks, whereas DSF only utilizes modular functions. Consequently, DSN generalizes DSF, making it applicable in a wider range of scenarios. The DSN utilizes the L-BFGS-B algorithm  \cite{byrd1995limited} for training, which is memory-efficient and suitable for maintaining non-negative weights. Manupriya et al. \cite{manupriya2022improving} design a novel approach called Submodular Ensembled Attribution for Neural Networks (SEA-NN), which aims to interpret the contribution of each input feature to the neural network's output, particularly in image-based tasks. The core component of SEA-NN is a submodular score function learned by DSF, which achieves the combination of several existing gradient-based attribution methods, such as Integrated Gradients and Smooth Integrated Gradients, and integrates their score bias. It is trained using heatmaps generated by baseline attribution methods, to increase the score for features that are highly relevant and specific. The learned scoring function re-evaluates the importance of features in the input by assessing the marginal gain of each feature, reducing the attribution scores of redundant features that may be present in the raw attribution maps. The SEA-NN is model-agnostic and can be applied to various scenarios. 

\subsubsection{Addressing the Limitation of DSF}\
\newline DSF models submodular functions as an aggregation of modular concave functions, but it does not provide methods for selecting these concave methods, complicating the practical application. To eliminate this gap, De et al. \cite{de2022neural} introduce novel neural networks, FLEXSUBNET, to estimate both monotone and non-monotone submodular functions. FLEXSUBNET models submodular functions by recursively applying concave functions to modular functions. This neural network allows for learning these concave functions from data, enhancing the expressiveness. The core of FLEXSUBNET is a simple recursive chain that is a restriction of complex topology, with each node in the chain sharing the learnable concave function. According to the monotonicity of learned functions, two scenarios are considered: 1) For monotone submodular function, it is learned through a recursive model. At each step, the model calculates a linear combination of a previously computed submodular function and a modular function, which is then processed by a learnable concave function to produce a composed submodular function. 2) For non-monotone submodular functions, it is also learned through a recursive model, where a non-monotone concave function is applied to a modular function. The model can be trained using (set, value) pairs or (perimeter-set, high-value-subset) pairs, with applications in subset selection tasks where high-value subsets need to be extracted from larger sets.

\subsection{Other Deep Learning methods}
\label{section of Other deep learning methods} 
In this section, we introduce other deep learning methods for set function learning problems, such as GaitSet, RepSet, and so on, which cannot be classified into the above several groups. 

Skianis et al. \cite{skianis2020rep} propose RepSet, a novel permutation-invariant neural network designed to address learning problems over sets of vectors. This model generates several hidden sets, with each containing a set of $d$-dimensional vectors. The correspondence between the input set and these hidden sets is established using a bipartite matching algorithm. These hidden sets can be updated through backpropagation during training to obtain the representation, which is then passed to a fully-connected layer to compute the output. In addition, ApproxRepSet, a relaxed version of RepSet that leverages fast matrix computations, is designed to handle large sets efficiently.

Li et al. \cite{li2020exchangeable} design an ordinary differential equations (ODE) based method called Exchangeable Neural ODE (ExNODE), capable of extracting the interdependences between set elements. ExNODE can be applied to both generative tasks and discriminative tasks. For generative tasks, the method employs continuous normalizing flows to model the distribution of sets and generate new samples. In set classification tasks, the embedding vector $v$ of input $x$ is $v=\mathrm{MaxPool}(\mathrm{ExNODE}\:\mathrm{Solve}(\phi(x)))$, where the linear function $\phi$ expands the feature dimensions of each input set element and the ExNODE learns the feature representations, which are aggregated by max-pooling. This model achieves fewer parameters and greater efficiency in point cloud classification and set likelihood estimation tasks.

Chao et al. \cite{chao2019gaitset} develop a gait recognition network called GaitSet, which regards gait as a set of independent frames containing gait silhouettes. In this framework, a CNN is employed to extract features from each frame of the gait set independently, capturing detailed spatial information. The Multilayer Global Pipeline extracts features at different levels from multiple layers of the CNN, combining them to form a comprehensive representation to preserve details of paces. These extracted frame-level features are then aggregated into set-level features by set pooling such as mean-pooling and attention mechanisms. The set-level features are further processed by Horizontal Pyramid Mapping, which splits the feature map into strips at multiple scales. This approach allows the model to capture both global and local features, enhancing the discriminative power of the representation. The experiments demonstrated the model's effectiveness with a limited number of frames and its ability to integrate information from different levels.

Shi et al. \cite{shi2020deep} propose a novel machine learning estimation method called Deep Message Passing on Sets (DMPS), which is designed to handle set-structured data by incorporating relational learning, bridging the gap between learning on graphs and learning on sets. This method begins by constructing a latent graph that represents the relational structure between set elements. This is achieved through a deep kernel learning approach, where each set element is transformed into a feature space and a kernel function is applied to capture similarities between elements. The message passing is adopted to update each set element based on a weighted sum of all elements, leveraging relational information. In particular, a stack of message passing is utilized to capture higher-order dependencies between set elements, but resulting in over-smoothing and vanishing gradient. To address such issues, two modules, Set-Denoising and Set-Residual blocks, are designed to integrate with DMPS. The Set-Denoising block eliminates over-smoothing by combining the original and updated features of set elements. Meanwhile, the Set-Residual block maintains distinctive features among set elements, preventing feature homogenization. 

Guo et al. \cite{dan2021learning} develop a prototype-oriented optimal transport (POT) approach to improve representation learning for set-structured data. In this framework, given a set $j$, a distribution $Q_j$ over these prototypes is computed, and each set is represented by a vector $h_j$, which indicates the mixture of prototypes relevant to that set. To align the distribution $Q_j$ with its empirical distribution $P_j$ over set elements, this model employs an Optimal Transport (OT) distance, which measures the effort required to transform $P_j$ into $Q_j$, providing a natural mechanism for training the model to capture the set’s statistics. The objective is to minimize this OT distance, encouraging the model to learn both global prototypes and set-specific representations $h_j$ effectively. POT can be integrated into existing architectures like summary networks and applied to different types of tasks, such as few-shot classification and meta-generative modeling.

Zhang et al. \cite{zhang2018learning} design an innovative model that optimizes permutation matrices to learn representations of set-structured data. The key component is the Permutation-Optimization (PO) module, which rearranges sets by minimizing a cost function via gradient descent. Given an input set $X$ represented as a matrix $X = [x_1, x_2, \ldots, x_n]^T$, where each $x_i$ is a feature vector, the algorithm initializes a permutation matrix $P^{(0)}$ either uniformly or through linear assignment. The total cost function $c(P)$ is defined as
\begin{equation}
    c(P) = \sum_{i,j} C_{ij} \Big( \sum_k P_{ik} \Big( \sum_{k^\prime > k} P_{jk^\prime} - \sum_{k^\prime < k} P_{j^\prime} \Big) \Big),
\nonumber
\end{equation}
where $C_{ij}$ is the pairwise ordering cost between elements $i$ and $j$, and $k$ and $k^\prime$ are element positions. $C_{ij}$ measures permutation quality to determine the optimal element ordering.  However, the overall complexity of the algorithm is $\Theta(n^3)$ per iteration, making it impractical for large sets.

Previous set encoding methods, such as DeepSets \cite{zaheer2017deep} and Set Transformer \cite{lee2019set}, implicitly assume that the entire set can be stored in memory and accessible simultaneously, while it is unrealistic when processing large sets or streaming data. To overcome this limitation, Bruno et al. \cite{bruno2021mini} define a new property called Mini-Batch Consistency (MBC), which is essential for maintaining consistent set representations across different mini-batches. MBC is defined as the property that ensures the encoding of a full set is equivalent to the aggregation of its mini-batches' encodings. To obtain the set representation, the authors develop Slot Set Encoder (SSE), where each slot is a learnable vector that interacts with set elements to capture their features. The attention mechanism in SSE computes attention weights using the dot product of the slots and the set elements, followed by a sigmoid activation function, eliminating the batch-dependent normalization issue that breaks MBC. To capture the interactions across set elements, the hierarchical slot set encoder is constructed by combining a stack of SSEs with a final PMA module in Set Transformer. The SSE can process sets in mini-batches, making it suitable for real-world applications with large sets. To overcome the limitations that SSE only supports sigmoid activation and cannot adopt more expressive non-MBC modules, Willette et al. \cite{willette2023scalable} develop the Universal MBC (UMBC) framework, which extends SSE to more activation functions, such as softmax, enabling more expressive set encoders. The authors also propose an efficient training algorithm that approximates the full set gradient by aggregating gradients from subsets of the set, maintaining constant memory overhead. This approximation provides an unbiased gradient estimate, significantly outperforming biased estimates derived from randomly sampled subsets. Notably, the UMBC framework is capable of universally approximating any continuous permutation-invariant function. 

\section{Other methods}
\label{section of Other methods}
This section discusses several non-deep learning set function methods, such as kernel and decision trees based methods. 

Kernel based approaches for learning set functions typically define a distance or similarity measure (kernel) to establish correspondences between sets. This measure is often combined with instance-based machine learning methods, such as Support Vector Machines (SVM). Nikolentzos et al. \cite{nikolentzos2017matching} utilize the Earth Mover’s Distance metric and SVM with Pyramid Match Graph Kernel. Buathong et al. \cite{buathong2020kernels} propose more efficient kernel methods by leveraging Reproducing Kernel Hilbert Space (RKHS) embeddings. They introduce Double Sum (DS) kernels, which calculate the sum of kernel evaluations over all pairs of elements. But DS kernels often lack strict positive definiteness, limiting their applicability. To overcome this limitation, the authors develop Deep Embedding (DE) kernels, applying a radial kernel in Hilbert space over the canonical distance induced by DS kernels. The proposed kernel methods enhance the Gaussian Process models in prediction and optimization tasks with set-valued inputs. However, these kernel based methods usually suffer from high computational complexity and memory overhead since they compare all sets to each other.

Wendler et al. \cite{wendler2021learning} develop novel algorithms for learning Fourier-sparse set functions using non-orthogonal Fourier transforms within a discrete-set signal processing framework (DSSP) \cite{puschel2020discrete}, which generalizes classical signal processing to set functions. The proposed algorithm, Sparse Set Function Fourier Transform (SSFT), computes the non-zero Fourier coefficients by utilizing Fourier support, which refers to the set of indices where the Fourier coefficients are non-zero. The Fourier support is determined by iteratively restricting the set function to subsets of its domain and identifying those subsets where the Fourier transform of the restricted function significantly contributes to the original set function. This algorithm requires $O(nk-k\log k)$ queries and $O(nk^2)$ operations, where $n$ is the size of the ground set and $k$ is the number of non-zero Fourier coefficients, achieving significant improvements over the naive fast Fourier transform.

Lu et al. \cite{lu2024slosh} propose Set Locality Sensitive Hashing (SLoSH), an efficient algorithm for set retrieval by leveraging Sliced-Wasserstein Embedding (SWE) and Locality-Sensitive Hashing (LSH). The SWE embeds each set into a lower-dimensional space through random linear projections, followed by sorting and calculating the Monge couplings, preserving the properties of the Wasserstein distance. The computational complexity of this embedding process is  $O(Ln(d+\log n))$, where $n$ is the set size and $L$ is the number of projections. Having obtained the vector representation, the LSH, sensitive to the Sliced-Wasserstein distance, is employed to find approximate nearest sets. The authors provide theoretical bounds for the SLoSH, ensuring the computational efficiency of both the embedding and hashing steps. 

Feldman et al. \cite{feldman2013representation} focus on the complexity of learning submodular functions on the Boolean hypercube $\{0, 1\}^n$. They prove that any submodular function $f$ can be approximated within $\epsilon$ in $\ell_2$ norm by a real-valued decision tree of depth $O(1/\epsilon^2)$. The function $f$ is represented as a binary decision tree $T$ with a rank of at most $4/\epsilon^2$, ensuring $\|T - f\|_2 \leq \epsilon$. Leveraging this approximation, the authors develop a Probably Approximately Correct (PAC) learning algorithm for submodular functions. This algorithm runs in time $\tilde{O}(n^2)\cdot 2^{O(1/\epsilon^4)}$, where $n$ is the number of variables, significantly improving learning efficiency under the uniform distribution. They also establish an information-theoretic lower bound of $2^{\Omega(1/{\epsilon}^{2/3})}$ and a computational lower bound of $n^{\Omega(1/{\epsilon}^{2/3})}$, implying optimality (up to the constant in the power of $\epsilon$) of their algorithms. Raskhodnikova et al. \cite{raskhodnikova2013learning} introduce a polynomial-time algorithm for learning submodular functions. This method builds on a structural result showing that any submodular function $f: \{0,1\}^n\rightarrow\{0,1,\ldots,k\}$ can be represented by pseudo-Boolean $2k$-disjunctive normal form (DNF) formula, which extends the traditional DNF formula to handle integer-valued functions, enabling learning submodular functions with techniques similar to those for Boolean functions. The authors propose a PAC learning algorithm, which is a generalization of Mansour’s PAC-learner for k-DNFs. This algorithm transforms the submodular function into a pseudo-Boolean k-DNF, applies random restrictions and utilizes Fourier analysis to identify significant coefficients.  The algorithm is efficient, with runtime polynomial in $n$, $k^{O(k \log k/\epsilon)}$, $1/\epsilon$, and $\log(1/\delta)$, where $\epsilon$ and $\delta$ are accuracy and confidence parameters respectively. The authors also establish lower bounds on the complexity of learning submodular functions, demonstrating the method's optimality.

\section{Applications and Relevant Datasets}
\label{section of Applications}
In this section, we introduce several applications of set function learning methods. These methods have shown great potential in scenarios where data can be naturally represented as sets, and the order of elements is not inherently important. As researchers continue to explore the capabilities of set function learning, the range of applications is expanding across various domains that require processing and reasoning over unordered sets of data.

\subsection{Point Cloud Processing}
Set function learning has the potential to revolutionize point cloud processing by treating the point cloud as a set of vectors, where each vector represents the features of a point. In point cloud applications, set function learning methods are mainly used for classification, segmentation, and detection tasks. As point cloud data is widely used in various applications, set function learning methods are likely to play an increasingly important role in this domain.

\textbf{Point cloud classification} aims to determine the category of objects represented by point clouds. Set function learning models such as DeepSets\cite{zaheer2017deep} PointNet \cite{qi2017pointnet}, Set Transformer \cite{lee2019set} and other methods \cite{aittala2018burst,bruno2021mini,hartford2018deep,kosiorek2020conditional,maron2019provably,maron2020learning,murphy2018janossy,prokudin2019efficient} extract relevant features from the entire point set. This enables accurate identification of objects such as cars, trees, and buildings, which is particularly important in fields such as robotics \cite{arnold2021fast,zhang2022relational} and autonomous driving \cite{lang2019pointpillars,zhou2018voxelnet}.

\textbf{Point cloud segmentation} is the process of labeling each point in a point cloud with a specific class, allowing for detailed scene understanding, such as distinguishing between vehicles and people in autonomous driving \cite{hackel2017semantic3d,milioto2019rangenet++}. Advanced models such as PointNet++ \cite{qi2017pointnet++}, Point Transformer \cite{zhao2021point} and other methods \cite{dan2021learning,li2020exchangeable,willette2023scalable,zhang2020set,zhang2021multiset} achieve good performance on this task by considering both local and global features. 

\textbf{Point cloud detection} focuses on identifying and localizing objects within a point cloud, providing bounding boxes around detected items. This application leverages set function learning models \cite{lee2019set,lee2022set,maron2020learning,prokudin2019efficient,rezatofighi2021learn,shi2020deep,zhang2022set} to propose regions of interest and refine these regions for precise localization. Point cloud detection ensures the safe and efficient operation of autonomous driving \cite{engelmann20203d,yang2018pixor} and robotic navigation \cite{hornung2013octomap,salas2013slam++} in dynamic environments.

Empirical comparisons show that PointNet++ and Point Transformer generally outperform DeepSets, PointNet, and Set Transformer in point cloud tasks on ModelNet40 and ShapeNet datasets \cite{zhao2021point,fei2022dumlp}. In particular, Point Transformer achieves the best performance in point cloud classification and segmentation by leveraging self-attention mechanism. SpiderCNN and PointNet++ excel in segmentation tasks by incorporating hierarchical and local feature extraction techniques. Meanwhile, DuMLP-Pin offers competitive performance while significantly reducing computational complexity, demonstrating its efficiency for classification and segmentation. 

\subsection{Set Anomaly Detection}
Set anomaly detection is another crucial application, aiming to identify outliers within a set by leveraging set function learning models such as DeepSets \cite{zaheer2017deep}, PointNet \cite{qi2017pointnet} and other variants \cite{maron2020learning,zhao2021point,qi2017pointnet++,zhang2020set,ou2022learning}. The process begins by extracting feature representations for each element in the set, which are subsequently aggregated to produce a unified set representation. The core of anomaly detection is to compare each element's feature against this aggregated representation, assigning an anomaly score to each element based on its deviation from the overall set pattern. This is typically accomplished through a subsequent layer that evaluates the extent of deviation. The framework then outputs a probability distribution over the elements, with higher probabilities indicating a greater likelihood of being outliers. Set anomaly detection is vital in various domains, such as detecting unusual behavior in sensor networks \cite{poornima2020anomaly} and identifying outlier faces in image sets \cite{sun2014deep}. 

Experiment results on the CelebA dataset highlight different strengths among models for set anomaly detection \cite{fei2022dumlp}. PointNet and DeepSets offer simplicity but struggle to capture complex interdependencies within sets, limiting their performance. Set Transformer improves performance by incorporating attention mechanisms to model relationships among set elements. However, DuMLP-Pin outperforms these methods by achieving the highest accuracy while significantly reducing parameter complexity. This makes DuMLP-Pin a competitive choice for anomaly detection tasks, particularly in resource-constrained environments.

\subsection{Recommendation Systems}
Set function learning methods \cite{zaheer2017deep,lee2019set} enhance recommendation systems by modeling user-item interactions as sets and effectively capturing user preferences regardless of the ordering, which facilitates the accurate representation of user profiles. For example, Reciptor \cite{li2020reciptor} and Recipebowl \cite{gim2021recipebowl} both employ Set Transformer to capture relationships between input elements, enabling more accurate recipe recommendations. Set function learning methods also integrate contextual information into the set, enabling context-aware recommendations \cite{adomavicius2010context,zhang2019deep} that adapt to different situations for more relevant suggestions. Additionally, they ensure diversity \cite{kunaver2017diversity} and fairness \cite{burke2017multisided} in recommendations by producing sets that cover a wide range of items, improving collaborative filtering \cite{sedhain2015autorec} through better aggregation of similar user preferences or items. Moreover, these methods require fewer parameters and enable efficient training on large datasets, making them promising for enhancing recommendation system accuracy and relevance.

Empirical results for recommendation task on Recipe1M dataset show that Reciptor and RecipeBowl outperform DeepSets by effectively modeling ingredient relationships and recipe context, with RecipeBowl achieving the best performance \cite{li2020reciptor, gim2021recipebowl}. Reciptor excels in cuisine classification and region prediction, while RecipeBowl focuses on context-aware ingredient and recipe recommendations. In contrast, while DeepSets is computationally efficient, it struggles to capture complex dependencies, resulting in weaker performance.

\subsection{Set Expansion and Set Retrieval}
Set expansion involves identifying new objects that are similar to a given set of objects and retrieving relevant candidates from a large pool. This process is closely related to set retrieval, where the goal is to efficiently retrieve items from a large dataset that match the characteristics of a target set. In the text concept set retrieval task, the goal is to identify and retrieve words that belong to a specific concept or category based on a given set of example words. For example, starting with \{apple, orange, pear\}, the aim is to retrieve additional related words such as banana and watermelon, which belong to the same "fruit" category. This task can be viewed as set expansion conditioned on a latent semantic concept, where DeepSets \cite{zaheer2017deep} and its variants \cite{bartunov2022equilibrium,lee2019set,qi2017pointnet} are particularly effective. In computational advertising, set function learning methods \cite{lee2019set,qi2017pointnet,zaheer2017deep} improve advertisement targeting by expanding the set of user preferences or behaviors with additional relevant interests, ensuring advertisements more relevant and effective. Experimentally, DeepSets outperforms all the conventional baselines on the COCO dataset for set retrieval tasks \cite{zaheer2017deep}. 

\subsection{Time Series Prediction}
Set function learning methods \cite{horn2020set,yu2023predicting} in time series prediction address the challenges posed by irregular, sparse and asynchronous data through treating each time series as an unordered set of observations. This mechanism eliminates the need for data regularization with interpolation, enabling models to operate directly on raw data and capture the inherent information more effectively. Sparsesense \cite{abedin2019sparsesense} processes the sparse and irregular data streams generated from batteryless passive wearables to tackle the problem of human activity recognition (HAR). DTS-ERA \cite{wang2023deep} combines evidential reinforced attention with deep temporal sets for detailed behavioral pattern analysis.

Empirically, SEFT-ATTN achieves competitive performance on mortality prediction tasks by effectively handling asynchronous and unaligned data \cite{horn2020set}. SparseSense outperforms traditional baselines in sparse data-stream classification by directly learning from unordered observations without interpolation.\cite{abedin2019sparsesense}.  DTS-ERA demonstrates superior predictive accuracy on 2D, 3D, and mixed Maze Painting data, further showing its generalization ability in behavior analysis \cite{wang2023deep}.

\subsection{Multi-label Classification}
Multi-label classification aims to assign multiple labels to a single instance, which is complex due to the dependencies between labels. Set function learning methods \cite{ghadimi2020deep,qin2019adapting,zaheer2017deep,rezatofighi2021learn,skianis2020rep,zhang2019fspool} can explicitly model these dependencies, improving classification performance. For example, in image tagging, labels such as beach and sun are likely to appear simultaneously, and modeling this relationship can lead to more accurate predictions. In particular, submodular function learning methods \cite{dolhansky2016deep,ghadimi2020deep,manupriya2022improving,de2022neural}, with the property of diminishing returns, can be used to capture the idea that adding a label to a smaller set of labels is more informative than adding it to a larger set, which is especially useful for modeling the dependencies between labels.

Experimentally, FSPOOL outperforms DeepSets, PointNet and Janossy Pooling on the CLEVR dataset by utilizing sorting-based pooling \cite{zhang2019fspool}.
RepSet consistently achieves superior performance across datasets by effectively modeling set relationships with a bipartite matching mechanism \cite{skianis2020rep}. In addition, Set-JDS and set-RNN both demonstrate competitive accuracy across multiple datasets \cite{qin2019adapting,rezatofighi2021learn}.

\subsection{Molecular Property Prediction}
Set function learning methods \cite{lee2022set,bartunov2022equilibrium} have achieved significant progress in molecular property prediction, capable of handling complex molecular datasets and enhancing prediction accuracy. For example, EMTO-CPA \cite{zhang2022composition} applies DeepSets to the design of high-entropy alloys (HEAs). By treating the composition of alloys as sets of elements, DeepSets can predict the properties of novel HEAs more accurately. This approach facilitates the exploration of a vast compositional space and the discovery of new materials with desirable properties. In drug discovery, EquiVSet \cite{ou2022learning} is utilized for compound selection in virtual screening by modeling the hierarchical selection process of compounds. 

Empirical evaluations on molecular property prediction tasks show that EquiVSet outperforms DeepSets on the PDBBind dataset by effectively capturing complex dependencies in molecular structures \cite{ou2022learning}. Similarly, Equilibrium Aggregation demonstrates superior representational power by optimizing a potential function over molecular sets \cite{bartunov2022equilibrium}, achieving better performance than GCN and GIN on MOLPCBA \cite{hu2020open}. 

\subsection{Amortized Inference}
Set function learning methods have found significant applications in amortized inference, where we train neural networks to approximate posterior distributions, thus replacing traditional iterative inference approaches with efficient forward passes. 
Set Transformer \cite{lee2019set} addresses amortized clustering by efficiently mapping datasets to cluster structures through set attention blocks. To overcome the limitation that Set Transformer assumes a fixed number of clusters, Lee et al. \cite{lee2019deep} propose Deep Amortized Clustering, which extends Set Transformer by incorporating recursive filtering steps, capable of generating varying number of clusters depending on dataset complexity. Building on this foundation, Pakman et al. \cite{pakman2020neural} apply set based architectures to approximate posterior sampling for probabilistic clustering models, which has been demonstrated effective in applications like spike sorting for high-dimensional neural data. Wang et al. \cite{wang2020amortized} introduce Neural Clustering Processes, a framework that combines set attention with GNN for flexible and efficient amortized clustering. Beyond clustering, set neural architectures have also been applied to general probabilistic inference. For instance, the Neural Process family employs set neural architectures to efficiently model functional variability and uncertainty across datasets \cite{jha2022neural}. Additionally, Müller et al. \cite{mullertransformers} develop Prior-Data Fitted Networks, which train set neural networks on synthetic priors, achieving fast and scalable Bayesian inference for structured data. 

Experimentally, these methods consistently achieve state-of-the-art performance in amortized inference \cite{lee2019deep,pakman2020neural,wang2020amortized,mullertransformers}, significantly outperforming traditional methods. For instance, Set Transformer outperforms variational methods in clustering tasks, achieving the highest accuracy on benchmark Gaussian mixtures and real-world datasets \cite{lee2019set}. 

\subsection{Other Applications}
In addition to the applications we have previously summarized, set function learning is also applied in other domains, such as human activity recognition. GaitSet \cite{chao2019gaitset} regards gait sequences as sets of frames, capturing the invariant features of human gait across different views and enhancing the ability to recognize individuals based on their walking patterns. CytoSet \cite{yi2021cytoset} leverages set modeling to handle the unordered and variable-sized nature of single-cell cytometry data. By utilizing permutation-invariant neural networks, CytoSet can predict clinical outcomes directly from the set of cells, enhancing the model's ability to capture complex biological patterns. Similarly, set function learning models such as UMBC \cite{willette2023scalable} can be employed to process high-resolution tissue images, improving the accuracy of cancer detection. Empirically, GaitSet, CytoSet, and UMBC demonstrate state-of-the-art performance in human activity recognition, clinical outcomes prediction, and cancer detection, respectively.

\subsection{Relevant Datasets}
\label{section of Datasets for set function learning}
In this section, we introduce some datasets that are commonly utilized to evaluate set function learning methods.

\subsubsection{Point Cloud Dataset}\
\newline There are some datasets commonly used for point cloud processing tasks.
\textbf{ModelNet40} dataset \cite{wu20153d} consists of 12,311 CAD models from 40 categories of man-made objects.
\textbf{ShapeNet} dataset \cite{yi2016scalable} contains 16,881 3D shapes from 16 categories, each annotated with 50 distinct parts. 
\textbf{Stanford 3D semantic parsing} dataset \cite{armeni20163d} includes Matterport 3D scans of 271 rooms across six areas, annotated with 13 semantic labels like chair, table, and floor.
\textbf{Point Cloud MNIST 2D} dataset converts MNIST \cite{lecun1998gradient} images into 2D point clouds, comprising 60,000 training and 10,000 testing samples, with each set containing 34–35 points.
\textbf{Oxford Buildings Dataset} \cite{philbin2007object} contains 5,062 images of eleven Oxford landmarks, each with 55 queries for evaluating object retrieval systems.


\subsubsection{Image Dataset}\
\newline We summarize some image datasets for anomaly detection, set retrieval, and multi-label classification.
\textbf{CelebA} dataset \cite{liu2015deep} contains 202,599 celebrity face images annotated with 40 boolean attributes, such as "smiling," "wearing glasses," and "blonde hair."
\textbf{Celebrity Together} dataset \cite{zhong2018compact} includes 194,000 images with 546,000 labeled faces, averaging 2.8 faces per image.
\textbf{MS COCO} dataset \cite{lin2014microsoft} comprises 123,000 images labeled with per-instance segmentation masks of 80 classes. Each image includes 0 to 18 objects, with most containing 1 to 3 labels.

\subsubsection{Recommendation Dataset}\
\newline
The following datasets are used to evaluate set function learning methods in recommendation systems.
\textbf{Amazon baby registry} dataset \cite{gillenwater2014expectation} contains 29,632 baby registries, each listing 5 to 100 products categorized into groups like "toys" and "furniture." 
\textbf{Recipe1M} dataset \cite{marin2021recipe1m+} consists of 1,029,720 cooking recipes with ingredients, instructions, images, and 1,047 semantic categories parsed from titles, covering 507,834 recipes.

\subsubsection{Chemical and Biological Dataset}\
\newline
Here are some datasets used for molecular property and hematocrit level prediction.
\textbf{Flow-RBC} dataset \cite{zhang2022set} contains 98,240 training and 23,104 test sets, each representing 1,000 red blood cells with volume and hemoglobin content measurements.
\textbf{PDBBind} dataset \cite{liu2015pdb} provides experimental binding data for 10,776 biomolecular complexes, including 8,302 protein–ligand and 2,474 other complexes.
\textbf{BindingDB9} dataset is a public database of measured binding affinities, consisting of 52,273 drug-targets with drug-like small molecules.

\subsubsection{Multi-modal Dataset}\
\newline The following datasets can be used for object detection and set property prediction.
\textbf{SHIFT15M} \cite{kimura2023shift15m} contains 15 million images and videos captured in diverse driving environments with annotations for object bounding boxes, instance segmentation masks, and semantic labels, covering vehicles, pedestrians, road signs, and more.
\textbf{CLEVR} \cite{johnson2017clevr} is a visual question answering dataset with 70,000 training images and 700,000 questions, plus additional validation and test sets. Questions fall into five types: existence, counting, integer comparison, attribute queries, and attribute comparisons. Each scene contains 3D-rendered objects characterized by size, shape, material and color, forming 96 unique combinations.

\section{Discussion and future directions}
\label{section of Conclusion and future directions}
In this survey, we have comprehensively reviewed and discussed various techniques to solve set function learning problems, covering deep learning and traditional learning methods. By investigating a wide range of methodologies, such as DeepSets \cite{zaheer2017deep} and Set Transformer \cite{lee2019set}, it is evident that significant progress has been achieved in learning complex set functions across various domains, from point cloud processing to recommendation systems. However, there still exist several challenges. A critical challenge is the lack of theoretical breakthroughs. Balcan et al. \cite{balcan2018submodular} introduce the probably mostly approximately correct (PMAC) model, extending the PAC model to real-valued functions. They demonstrate that submodular functions can be PMAC-learned with an approximation factor $O(n^{1/2})$ using a polynomial number of samples. But there still lacks research that focuses on the learnability of general set functions. Another major limitation is that most current methods assume that the entire set can be accessed at once, which is impractical for large sets due to memory constraints. Moreover, in streaming data scenarios, it is crucial that set representations can be updated in real time. Additionally, the potential and advantages of set function learning methods in specific fields have not been fully explored. These challenges highlight several open research directions worthy of further investigation.

\begin{itemize}
    \item \textbf{Theoretical analysis}: It is significant to conduct in-depth theoretical analysis for advancing set function learning. This involves analyzing learnability of various classes of set functions, assessing the expressiveness and limitations of different models, and establishing generalization bounds. Additionally, exploring the impact of set size and element distributions on model performance can reveal crucial factors affecting performance. 
    These theoretical advancements provide deeper insights into set function models and valuable guidance for designing more efficient and interpretable frameworks.
    
    \item \textbf{Mini-batch consistency}: It is vital to ensure stable predictions across mini-batches during training for resource-constrained environments. Developing techniques such as consistency regularization and batch normalization specifically designed for set inputs can mitigate instability arising from variations in set size and composition. Furthermore, investigating the impact of set composition, diversity, and size on mini-batch consistency enables the development of more stable and robust training strategies for set function models. 
     
    \item \textbf{Dynamic data handling}: In scenarios such as sensor networks, where data streams continuously, it is critical to develop adaptive architectures that can process sets of varying sizes efficiently and handle streaming data. The key idea is developing online learning algorithms for set functions, incrementally updating models without requiring retraining on the entire dataset. Additionally, exploring techniques to manage concept drift, where the data distribution evolves over time, is also important for maintaining sustained model performance in set-based data streams.   
    
    \item \textbf{Domain-specific enhancements}: To further exploit the potential of set function learning methods in processing set structured data, it is significant to tailor set function models for specific domains such as multi-object detection, document classification, and drug discovery. These adaptations should preserve permutation-invariance and consider the unique requirements of different data types. By incorporating domain knowledge and optimizing architectures to account for specific relational patterns or employing well-designed loss functions, these specialized models will outperform universal frameworks. 
    
    \item \textbf{Hybrid approaches}: Combining set function learning with other machine learning paradigms can significantly improve their applicability and performance. For instance, integrating set function learning with graph neural networks can enhance relational reasoning, while incorporating sequence models enables handling tasks involving both sequential and unordered data. Additionally, exploring the synergy between set function learning and reinforcement learning can unlock new possibilities for complex decision-making in set-based dynamic environments, such as resource allocation and planning.
    
    \item \textbf{Few-shot and transfer learning}: Exploring few-shot and transfer learning techniques is meaningful for improving the generalization ability of set function models with limited data and enabling effective knowledge transfer across related tasks.
    Meta-learning algorithms specifically designed for set functions can help adapt to new tasks with minimal examples, while transfer learning mechanisms leveraging pre-trained set function models can accelerate domain adaptation. Additionally, employing self-supervised learning that exploits inherent set structures can further enhance few-shot performance for set functions.
    
    \item \textbf{Graph prediction}: Extending the principles of set function learning to graph structures offers significant potential for predicting complex structures and relationships. Developing set-to-graph architectures that effectively transition from unordered sets to structured graph outputs can advance applications such as scene understanding and relationship inference. Moreover, adapting set function learning models for graph-based tasks can improve performance in domains requiring hierarchical reasoning, such as molecular property prediction or knowledge graph construction.
    
\end{itemize}

\begin{acks}
This project is supported in part by National Science Foundation under IIS-2144285 and IIS-2414308.
\end{acks}

\bibliographystyle{ACM-Reference-Format}
\bibliography{sample-base}


\begin{thebibliography}{153}


\ifx \showCODEN    \undefined \def \showCODEN     #1{\unskip}     \fi
\ifx \showDOI      \undefined \def \showDOI       #1{#1}\fi
\ifx \showISBNx    \undefined \def \showISBNx     #1{\unskip}     \fi
\ifx \showISBNxiii \undefined \def \showISBNxiii  #1{\unskip}     \fi
\ifx \showISSN     \undefined \def \showISSN      #1{\unskip}     \fi
\ifx \showLCCN     \undefined \def \showLCCN      #1{\unskip}     \fi
\ifx \shownote     \undefined \def \shownote      #1{#1}          \fi
\ifx \showarticletitle \undefined \def \showarticletitle #1{#1}   \fi
\ifx \showURL      \undefined \def \showURL       {\relax}        \fi
\providecommand\bibfield[2]{#2}
\providecommand\bibinfo[2]{#2}
\providecommand\natexlab[1]{#1}
\providecommand\showeprint[2][]{arXiv:#2}

\bibitem[Abedin et~al\mbox{.}(2019)]%
        {abedin2019sparsesense}
\bibfield{author}{\bibinfo{person}{Alireza Abedin}, \bibinfo{person}{S~Hamid Rezatofighi}, \bibinfo{person}{Qinfeng Shi}, {and} \bibinfo{person}{Damith~C Ranasinghe}.} \bibinfo{year}{2019}\natexlab{}.
\newblock \showarticletitle{SparseSense: human activity recognition from highly sparse sensor data-streams using set-based neural networks}. In \bibinfo{booktitle}{\emph{International Joint Conference on Artificial Intelligence 2019}}. Association for the Advancement of Artificial Intelligence (AAAI), \bibinfo{pages}{5780--5786}.
\newblock


\bibitem[Adomavicius and Tuzhilin(2010)]%
        {adomavicius2010context}
\bibfield{author}{\bibinfo{person}{Gediminas Adomavicius} {and} \bibinfo{person}{Alexander Tuzhilin}.} \bibinfo{year}{2010}\natexlab{}.
\newblock \showarticletitle{Context-aware recommender systems}.
\newblock In \bibinfo{booktitle}{\emph{Recommender systems handbook}}. \bibinfo{publisher}{Springer}, \bibinfo{pages}{217--253}.
\newblock


\bibitem[Aittala and Durand(2018)]%
        {aittala2018burst}
\bibfield{author}{\bibinfo{person}{Miika Aittala} {and} \bibinfo{person}{Fr{\'e}do Durand}.} \bibinfo{year}{2018}\natexlab{}.
\newblock \showarticletitle{Burst image deblurring using permutation invariant convolutional neural networks}. In \bibinfo{booktitle}{\emph{Proceedings of the European conference on computer vision (ECCV)}}. \bibinfo{pages}{731--747}.
\newblock


\bibitem[Arandjelovic et~al\mbox{.}(2016)]%
        {arandjelovic2016netvlad}
\bibfield{author}{\bibinfo{person}{Relja Arandjelovic}, \bibinfo{person}{Petr Gronat}, \bibinfo{person}{Akihiko Torii}, \bibinfo{person}{Tomas Pajdla}, {and} \bibinfo{person}{Josef Sivic}.} \bibinfo{year}{2016}\natexlab{}.
\newblock \showarticletitle{NetVLAD: CNN architecture for weakly supervised place recognition}. In \bibinfo{booktitle}{\emph{Proceedings of the IEEE conference on computer vision and pattern recognition}}. \bibinfo{pages}{5297--5307}.
\newblock


\bibitem[Armeni et~al\mbox{.}(2016)]%
        {armeni20163d}
\bibfield{author}{\bibinfo{person}{Iro Armeni}, \bibinfo{person}{Ozan Sener}, \bibinfo{person}{Amir~R Zamir}, \bibinfo{person}{Helen Jiang}, \bibinfo{person}{Ioannis Brilakis}, \bibinfo{person}{Martin Fischer}, {and} \bibinfo{person}{Silvio Savarese}.} \bibinfo{year}{2016}\natexlab{}.
\newblock \showarticletitle{3d semantic parsing of large-scale indoor spaces}. In \bibinfo{booktitle}{\emph{Proceedings of the IEEE conference on computer vision and pattern recognition}}. \bibinfo{pages}{1534--1543}.
\newblock


\bibitem[Arnold et~al\mbox{.}(2021)]%
        {arnold2021fast}
\bibfield{author}{\bibinfo{person}{Eduardo Arnold}, \bibinfo{person}{Sajjad Mozaffari}, {and} \bibinfo{person}{Mehrdad Dianati}.} \bibinfo{year}{2021}\natexlab{}.
\newblock \showarticletitle{Fast and robust registration of partially overlapping point clouds}.
\newblock \bibinfo{journal}{\emph{IEEE Robotics and Automation Letters}} \bibinfo{volume}{7}, \bibinfo{number}{2} (\bibinfo{year}{2021}), \bibinfo{pages}{1502--1509}.
\newblock


\bibitem[Arvaniti and Claassen(2017)]%
        {arvaniti2017sensitive}
\bibfield{author}{\bibinfo{person}{Eirini Arvaniti} {and} \bibinfo{person}{Manfred Claassen}.} \bibinfo{year}{2017}\natexlab{}.
\newblock \showarticletitle{Sensitive detection of rare disease-associated cell subsets via representation learning}.
\newblock \bibinfo{journal}{\emph{Nature communications}} \bibinfo{volume}{8}, \bibinfo{number}{1} (\bibinfo{year}{2017}), \bibinfo{pages}{14825}.
\newblock


\bibitem[Ba et~al\mbox{.}(2016)]%
        {ba2016layer}
\bibfield{author}{\bibinfo{person}{Jimmy~Lei Ba}, \bibinfo{person}{Jamie~Ryan Kiros}, {and} \bibinfo{person}{Geoffrey~E Hinton}.} \bibinfo{year}{2016}\natexlab{}.
\newblock \showarticletitle{Layer Normalization}.
\newblock \bibinfo{journal}{\emph{stat}}  \bibinfo{volume}{1050} (\bibinfo{year}{2016}), \bibinfo{pages}{21}.
\newblock


\bibitem[Bahdanau et~al\mbox{.}(2015)]%
        {bahdanau2015neural}
\bibfield{author}{\bibinfo{person}{Dzmitry Bahdanau}, \bibinfo{person}{Kyung~Hyun Cho}, {and} \bibinfo{person}{Yoshua Bengio}.} \bibinfo{year}{2015}\natexlab{}.
\newblock \showarticletitle{Neural machine translation by jointly learning to align and translate}. In \bibinfo{booktitle}{\emph{3rd International Conference on Learning Representations, ICLR 2015}}.
\newblock


\bibitem[Bai et~al\mbox{.}(2023)]%
        {bai2023permutation}
\bibfield{author}{\bibinfo{person}{Jun Bai}, \bibinfo{person}{Chuantao Yin}, \bibinfo{person}{Hanhua Hong}, \bibinfo{person}{Jianfei Zhang}, \bibinfo{person}{Chen Li}, \bibinfo{person}{Yanmeng Wang}, {and} \bibinfo{person}{Wenge Rong}.} \bibinfo{year}{2023}\natexlab{}.
\newblock \showarticletitle{Permutation Invariant Training for Paraphrase Identification}. In \bibinfo{booktitle}{\emph{ICASSP 2023-2023 IEEE International Conference on Acoustics, Speech and Signal Processing (ICASSP)}}. IEEE, \bibinfo{pages}{1--5}.
\newblock


\bibitem[Balcan and Harvey(2018)]%
        {balcan2018submodular}
\bibfield{author}{\bibinfo{person}{Maria-Florina Balcan} {and} \bibinfo{person}{Nicholas~JA Harvey}.} \bibinfo{year}{2018}\natexlab{}.
\newblock \showarticletitle{Submodular functions: Learnability, structure, and optimization}.
\newblock \bibinfo{journal}{\emph{SIAM J. Comput.}} \bibinfo{volume}{47}, \bibinfo{number}{3} (\bibinfo{year}{2018}), \bibinfo{pages}{703--754}.
\newblock


\bibitem[Bartunov et~al\mbox{.}(2022)]%
        {bartunov2022equilibrium}
\bibfield{author}{\bibinfo{person}{Sergey Bartunov}, \bibinfo{person}{Fabian~B Fuchs}, {and} \bibinfo{person}{Timothy~P Lillicrap}.} \bibinfo{year}{2022}\natexlab{}.
\newblock \showarticletitle{Equilibrium aggregation: encoding sets via optimization}. In \bibinfo{booktitle}{\emph{Uncertainty in Artificial Intelligence}}. PMLR, \bibinfo{pages}{139--149}.
\newblock


\bibitem[Bebis and Georgiopoulos(1994)]%
        {bebis1994feed}
\bibfield{author}{\bibinfo{person}{George Bebis} {and} \bibinfo{person}{Michael Georgiopoulos}.} \bibinfo{year}{1994}\natexlab{}.
\newblock \showarticletitle{Feed-forward neural networks}.
\newblock \bibinfo{journal}{\emph{Ieee Potentials}} \bibinfo{volume}{13}, \bibinfo{number}{4} (\bibinfo{year}{1994}), \bibinfo{pages}{27--31}.
\newblock


\bibitem[Blondel et~al\mbox{.}(2022)]%
        {blondel2022efficient}
\bibfield{author}{\bibinfo{person}{Mathieu Blondel}, \bibinfo{person}{Quentin Berthet}, \bibinfo{person}{Marco Cuturi}, \bibinfo{person}{Roy Frostig}, \bibinfo{person}{Stephan Hoyer}, \bibinfo{person}{Felipe Llinares-L{\'o}pez}, \bibinfo{person}{Fabian Pedregosa}, {and} \bibinfo{person}{Jean-Philippe Vert}.} \bibinfo{year}{2022}\natexlab{}.
\newblock \showarticletitle{Efficient and modular implicit differentiation}.
\newblock \bibinfo{journal}{\emph{Advances in neural information processing systems}}  \bibinfo{volume}{35} (\bibinfo{year}{2022}), \bibinfo{pages}{5230--5242}.
\newblock


\bibitem[Bronstein et~al\mbox{.}(2017)]%
        {bronstein2017geometric}
\bibfield{author}{\bibinfo{person}{Michael~M Bronstein}, \bibinfo{person}{Joan Bruna}, \bibinfo{person}{Yann LeCun}, \bibinfo{person}{Arthur Szlam}, {and} \bibinfo{person}{Pierre Vandergheynst}.} \bibinfo{year}{2017}\natexlab{}.
\newblock \showarticletitle{Geometric deep learning: going beyond euclidean data}.
\newblock \bibinfo{journal}{\emph{IEEE Signal Processing Magazine}} \bibinfo{volume}{34}, \bibinfo{number}{4} (\bibinfo{year}{2017}), \bibinfo{pages}{18--42}.
\newblock


\bibitem[Bruno et~al\mbox{.}(2021)]%
        {bruno2021mini}
\bibfield{author}{\bibinfo{person}{Andreis Bruno}, \bibinfo{person}{Jeffrey Willette}, \bibinfo{person}{Juho Lee}, {and} \bibinfo{person}{Sung~Ju Hwang}.} \bibinfo{year}{2021}\natexlab{}.
\newblock \showarticletitle{Mini-batch consistent slot set encoder for scalable set encoding}.
\newblock \bibinfo{journal}{\emph{Advances in Neural Information Processing Systems}}  \bibinfo{volume}{34} (\bibinfo{year}{2021}), \bibinfo{pages}{21365--21374}.
\newblock


\bibitem[Buathong et~al\mbox{.}(2020)]%
        {buathong2020kernels}
\bibfield{author}{\bibinfo{person}{Poompol Buathong}, \bibinfo{person}{David Ginsbourger}, {and} \bibinfo{person}{Tipaluck Krityakierne}.} \bibinfo{year}{2020}\natexlab{}.
\newblock \showarticletitle{Kernels over sets of finite sets using rkhs embeddings, with application to bayesian (combinatorial) optimization}. In \bibinfo{booktitle}{\emph{International Conference on Artificial Intelligence and Statistics}}. PMLR, \bibinfo{pages}{2731--2741}.
\newblock


\bibitem[Bueno and Hylton(2021)]%
        {bueno2021representation}
\bibfield{author}{\bibinfo{person}{Christian Bueno} {and} \bibinfo{person}{Alan Hylton}.} \bibinfo{year}{2021}\natexlab{}.
\newblock \showarticletitle{On the representation power of set pooling networks}.
\newblock \bibinfo{journal}{\emph{Advances in Neural Information Processing Systems}}  \bibinfo{volume}{34} (\bibinfo{year}{2021}), \bibinfo{pages}{17170--17182}.
\newblock


\bibitem[Burke(2017)]%
        {burke2017multisided}
\bibfield{author}{\bibinfo{person}{Robin Burke}.} \bibinfo{year}{2017}\natexlab{}.
\newblock \showarticletitle{Multisided fairness for recommendation}.
\newblock \bibinfo{journal}{\emph{arXiv preprint arXiv:1707.00093}} (\bibinfo{year}{2017}).
\newblock


\bibitem[Byrd et~al\mbox{.}(1995)]%
        {byrd1995limited}
\bibfield{author}{\bibinfo{person}{Richard~H Byrd}, \bibinfo{person}{Peihuang Lu}, \bibinfo{person}{Jorge Nocedal}, {and} \bibinfo{person}{Ciyou Zhu}.} \bibinfo{year}{1995}\natexlab{}.
\newblock \showarticletitle{A limited memory algorithm for bound constrained optimization}.
\newblock \bibinfo{journal}{\emph{SIAM Journal on scientific computing}} \bibinfo{volume}{16}, \bibinfo{number}{5} (\bibinfo{year}{1995}), \bibinfo{pages}{1190--1208}.
\newblock


\bibitem[Chao et~al\mbox{.}(2019)]%
        {chao2019gaitset}
\bibfield{author}{\bibinfo{person}{Hanqing Chao}, \bibinfo{person}{Yiwei He}, \bibinfo{person}{Junping Zhang}, {and} \bibinfo{person}{Jianfeng Feng}.} \bibinfo{year}{2019}\natexlab{}.
\newblock \showarticletitle{Gaitset: Regarding gait as a set for cross-view gait recognition}. In \bibinfo{booktitle}{\emph{Proceedings of the AAAI conference on artificial intelligence}}, Vol.~\bibinfo{volume}{33}. \bibinfo{address}{null}, \bibinfo{pages}{8126--8133}.
\newblock


\bibitem[Chechik et~al\mbox{.}(2010)]%
        {chechik2010large}
\bibfield{author}{\bibinfo{person}{Gal Chechik}, \bibinfo{person}{Varun Sharma}, \bibinfo{person}{Uri Shalit}, {and} \bibinfo{person}{Samy Bengio}.} \bibinfo{year}{2010}\natexlab{}.
\newblock \showarticletitle{Large Scale Online Learning of Image Similarity Through Ranking}.
\newblock \bibinfo{journal}{\emph{Journal of Machine Learning Research}}  \bibinfo{volume}{11} (\bibinfo{year}{2010}), \bibinfo{pages}{1109--1135}.
\newblock


\bibitem[Chen et~al\mbox{.}(2017)]%
        {chen2017deeplab}
\bibfield{author}{\bibinfo{person}{Liang-Chieh Chen}, \bibinfo{person}{George Papandreou}, \bibinfo{person}{Iasonas Kokkinos}, \bibinfo{person}{Kevin Murphy}, {and} \bibinfo{person}{Alan~L Yuille}.} \bibinfo{year}{2017}\natexlab{}.
\newblock \showarticletitle{Deeplab: Semantic image segmentation with deep convolutional nets, atrous convolution, and fully connected crfs}.
\newblock \bibinfo{journal}{\emph{IEEE transactions on pattern analysis and machine intelligence}} \bibinfo{volume}{40}, \bibinfo{number}{4} (\bibinfo{year}{2017}), \bibinfo{pages}{834--848}.
\newblock


\bibitem[dan Guo et~al\mbox{.}(2021)]%
        {dan2021learning}
\bibfield{author}{\bibinfo{person}{Dan dan Guo}, \bibinfo{person}{Long Tian}, \bibinfo{person}{Minghe Zhang}, \bibinfo{person}{Mingyuan Zhou}, {and} \bibinfo{person}{Hongyuan Zha}.} \bibinfo{year}{2021}\natexlab{}.
\newblock \showarticletitle{Learning Prototype-oriented Set Representations for Meta-Learning}. In \bibinfo{booktitle}{\emph{International Conference on Learning Representations}}.
\newblock


\bibitem[De and Chakrabarti(2022)]%
        {de2022neural}
\bibfield{author}{\bibinfo{person}{Abir De} {and} \bibinfo{person}{Soumen Chakrabarti}.} \bibinfo{year}{2022}\natexlab{}.
\newblock \showarticletitle{Neural Estimation of Submodular Functions with Applications to Differentiable Subset Selection}.
\newblock \bibinfo{journal}{\emph{Advances in Neural Information Processing Systems}}  \bibinfo{volume}{35} (\bibinfo{year}{2022}), \bibinfo{pages}{19537--19552}.
\newblock


\bibitem[Dennis et~al\mbox{.}(2019)]%
        {dennis2019shallow}
\bibfield{author}{\bibinfo{person}{Don Dennis}, \bibinfo{person}{Durmus Alp~Emre Acar}, \bibinfo{person}{Vikram Mandikal}, \bibinfo{person}{Vinu~Sankar Sadasivan}, \bibinfo{person}{Venkatesh Saligrama}, \bibinfo{person}{Harsha~Vardhan Simhadri}, {and} \bibinfo{person}{Prateek Jain}.} \bibinfo{year}{2019}\natexlab{}.
\newblock \showarticletitle{Shallow rnn: accurate time-series classification on resource constrained devices}.
\newblock \bibinfo{journal}{\emph{Advances in neural information processing systems}}  \bibinfo{volume}{32} (\bibinfo{year}{2019}).
\newblock


\bibitem[Doerr et~al\mbox{.}(2020)]%
        {doerr2020optimization}
\bibfield{author}{\bibinfo{person}{Benjamin Doerr}, \bibinfo{person}{Carola Doerr}, \bibinfo{person}{Aneta Neumann}, \bibinfo{person}{Frank Neumann}, {and} \bibinfo{person}{Andrew Sutton}.} \bibinfo{year}{2020}\natexlab{}.
\newblock \showarticletitle{Optimization of chance-constrained submodular functions}. In \bibinfo{booktitle}{\emph{Proceedings of the AAAI Conference on Artificial Intelligence}}, Vol.~\bibinfo{volume}{34}. \bibinfo{pages}{1460--1467}.
\newblock


\bibitem[Dolhansky and Bilmes(2016)]%
        {dolhansky2016deep}
\bibfield{author}{\bibinfo{person}{Brian~W Dolhansky} {and} \bibinfo{person}{Jeff~A Bilmes}.} \bibinfo{year}{2016}\natexlab{}.
\newblock \showarticletitle{Deep submodular functions: Definitions and learning}.
\newblock \bibinfo{journal}{\emph{Advances in Neural Information Processing Systems}}  \bibinfo{volume}{29} (\bibinfo{year}{2016}), \bibinfo{pages}{3404--3412}.
\newblock


\bibitem[Duvenaud et~al\mbox{.}(2015)]%
        {duvenaud2015convolutional}
\bibfield{author}{\bibinfo{person}{David~K Duvenaud}, \bibinfo{person}{Dougal Maclaurin}, \bibinfo{person}{Jorge Iparraguirre}, \bibinfo{person}{Rafael Bombarell}, \bibinfo{person}{Timothy Hirzel}, \bibinfo{person}{Al{\'a}n Aspuru-Guzik}, {and} \bibinfo{person}{Ryan~P Adams}.} \bibinfo{year}{2015}\natexlab{}.
\newblock \showarticletitle{Convolutional networks on graphs for learning molecular fingerprints}.
\newblock \bibinfo{journal}{\emph{Advances in neural information processing systems}}  \bibinfo{volume}{28} (\bibinfo{year}{2015}), \bibinfo{pages}{2224--2232}.
\newblock


\bibitem[Engelmann et~al\mbox{.}(2020)]%
        {engelmann20203d}
\bibfield{author}{\bibinfo{person}{Francis Engelmann}, \bibinfo{person}{Martin Bokeloh}, \bibinfo{person}{Alireza Fathi}, \bibinfo{person}{Bastian Leibe}, {and} \bibinfo{person}{Matthias Nie{\ss}ner}.} \bibinfo{year}{2020}\natexlab{}.
\newblock \showarticletitle{3d-mpa: Multi-proposal aggregation for 3d semantic instance segmentation}. In \bibinfo{booktitle}{\emph{Proceedings of the IEEE/CVF conference on computer vision and pattern recognition}}. \bibinfo{pages}{9031--9040}.
\newblock


\bibitem[Faber et~al\mbox{.}(2016)]%
        {faber2016machine}
\bibfield{author}{\bibinfo{person}{Felix~A Faber}, \bibinfo{person}{Alexander Lindmaa}, \bibinfo{person}{O~Anatole Von~Lilienfeld}, {and} \bibinfo{person}{Rickard Armiento}.} \bibinfo{year}{2016}\natexlab{}.
\newblock \showarticletitle{Machine learning energies of 2 million elpasolite (A B C 2 D 6) crystals}.
\newblock \bibinfo{journal}{\emph{Physical review letters}} \bibinfo{volume}{117}, \bibinfo{number}{13} (\bibinfo{year}{2016}), \bibinfo{pages}{135502}.
\newblock


\bibitem[Fei et~al\mbox{.}(2022)]%
        {fei2022dumlp}
\bibfield{author}{\bibinfo{person}{Jiajun Fei}, \bibinfo{person}{Ziyu Zhu}, \bibinfo{person}{Wenlei Liu}, \bibinfo{person}{Zhidong Deng}, \bibinfo{person}{Mingyang Li}, \bibinfo{person}{Huanjun Deng}, {and} \bibinfo{person}{Shuo Zhang}.} \bibinfo{year}{2022}\natexlab{}.
\newblock \showarticletitle{Dumlp-pin: a dual-mlp-dot-product permutation-invariant network for set feature extraction}. In \bibinfo{booktitle}{\emph{Proceedings of the AAAI conference on artificial intelligence}}, Vol.~\bibinfo{volume}{36}. \bibinfo{pages}{598--606}.
\newblock


\bibitem[Feldman et~al\mbox{.}(2013)]%
        {feldman2013representation}
\bibfield{author}{\bibinfo{person}{Vitaly Feldman}, \bibinfo{person}{Pravesh Kothari}, {and} \bibinfo{person}{Jan Vondr{\'a}k}.} \bibinfo{year}{2013}\natexlab{}.
\newblock \showarticletitle{Representation, approximation and learning of submodular functions using low-rank decision trees}. In \bibinfo{booktitle}{\emph{Conference on Learning Theory}}. PMLR, \bibinfo{pages}{711--740}.
\newblock


\bibitem[Gelly and Gauvain(2017)]%
        {gelly2017optimization}
\bibfield{author}{\bibinfo{person}{Gregory Gelly} {and} \bibinfo{person}{Jean-Luc Gauvain}.} \bibinfo{year}{2017}\natexlab{}.
\newblock \showarticletitle{Optimization of RNN-based speech activity detection}.
\newblock \bibinfo{journal}{\emph{IEEE/ACM Transactions on Audio, Speech, and Language Processing}} \bibinfo{volume}{26}, \bibinfo{number}{3} (\bibinfo{year}{2017}), \bibinfo{pages}{646--656}.
\newblock


\bibitem[Ghadimi and Beigy(2020)]%
        {ghadimi2020deep}
\bibfield{author}{\bibinfo{person}{Alireza Ghadimi} {and} \bibinfo{person}{Hamid Beigy}.} \bibinfo{year}{2020}\natexlab{}.
\newblock \showarticletitle{Deep submodular network: An application to multi-document summarization}.
\newblock \bibinfo{journal}{\emph{Expert Systems with Applications}}  \bibinfo{volume}{152} (\bibinfo{year}{2020}), \bibinfo{pages}{113392}.
\newblock


\bibitem[Gillenwater et~al\mbox{.}(2014)]%
        {gillenwater2014expectation}
\bibfield{author}{\bibinfo{person}{Jennifer~A Gillenwater}, \bibinfo{person}{Alex Kulesza}, \bibinfo{person}{Emily Fox}, {and} \bibinfo{person}{Ben Taskar}.} \bibinfo{year}{2014}\natexlab{}.
\newblock \showarticletitle{Expectation-maximization for learning determinantal point processes}.
\newblock \bibinfo{journal}{\emph{Advances in Neural Information Processing Systems}}  \bibinfo{volume}{27} (\bibinfo{year}{2014}).
\newblock


\bibitem[Gilmer et~al\mbox{.}(2017)]%
        {gilmer2017neural}
\bibfield{author}{\bibinfo{person}{Justin Gilmer}, \bibinfo{person}{Samuel~S Schoenholz}, \bibinfo{person}{Patrick~F Riley}, \bibinfo{person}{Oriol Vinyals}, {and} \bibinfo{person}{George~E Dahl}.} \bibinfo{year}{2017}\natexlab{}.
\newblock \showarticletitle{Neural message passing for quantum chemistry}. In \bibinfo{booktitle}{\emph{International conference on machine learning}}. PMLR, \bibinfo{pages}{1263--1272}.
\newblock


\bibitem[Gim et~al\mbox{.}(2021)]%
        {gim2021recipebowl}
\bibfield{author}{\bibinfo{person}{Mogan Gim}, \bibinfo{person}{Donghyeon Park}, \bibinfo{person}{Michael Spranger}, \bibinfo{person}{Kana Maruyama}, {and} \bibinfo{person}{Jaewoo Kang}.} \bibinfo{year}{2021}\natexlab{}.
\newblock \showarticletitle{Recipebowl: A cooking recommender for ingredients and recipes using set transformer}.
\newblock \bibinfo{journal}{\emph{IEEE Access}}  \bibinfo{volume}{9} (\bibinfo{year}{2021}), \bibinfo{pages}{143623--143633}.
\newblock


\bibitem[Girgis et~al\mbox{.}(2021)]%
        {girgis2021latent}
\bibfield{author}{\bibinfo{person}{Roger Girgis}, \bibinfo{person}{Florian Golemo}, \bibinfo{person}{Felipe Codevilla}, \bibinfo{person}{Martin Weiss}, \bibinfo{person}{Jim~Aldon D'Souza}, \bibinfo{person}{Samira~Ebrahimi Kahou}, \bibinfo{person}{Felix Heide}, {and} \bibinfo{person}{Christopher Pal}.} \bibinfo{year}{2021}\natexlab{}.
\newblock \showarticletitle{Latent Variable Sequential Set Transformers for Joint Multi-Agent Motion Prediction}. In \bibinfo{booktitle}{\emph{International Conference on Learning Representations}}.
\newblock


\bibitem[Hackel et~al\mbox{.}(2017)]%
        {hackel2017semantic3d}
\bibfield{author}{\bibinfo{person}{T Hackel}, \bibinfo{person}{N Savinov}, \bibinfo{person}{L Ladicky}, \bibinfo{person}{JD Wegner}, \bibinfo{person}{K Schindler}, {and} \bibinfo{person}{M Pollefeys}.} \bibinfo{year}{2017}\natexlab{}.
\newblock \showarticletitle{SEMANTIC3D. NET: A NEW LARGE-SCALE POINT CLOUD CLASSIFICATION BENCHMARK}.
\newblock \bibinfo{journal}{\emph{ISPRS Annals of the Photogrammetry, Remote Sensing and Spatial Information Sciences}}  \bibinfo{volume}{4} (\bibinfo{year}{2017}), \bibinfo{pages}{91--98}.
\newblock


\bibitem[Hartford et~al\mbox{.}(2018)]%
        {hartford2018deep}
\bibfield{author}{\bibinfo{person}{Jason Hartford}, \bibinfo{person}{Devon Graham}, \bibinfo{person}{Kevin Leyton-Brown}, {and} \bibinfo{person}{Siamak Ravanbakhsh}.} \bibinfo{year}{2018}\natexlab{}.
\newblock \showarticletitle{Deep models of interactions across sets}. In \bibinfo{booktitle}{\emph{International Conference on Machine Learning}}. PMLR, \bibinfo{pages}{1909--1918}.
\newblock


\bibitem[He et~al\mbox{.}(2016)]%
        {he2016deep}
\bibfield{author}{\bibinfo{person}{Kaiming He}, \bibinfo{person}{Xiangyu Zhang}, \bibinfo{person}{Shaoqing Ren}, {and} \bibinfo{person}{Jian Sun}.} \bibinfo{year}{2016}\natexlab{}.
\newblock \showarticletitle{Deep residual learning for image recognition}. In \bibinfo{booktitle}{\emph{Proceedings of the IEEE conference on computer vision and pattern recognition}}. \bibinfo{pages}{770--778}.
\newblock


\bibitem[Horn et~al\mbox{.}(2020)]%
        {horn2020set}
\bibfield{author}{\bibinfo{person}{Max Horn}, \bibinfo{person}{Michael Moor}, \bibinfo{person}{Christian Bock}, \bibinfo{person}{Bastian Rieck}, {and} \bibinfo{person}{Karsten Borgwardt}.} \bibinfo{year}{2020}\natexlab{}.
\newblock \showarticletitle{Set functions for time series}. In \bibinfo{booktitle}{\emph{International Conference on Machine Learning}}. PMLR, \bibinfo{pages}{4353--4363}.
\newblock


\bibitem[Hornung et~al\mbox{.}(2013)]%
        {hornung2013octomap}
\bibfield{author}{\bibinfo{person}{Armin Hornung}, \bibinfo{person}{Kai~M Wurm}, \bibinfo{person}{Maren Bennewitz}, \bibinfo{person}{Cyrill Stachniss}, {and} \bibinfo{person}{Wolfram Burgard}.} \bibinfo{year}{2013}\natexlab{}.
\newblock \showarticletitle{OctoMap: An efficient probabilistic 3D mapping framework based on octrees}.
\newblock \bibinfo{journal}{\emph{Autonomous robots}}  \bibinfo{volume}{34} (\bibinfo{year}{2013}), \bibinfo{pages}{189--206}.
\newblock


\bibitem[Hu et~al\mbox{.}(2020)]%
        {hu2020open}
\bibfield{author}{\bibinfo{person}{Weihua Hu}, \bibinfo{person}{Matthias Fey}, \bibinfo{person}{Marinka Zitnik}, \bibinfo{person}{Yuxiao Dong}, \bibinfo{person}{Hongyu Ren}, \bibinfo{person}{Bowen Liu}, \bibinfo{person}{Michele Catasta}, {and} \bibinfo{person}{Jure Leskovec}.} \bibinfo{year}{2020}\natexlab{}.
\newblock \showarticletitle{Open graph benchmark: Datasets for machine learning on graphs}.
\newblock \bibinfo{journal}{\emph{Advances in neural information processing systems}}  \bibinfo{volume}{33} (\bibinfo{year}{2020}), \bibinfo{pages}{22118--22133}.
\newblock


\bibitem[Hu et~al\mbox{.}(2019)]%
        {hu2019robust}
\bibfield{author}{\bibinfo{person}{Zicheng Hu}, \bibinfo{person}{Benjamin~S Glicksberg}, {and} \bibinfo{person}{Atul~J Butte}.} \bibinfo{year}{2019}\natexlab{}.
\newblock \showarticletitle{Robust prediction of clinical outcomes using cytometry data}.
\newblock \bibinfo{journal}{\emph{Bioinformatics}} \bibinfo{volume}{35}, \bibinfo{number}{7} (\bibinfo{year}{2019}), \bibinfo{pages}{1197--1203}.
\newblock


\bibitem[Hua et~al\mbox{.}(2018)]%
        {hua2018pointwise}
\bibfield{author}{\bibinfo{person}{Binh-Son Hua}, \bibinfo{person}{Minh-Khoi Tran}, {and} \bibinfo{person}{Sai-Kit Yeung}.} \bibinfo{year}{2018}\natexlab{}.
\newblock \showarticletitle{Pointwise convolutional neural networks}. In \bibinfo{booktitle}{\emph{Proceedings of the IEEE conference on computer vision and pattern recognition}}. \bibinfo{pages}{984--993}.
\newblock


\bibitem[Huang et~al\mbox{.}(2000)]%
        {huang2000classification}
\bibfield{author}{\bibinfo{person}{Guang-Bin Huang}, \bibinfo{person}{Yan-Qiu Chen}, {and} \bibinfo{person}{Haroon~A Babri}.} \bibinfo{year}{2000}\natexlab{}.
\newblock \showarticletitle{Classification ability of single hidden layer feedforward neural networks}.
\newblock \bibinfo{journal}{\emph{IEEE transactions on neural networks}} \bibinfo{volume}{11}, \bibinfo{number}{3} (\bibinfo{year}{2000}), \bibinfo{pages}{799--801}.
\newblock


\bibitem[Iyer et~al\mbox{.}(2021)]%
        {iyer2021submodular}
\bibfield{author}{\bibinfo{person}{Rishabh Iyer}, \bibinfo{person}{Ninad Khargoankar}, \bibinfo{person}{Jeff Bilmes}, {and} \bibinfo{person}{Himanshu Asanani}.} \bibinfo{year}{2021}\natexlab{}.
\newblock \showarticletitle{Submodular combinatorial information measures with applications in machine learning}. In \bibinfo{booktitle}{\emph{Algorithmic Learning Theory}}. PMLR, \bibinfo{pages}{722--754}.
\newblock


\bibitem[Jha et~al\mbox{.}(2022)]%
        {jha2022neural}
\bibfield{author}{\bibinfo{person}{Saurav Jha}, \bibinfo{person}{Dong Gong}, \bibinfo{person}{Xuesong Wang}, \bibinfo{person}{Richard~E Turner}, {and} \bibinfo{person}{Lina Yao}.} \bibinfo{year}{2022}\natexlab{}.
\newblock \showarticletitle{The neural process family: Survey, applications and perspectives}.
\newblock \bibinfo{journal}{\emph{arXiv preprint arXiv:2209.00517}} (\bibinfo{year}{2022}).
\newblock


\bibitem[Jiang et~al\mbox{.}(2023)]%
        {jiang2023pharmacophoric}
\bibfield{author}{\bibinfo{person}{Yinghui Jiang}, \bibinfo{person}{Shuting Jin}, \bibinfo{person}{Xurui Jin}, \bibinfo{person}{Xianglu Xiao}, \bibinfo{person}{Wenfan Wu}, \bibinfo{person}{Xiangrong Liu}, \bibinfo{person}{Qiang Zhang}, \bibinfo{person}{Xiangxiang Zeng}, \bibinfo{person}{Guang Yang}, {and} \bibinfo{person}{Zhangming Niu}.} \bibinfo{year}{2023}\natexlab{}.
\newblock \showarticletitle{Pharmacophoric-constrained heterogeneous graph transformer model for molecular property prediction}.
\newblock \bibinfo{journal}{\emph{Communications Chemistry}} \bibinfo{volume}{6}, \bibinfo{number}{1} (\bibinfo{year}{2023}), \bibinfo{pages}{60}.
\newblock


\bibitem[Johnson et~al\mbox{.}(2017)]%
        {johnson2017clevr}
\bibfield{author}{\bibinfo{person}{Justin Johnson}, \bibinfo{person}{Bharath Hariharan}, \bibinfo{person}{Laurens Van Der~Maaten}, \bibinfo{person}{Li Fei-Fei}, \bibinfo{person}{C Lawrence~Zitnick}, {and} \bibinfo{person}{Ross Girshick}.} \bibinfo{year}{2017}\natexlab{}.
\newblock \showarticletitle{Clevr: A diagnostic dataset for compositional language and elementary visual reasoning}. In \bibinfo{booktitle}{\emph{Proceedings of the IEEE conference on computer vision and pattern recognition}}. \bibinfo{pages}{2901--2910}.
\newblock


\bibitem[Jumper et~al\mbox{.}(2021)]%
        {jumper2021highly}
\bibfield{author}{\bibinfo{person}{John Jumper}, \bibinfo{person}{Richard Evans}, \bibinfo{person}{Alexander Pritzel}, \bibinfo{person}{Tim Green}, \bibinfo{person}{Michael Figurnov}, \bibinfo{person}{Olaf Ronneberger}, \bibinfo{person}{Kathryn Tunyasuvunakool}, \bibinfo{person}{Russ Bates}, \bibinfo{person}{Augustin {\v{Z}}{\'\i}dek}, \bibinfo{person}{Anna Potapenko}, {et~al\mbox{.}}} \bibinfo{year}{2021}\natexlab{}.
\newblock \showarticletitle{Highly accurate protein structure prediction with AlphaFold}.
\newblock \bibinfo{journal}{\emph{Nature}} \bibinfo{volume}{596}, \bibinfo{number}{7873} (\bibinfo{year}{2021}), \bibinfo{pages}{583--589}.
\newblock


\bibitem[Jung et~al\mbox{.}(2015)]%
        {jung2015exploration}
\bibfield{author}{\bibinfo{person}{In-Soo Jung}, \bibinfo{person}{Mario Berges}, \bibinfo{person}{James~H Garrett~Jr}, {and} \bibinfo{person}{Barnabas Poczos}.} \bibinfo{year}{2015}\natexlab{}.
\newblock \showarticletitle{Exploration and evaluation of AR, MPCA and KL anomaly detection techniques to embankment dam piezometer data}.
\newblock \bibinfo{journal}{\emph{Advanced Engineering Informatics}} \bibinfo{volume}{29}, \bibinfo{number}{4} (\bibinfo{year}{2015}), \bibinfo{pages}{902--917}.
\newblock


\bibitem[Jurewicz and Derczynski(2022)]%
        {jurewicz2022set}
\bibfield{author}{\bibinfo{person}{Mateusz Jurewicz} {and} \bibinfo{person}{Leon Derczynski}.} \bibinfo{year}{2022}\natexlab{}.
\newblock \showarticletitle{Set Interdependence Transformer: Set-to-Sequence Neural Networks for Permutation Learning and Structure Prediction}. In \bibinfo{booktitle}{\emph{Proceedings on the International Joint Conferences on Artificial Intelligence}}.
\newblock


\bibitem[Kimura et~al\mbox{.}(2023)]%
        {kimura2023shift15m}
\bibfield{author}{\bibinfo{person}{Masanari Kimura}, \bibinfo{person}{Takuma Nakamura}, {and} \bibinfo{person}{Yuki Saito}.} \bibinfo{year}{2023}\natexlab{}.
\newblock \showarticletitle{SHIFT15M: Fashion-specific dataset for set-to-set matching with several distribution shifts}. In \bibinfo{booktitle}{\emph{Proceedings of the IEEE/CVF Conference on Computer Vision and Pattern Recognition}}. \bibinfo{pages}{3508--3513}.
\newblock


\bibitem[Kimura et~al\mbox{.}(2024)]%
        {kimura2024permutation}
\bibfield{author}{\bibinfo{person}{Masanari Kimura}, \bibinfo{person}{Ryotaro Shimizu}, \bibinfo{person}{Yuki Hirakawa}, \bibinfo{person}{Ryosuke Goto}, {and} \bibinfo{person}{Yuki Saito}.} \bibinfo{year}{2024}\natexlab{}.
\newblock \showarticletitle{On permutation-invariant neural networks}.
\newblock \bibinfo{journal}{\emph{arXiv preprint arXiv:2403.17410}} (\bibinfo{year}{2024}).
\newblock


\bibitem[Kosiorek et~al\mbox{.}(2020)]%
        {kosiorek2020conditional}
\bibfield{author}{\bibinfo{person}{Adam~R Kosiorek}, \bibinfo{person}{Hyunjik Kim}, {and} \bibinfo{person}{Danilo~J Rezende}.} \bibinfo{year}{2020}\natexlab{}.
\newblock \showarticletitle{Conditional set generation with transformers}.
\newblock \bibinfo{journal}{\emph{arXiv preprint arXiv:2006.16841}} (\bibinfo{year}{2020}), \bibinfo{pages}{arXiv--2006}.
\newblock


\bibitem[Krizhevsky et~al\mbox{.}(2012)]%
        {krizhevsky2012imagenet}
\bibfield{author}{\bibinfo{person}{Alex Krizhevsky}, \bibinfo{person}{Ilya Sutskever}, {and} \bibinfo{person}{Geoffrey~E Hinton}.} \bibinfo{year}{2012}\natexlab{}.
\newblock \showarticletitle{Imagenet classification with deep convolutional neural networks}.
\newblock \bibinfo{journal}{\emph{Advances in neural information processing systems}}  \bibinfo{volume}{25} (\bibinfo{year}{2012}).
\newblock


\bibitem[Kunaver and Po{\v{z}}rl(2017)]%
        {kunaver2017diversity}
\bibfield{author}{\bibinfo{person}{Matev{\v{z}} Kunaver} {and} \bibinfo{person}{Toma{\v{z}} Po{\v{z}}rl}.} \bibinfo{year}{2017}\natexlab{}.
\newblock \showarticletitle{Diversity in recommender systems--A survey}.
\newblock \bibinfo{journal}{\emph{Knowledge-based systems}}  \bibinfo{volume}{123} (\bibinfo{year}{2017}), \bibinfo{pages}{154--162}.
\newblock


\bibitem[Lang et~al\mbox{.}(2019)]%
        {lang2019pointpillars}
\bibfield{author}{\bibinfo{person}{Alex~H Lang}, \bibinfo{person}{Sourabh Vora}, \bibinfo{person}{Holger Caesar}, \bibinfo{person}{Lubing Zhou}, \bibinfo{person}{Jiong Yang}, {and} \bibinfo{person}{Oscar Beijbom}.} \bibinfo{year}{2019}\natexlab{}.
\newblock \showarticletitle{Pointpillars: Fast encoders for object detection from point clouds}. In \bibinfo{booktitle}{\emph{Proceedings of the IEEE/CVF conference on computer vision and pattern recognition}}. \bibinfo{pages}{12697--12705}.
\newblock


\bibitem[LeCun et~al\mbox{.}(2015)]%
        {lecun2015deep}
\bibfield{author}{\bibinfo{person}{Yann LeCun}, \bibinfo{person}{Yoshua Bengio}, {and} \bibinfo{person}{Geoffrey Hinton}.} \bibinfo{year}{2015}\natexlab{}.
\newblock \showarticletitle{Deep learning}.
\newblock \bibinfo{journal}{\emph{nature}} \bibinfo{volume}{521}, \bibinfo{number}{7553} (\bibinfo{year}{2015}), \bibinfo{pages}{436--444}.
\newblock


\bibitem[LeCun et~al\mbox{.}(1998)]%
        {lecun1998gradient}
\bibfield{author}{\bibinfo{person}{Yann LeCun}, \bibinfo{person}{L{\'e}on Bottou}, \bibinfo{person}{Yoshua Bengio}, {and} \bibinfo{person}{Patrick Haffner}.} \bibinfo{year}{1998}\natexlab{}.
\newblock \showarticletitle{Gradient-based learning applied to document recognition}.
\newblock \bibinfo{journal}{\emph{Proc. IEEE}} \bibinfo{volume}{86}, \bibinfo{number}{11} (\bibinfo{year}{1998}), \bibinfo{pages}{2278--2324}.
\newblock


\bibitem[Lee et~al\mbox{.}(2019b)]%
        {lee2019set}
\bibfield{author}{\bibinfo{person}{Juho Lee}, \bibinfo{person}{Yoonho Lee}, \bibinfo{person}{Jungtaek Kim}, \bibinfo{person}{Adam Kosiorek}, \bibinfo{person}{Seungjin Choi}, {and} \bibinfo{person}{Yee~Whye Teh}.} \bibinfo{year}{2019}\natexlab{b}.
\newblock \showarticletitle{Set transformer: A framework for attention-based permutation-invariant neural networks}. In \bibinfo{booktitle}{\emph{International conference on machine learning}}. PMLR, \bibinfo{pages}{3744--3753}.
\newblock


\bibitem[Lee et~al\mbox{.}(2019a)]%
        {lee2019deep}
\bibfield{author}{\bibinfo{person}{Juho Lee}, \bibinfo{person}{Yoonho Lee}, {and} \bibinfo{person}{Yee~Whye Teh}.} \bibinfo{year}{2019}\natexlab{a}.
\newblock \showarticletitle{Deep amortized clustering}.
\newblock \bibinfo{journal}{\emph{arXiv preprint arXiv:1909.13433}} (\bibinfo{year}{2019}).
\newblock


\bibitem[Lee et~al\mbox{.}(2022)]%
        {lee2022set}
\bibfield{author}{\bibinfo{person}{Seanie Lee}, \bibinfo{person}{Bruno Andreis}, \bibinfo{person}{Kenji Kawaguchi}, \bibinfo{person}{Juho Lee}, {and} \bibinfo{person}{Sung~Ju Hwang}.} \bibinfo{year}{2022}\natexlab{}.
\newblock \showarticletitle{Set-based meta-interpolation for few-task meta-learning}.
\newblock \bibinfo{journal}{\emph{Advances in Neural Information Processing Systems}}  \bibinfo{volume}{35} (\bibinfo{year}{2022}), \bibinfo{pages}{6775--6788}.
\newblock


\bibitem[Li and Zaki(2020)]%
        {li2020reciptor}
\bibfield{author}{\bibinfo{person}{Diya Li} {and} \bibinfo{person}{Mohammed~J Zaki}.} \bibinfo{year}{2020}\natexlab{}.
\newblock \showarticletitle{Reciptor: An effective pretrained model for recipe representation learning}. In \bibinfo{booktitle}{\emph{Proceedings of the 26th ACM SIGKDD international conference on knowledge discovery \& data mining}}. \bibinfo{pages}{1719--1727}.
\newblock


\bibitem[Li et~al\mbox{.}(2023)]%
        {li2023set}
\bibfield{author}{\bibinfo{person}{Jiangnan Li}, \bibinfo{person}{Yice Zhang}, \bibinfo{person}{Bin Liang}, \bibinfo{person}{Kam-Fai Wong}, {and} \bibinfo{person}{Ruifeng Xu}.} \bibinfo{year}{2023}\natexlab{}.
\newblock \showarticletitle{Set learning for generative information extraction}. In \bibinfo{booktitle}{\emph{Proceedings of the 2023 Conference on Empirical Methods in Natural Language Processing}}. \bibinfo{pages}{13043--13052}.
\newblock


\bibitem[Li et~al\mbox{.}(2020)]%
        {li2020exchangeable}
\bibfield{author}{\bibinfo{person}{Yang Li}, \bibinfo{person}{Haidong Yi}, \bibinfo{person}{Christopher Bender}, \bibinfo{person}{Siyuan Shan}, {and} \bibinfo{person}{Junier~B Oliva}.} \bibinfo{year}{2020}\natexlab{}.
\newblock \showarticletitle{Exchangeable neural ode for set modeling}.
\newblock \bibinfo{journal}{\emph{Advances in Neural Information Processing Systems}}  \bibinfo{volume}{33} (\bibinfo{year}{2020}), \bibinfo{pages}{6936--6946}.
\newblock


\bibitem[Lin et~al\mbox{.}(2014)]%
        {lin2014microsoft}
\bibfield{author}{\bibinfo{person}{Tsung-Yi Lin}, \bibinfo{person}{Michael Maire}, \bibinfo{person}{Serge Belongie}, \bibinfo{person}{James Hays}, \bibinfo{person}{Pietro Perona}, \bibinfo{person}{Deva Ramanan}, \bibinfo{person}{Piotr Doll{\'a}r}, {and} \bibinfo{person}{C~Lawrence Zitnick}.} \bibinfo{year}{2014}\natexlab{}.
\newblock \showarticletitle{Microsoft coco: Common objects in context}. In \bibinfo{booktitle}{\emph{Computer Vision--ECCV 2014: 13th European Conference, Zurich, Switzerland, September 6-12, 2014, Proceedings, Part V 13}}. Springer, \bibinfo{pages}{740--755}.
\newblock


\bibitem[Liu and Lapata(2019)]%
        {liu-lapata-2019-hierarchical}
\bibfield{author}{\bibinfo{person}{Yang Liu} {and} \bibinfo{person}{Mirella Lapata}.} \bibinfo{year}{2019}\natexlab{}.
\newblock \showarticletitle{Hierarchical Transformers for Multi-Document Summarization}. In \bibinfo{booktitle}{\emph{Proceedings of the 57th Annual Meeting of the Association for Computational Linguistics}}. \bibinfo{publisher}{Association for Computational Linguistics}, \bibinfo{address}{Florence, Italy}, \bibinfo{pages}{5070--5081}.
\newblock


\bibitem[Liu et~al\mbox{.}(2015a)]%
        {liu2015pdb}
\bibfield{author}{\bibinfo{person}{Zhihai Liu}, \bibinfo{person}{Yan Li}, \bibinfo{person}{Li Han}, \bibinfo{person}{Jie Li}, \bibinfo{person}{Jie Liu}, \bibinfo{person}{Zhixiong Zhao}, \bibinfo{person}{Wei Nie}, \bibinfo{person}{Yuchen Liu}, {and} \bibinfo{person}{Renxiao Wang}.} \bibinfo{year}{2015}\natexlab{a}.
\newblock \showarticletitle{PDB-wide collection of binding data: current status of the PDBbind database}.
\newblock \bibinfo{journal}{\emph{Bioinformatics}} \bibinfo{volume}{31}, \bibinfo{number}{3} (\bibinfo{year}{2015}), \bibinfo{pages}{405--412}.
\newblock


\bibitem[Liu et~al\mbox{.}(2015b)]%
        {liu2015deep}
\bibfield{author}{\bibinfo{person}{Ziwei Liu}, \bibinfo{person}{Ping Luo}, \bibinfo{person}{Xiaogang Wang}, {and} \bibinfo{person}{Xiaoou Tang}.} \bibinfo{year}{2015}\natexlab{b}.
\newblock \showarticletitle{Deep learning face attributes in the wild}. In \bibinfo{booktitle}{\emph{Proceedings of the IEEE international conference on computer vision}}. \bibinfo{pages}{3730--3738}.
\newblock


\bibitem[Lu(2016)]%
        {lu2016practical}
\bibfield{author}{\bibinfo{person}{Yi Lu}.} \bibinfo{year}{2016}\natexlab{}.
\newblock \showarticletitle{Practical tera-scale Walsh-Hadamard transform}. In \bibinfo{booktitle}{\emph{2016 Future Technologies Conference (FTC)}}. IEEE, \bibinfo{pages}{1230--1236}.
\newblock


\bibitem[Lu et~al\mbox{.}(2024)]%
        {lu2024slosh}
\bibfield{author}{\bibinfo{person}{Yuzhe Lu}, \bibinfo{person}{Xinran Liu}, \bibinfo{person}{Andrea Soltoggio}, {and} \bibinfo{person}{Soheil Kolouri}.} \bibinfo{year}{2024}\natexlab{}.
\newblock \showarticletitle{Slosh: Set locality sensitive hashing via sliced-wasserstein embeddings}. In \bibinfo{booktitle}{\emph{Proceedings of the IEEE/CVF Winter Conference on Applications of Computer Vision}}. \bibinfo{pages}{2566--2576}.
\newblock


\bibitem[Manupriya et~al\mbox{.}(2022)]%
        {manupriya2022improving}
\bibfield{author}{\bibinfo{person}{Piyushi Manupriya}, \bibinfo{person}{Tarun~Ram Menta}, \bibinfo{person}{Sakethanath~N Jagarlapudi}, {and} \bibinfo{person}{Vineeth~N Balasubramanian}.} \bibinfo{year}{2022}\natexlab{}.
\newblock \showarticletitle{Improving attribution methods by learning submodular functions}. In \bibinfo{booktitle}{\emph{International Conference on Artificial Intelligence and Statistics}}. PMLR, \bibinfo{pages}{2173--2190}.
\newblock


\bibitem[Mar{\i}n et~al\mbox{.}(2021)]%
        {marin2021recipe1m+}
\bibfield{author}{\bibinfo{person}{Javier Mar{\i}n}, \bibinfo{person}{Aritro Biswas}, \bibinfo{person}{Ferda Ofli}, \bibinfo{person}{Nicholas Hynes}, \bibinfo{person}{Amaia Salvador}, \bibinfo{person}{Yusuf Aytar}, \bibinfo{person}{Ingmar Weber}, {and} \bibinfo{person}{Antonio Torralba}.} \bibinfo{year}{2021}\natexlab{}.
\newblock \showarticletitle{Recipe1m+: A dataset for learning cross-modal embeddings for cooking recipes and food images}.
\newblock \bibinfo{journal}{\emph{IEEE Transactions on Pattern Analysis and Machine Intelligence}} \bibinfo{volume}{43}, \bibinfo{number}{1} (\bibinfo{year}{2021}), \bibinfo{pages}{187--203}.
\newblock


\bibitem[Maron et~al\mbox{.}(2019a)]%
        {maron2019provably}
\bibfield{author}{\bibinfo{person}{Haggai Maron}, \bibinfo{person}{Heli Ben-Hamu}, \bibinfo{person}{Hadar Serviansky}, {and} \bibinfo{person}{Yaron Lipman}.} \bibinfo{year}{2019}\natexlab{a}.
\newblock \showarticletitle{Provably powerful graph networks}.
\newblock \bibinfo{journal}{\emph{Advances in neural information processing systems}}  \bibinfo{volume}{32} (\bibinfo{year}{2019}).
\newblock


\bibitem[Maron et~al\mbox{.}(2019b)]%
        {maron2019universality}
\bibfield{author}{\bibinfo{person}{Haggai Maron}, \bibinfo{person}{Ethan Fetaya}, \bibinfo{person}{Nimrod Segol}, {and} \bibinfo{person}{Yaron Lipman}.} \bibinfo{year}{2019}\natexlab{b}.
\newblock \showarticletitle{On the universality of invariant networks}. In \bibinfo{booktitle}{\emph{International conference on machine learning}}. PMLR, \bibinfo{pages}{4363--4371}.
\newblock


\bibitem[Maron et~al\mbox{.}(2020)]%
        {maron2020learning}
\bibfield{author}{\bibinfo{person}{Haggai Maron}, \bibinfo{person}{Or Litany}, \bibinfo{person}{Gal Chechik}, {and} \bibinfo{person}{Ethan Fetaya}.} \bibinfo{year}{2020}\natexlab{}.
\newblock \showarticletitle{On learning sets of symmetric elements}. In \bibinfo{booktitle}{\emph{International conference on machine learning}}. PMLR, \bibinfo{pages}{6734--6744}.
\newblock


\bibitem[Milioto et~al\mbox{.}(2019)]%
        {milioto2019rangenet++}
\bibfield{author}{\bibinfo{person}{Andres Milioto}, \bibinfo{person}{Ignacio Vizzo}, \bibinfo{person}{Jens Behley}, {and} \bibinfo{person}{Cyrill Stachniss}.} \bibinfo{year}{2019}\natexlab{}.
\newblock \showarticletitle{Rangenet++: Fast and accurate lidar semantic segmentation}. In \bibinfo{booktitle}{\emph{2019 IEEE/RSJ international conference on intelligent robots and systems (IROS)}}. IEEE, \bibinfo{pages}{4213--4220}.
\newblock


\bibitem[M{\"u}ller et~al\mbox{.}(2022)]%
        {mullertransformers}
\bibfield{author}{\bibinfo{person}{Samuel M{\"u}ller}, \bibinfo{person}{Noah Hollmann}, \bibinfo{person}{Sebastian~Pineda Arango}, \bibinfo{person}{Josif Grabocka}, {and} \bibinfo{person}{Frank Hutter}.} \bibinfo{year}{2022}\natexlab{}.
\newblock \showarticletitle{Transformers Can Do Bayesian Inference}. In \bibinfo{booktitle}{\emph{International Conference on Learning Representations}}.
\newblock


\bibitem[Murphy et~al\mbox{.}(2018)]%
        {murphy2018janossy}
\bibfield{author}{\bibinfo{person}{Ryan~L Murphy}, \bibinfo{person}{Balasubramaniam Srinivasan}, \bibinfo{person}{Vinayak Rao}, {and} \bibinfo{person}{Bruno Ribeiro}.} \bibinfo{year}{2018}\natexlab{}.
\newblock \showarticletitle{Janossy pooling: Learning deep permutation-invariant functions for variable-size inputs}.
\newblock \bibinfo{journal}{\emph{arXiv preprint arXiv:1811.01900}} (\bibinfo{year}{2018}).
\newblock


\bibitem[Nikolentzos et~al\mbox{.}(2017)]%
        {nikolentzos2017matching}
\bibfield{author}{\bibinfo{person}{Giannis Nikolentzos}, \bibinfo{person}{Polykarpos Meladianos}, {and} \bibinfo{person}{Michalis Vazirgiannis}.} \bibinfo{year}{2017}\natexlab{}.
\newblock \showarticletitle{Matching node embeddings for graph similarity}. In \bibinfo{booktitle}{\emph{Proceedings of the AAAI conference on Artificial Intelligence}}, Vol.~\bibinfo{volume}{31}.
\newblock


\bibitem[Oh et~al\mbox{.}(2019)]%
        {oh2019speech2face}
\bibfield{author}{\bibinfo{person}{Tae-Hyun Oh}, \bibinfo{person}{Tali Dekel}, \bibinfo{person}{Changil Kim}, \bibinfo{person}{Inbar Mosseri}, \bibinfo{person}{William~T Freeman}, \bibinfo{person}{Michael Rubinstein}, {and} \bibinfo{person}{Wojciech Matusik}.} \bibinfo{year}{2019}\natexlab{}.
\newblock \showarticletitle{Speech2face: Learning the face behind a voice}. In \bibinfo{booktitle}{\emph{Proceedings of the IEEE/CVF conference on computer vision and pattern recognition}}. \bibinfo{pages}{7539--7548}.
\newblock


\bibitem[Ou et~al\mbox{.}(2022)]%
        {ou2022learning}
\bibfield{author}{\bibinfo{person}{Zijing Ou}, \bibinfo{person}{Tingyang Xu}, \bibinfo{person}{Qinliang Su}, \bibinfo{person}{Yingzhen Li}, \bibinfo{person}{Peilin Zhao}, {and} \bibinfo{person}{Yatao Bian}.} \bibinfo{year}{2022}\natexlab{}.
\newblock \showarticletitle{Learning neural set functions under the optimal subset oracle}.
\newblock \bibinfo{journal}{\emph{Advances in Neural Information Processing Systems}}  \bibinfo{volume}{35} (\bibinfo{year}{2022}), \bibinfo{pages}{35021--35034}.
\newblock


\bibitem[Pakman et~al\mbox{.}(2020)]%
        {pakman2020neural}
\bibfield{author}{\bibinfo{person}{Ari Pakman}, \bibinfo{person}{Yueqi Wang}, \bibinfo{person}{Catalin Mitelut}, \bibinfo{person}{JinHyung Lee}, {and} \bibinfo{person}{Liam Paninski}.} \bibinfo{year}{2020}\natexlab{}.
\newblock \showarticletitle{Neural clustering processes}. In \bibinfo{booktitle}{\emph{International Conference on Machine Learning}}. PMLR, \bibinfo{pages}{7455--7465}.
\newblock


\bibitem[Pandey and Wang(2022)]%
        {pandey2022self}
\bibfield{author}{\bibinfo{person}{Ashutosh Pandey} {and} \bibinfo{person}{DeLiang Wang}.} \bibinfo{year}{2022}\natexlab{}.
\newblock \showarticletitle{Self-attending RNN for speech enhancement to improve cross-corpus generalization}.
\newblock \bibinfo{journal}{\emph{IEEE/ACM Transactions on Audio, Speech, and Language Processing}}  \bibinfo{volume}{30} (\bibinfo{year}{2022}), \bibinfo{pages}{1374--1385}.
\newblock


\bibitem[Pascanu et~al\mbox{.}(2013)]%
        {pascanu2013difficulty}
\bibfield{author}{\bibinfo{person}{Razvan Pascanu}, \bibinfo{person}{Tomas Mikolov}, {and} \bibinfo{person}{Yoshua Bengio}.} \bibinfo{year}{2013}\natexlab{}.
\newblock \showarticletitle{On the difficulty of training recurrent neural networks}. In \bibinfo{booktitle}{\emph{International conference on machine learning}}. Pmlr, \bibinfo{pages}{1310--1318}.
\newblock


\bibitem[Philbin et~al\mbox{.}(2007)]%
        {philbin2007object}
\bibfield{author}{\bibinfo{person}{James Philbin}, \bibinfo{person}{Ondrej Chum}, \bibinfo{person}{Michael Isard}, \bibinfo{person}{Josef Sivic}, {and} \bibinfo{person}{Andrew Zisserman}.} \bibinfo{year}{2007}\natexlab{}.
\newblock \showarticletitle{Object retrieval with large vocabularies and fast spatial matching}. In \bibinfo{booktitle}{\emph{2007 IEEE conference on computer vision and pattern recognition}}. IEEE, \bibinfo{pages}{1--8}.
\newblock


\bibitem[Poornima and Paramasivan(2020)]%
        {poornima2020anomaly}
\bibfield{author}{\bibinfo{person}{I~Gethzi~Ahila Poornima} {and} \bibinfo{person}{B Paramasivan}.} \bibinfo{year}{2020}\natexlab{}.
\newblock \showarticletitle{Anomaly detection in wireless sensor network using machine learning algorithm}.
\newblock \bibinfo{journal}{\emph{Computer communications}}  \bibinfo{volume}{151} (\bibinfo{year}{2020}), \bibinfo{pages}{331--337}.
\newblock


\bibitem[Prokudin et~al\mbox{.}(2019)]%
        {prokudin2019efficient}
\bibfield{author}{\bibinfo{person}{Sergey Prokudin}, \bibinfo{person}{Christoph Lassner}, {and} \bibinfo{person}{Javier Romero}.} \bibinfo{year}{2019}\natexlab{}.
\newblock \showarticletitle{Efficient learning on point clouds with basis point sets}. In \bibinfo{booktitle}{\emph{Proceedings of the IEEE/CVF international conference on computer vision}}. \bibinfo{pages}{4332--4341}.
\newblock


\bibitem[P{\"u}schel(2018)]%
        {puschel2018discrete}
\bibfield{author}{\bibinfo{person}{Markus P{\"u}schel}.} \bibinfo{year}{2018}\natexlab{}.
\newblock \showarticletitle{A discrete signal processing framework for set functions}. In \bibinfo{booktitle}{\emph{2018 IEEE International Conference on Acoustics, Speech and Signal Processing (ICASSP)}}. IEEE, \bibinfo{pages}{4359--4363}.
\newblock


\bibitem[P{\"u}schel and Wendler(2020)]%
        {puschel2020discrete}
\bibfield{author}{\bibinfo{person}{Markus P{\"u}schel} {and} \bibinfo{person}{Chris Wendler}.} \bibinfo{year}{2020}\natexlab{}.
\newblock \showarticletitle{Discrete signal processing with set functions}.
\newblock \bibinfo{journal}{\emph{IEEE Transactions on Signal Processing}}  \bibinfo{volume}{69} (\bibinfo{year}{2020}), \bibinfo{pages}{1039--1053}.
\newblock


\bibitem[Qi et~al\mbox{.}(2017a)]%
        {qi2017pointnet}
\bibfield{author}{\bibinfo{person}{Charles~R Qi}, \bibinfo{person}{Hao Su}, \bibinfo{person}{Kaichun Mo}, {and} \bibinfo{person}{Leonidas~J Guibas}.} \bibinfo{year}{2017}\natexlab{a}.
\newblock \showarticletitle{Pointnet: Deep learning on point sets for 3d classification and segmentation}. In \bibinfo{booktitle}{\emph{Proceedings of the IEEE conference on computer vision and pattern recognition}}. \bibinfo{pages}{652--660}.
\newblock


\bibitem[Qi et~al\mbox{.}(2017b)]%
        {qi2017pointnet++}
\bibfield{author}{\bibinfo{person}{Charles~Ruizhongtai Qi}, \bibinfo{person}{Li Yi}, \bibinfo{person}{Hao Su}, {and} \bibinfo{person}{Leonidas~J Guibas}.} \bibinfo{year}{2017}\natexlab{b}.
\newblock \showarticletitle{Pointnet++: Deep hierarchical feature learning on point sets in a metric space}.
\newblock \bibinfo{journal}{\emph{Advances in neural information processing systems}}  \bibinfo{volume}{30} (\bibinfo{year}{2017}).
\newblock


\bibitem[Qin et~al\mbox{.}(2019)]%
        {qin2019adapting}
\bibfield{author}{\bibinfo{person}{Kechen Qin}, \bibinfo{person}{Cheng Li}, \bibinfo{person}{Virgil Pavlu}, {and} \bibinfo{person}{Javed Aslam}.} \bibinfo{year}{2019}\natexlab{}.
\newblock \showarticletitle{Adapting RNN Sequence Prediction Model to Multi-label Set Prediction}. In \bibinfo{booktitle}{\emph{Proceedings of the 2019 Conference of the North American Chapter of the Association for Computational Linguistics: Human Language Technologies, Volume 1 (Long and Short Papers)}}. \bibinfo{pages}{3181--3190}.
\newblock


\bibitem[Rafiey and Yoshida(2020)]%
        {rafiey2020fast}
\bibfield{author}{\bibinfo{person}{Akbar Rafiey} {and} \bibinfo{person}{Yuichi Yoshida}.} \bibinfo{year}{2020}\natexlab{}.
\newblock \showarticletitle{Fast and Private Submodular and $ k $-Submodular Functions Maximization with Matroid Constraints}. In \bibinfo{booktitle}{\emph{International conference on machine learning}}. PMLR, \bibinfo{pages}{7887--7897}.
\newblock


\bibitem[Rajput and Verma(2014)]%
        {rajput2014back}
\bibfield{author}{\bibinfo{person}{Neelima Rajput} {and} \bibinfo{person}{SK Verma}.} \bibinfo{year}{2014}\natexlab{}.
\newblock \showarticletitle{Back propagation feed forward neural network approach for speech recognition}. In \bibinfo{booktitle}{\emph{Proceedings of 3rd international conference on reliability, infocom technologies and optimization}}. IEEE, \bibinfo{pages}{1--6}.
\newblock


\bibitem[Raskhodnikova and Yaroslavtsev(2013)]%
        {raskhodnikova2013learning}
\bibfield{author}{\bibinfo{person}{Sofya Raskhodnikova} {and} \bibinfo{person}{Grigory Yaroslavtsev}.} \bibinfo{year}{2013}\natexlab{}.
\newblock \showarticletitle{Learning pseudo-boolean k-dnf and submodular functions}. In \bibinfo{booktitle}{\emph{Proceedings of the Twenty-Fourth Annual ACM-SIAM Symposium on Discrete Algorithms}}. SIAM, \bibinfo{pages}{1356--1368}.
\newblock


\bibitem[Redmon et~al\mbox{.}(2016)]%
        {redmon2016you}
\bibfield{author}{\bibinfo{person}{Joseph Redmon}, \bibinfo{person}{Santosh Divvala}, \bibinfo{person}{Ross Girshick}, {and} \bibinfo{person}{Ali Farhadi}.} \bibinfo{year}{2016}\natexlab{}.
\newblock \showarticletitle{You only look once: Unified, real-time object detection}. In \bibinfo{booktitle}{\emph{Proceedings of the IEEE conference on computer vision and pattern recognition}}. \bibinfo{pages}{779--788}.
\newblock


\bibitem[Ren et~al\mbox{.}(2016)]%
        {ren2016faster}
\bibfield{author}{\bibinfo{person}{Shaoqing Ren}, \bibinfo{person}{Kaiming He}, \bibinfo{person}{Ross Girshick}, {and} \bibinfo{person}{Jian Sun}.} \bibinfo{year}{2016}\natexlab{}.
\newblock \showarticletitle{Faster R-CNN: Towards real-time object detection with region proposal networks}.
\newblock \bibinfo{journal}{\emph{IEEE transactions on pattern analysis and machine intelligence}} \bibinfo{volume}{39}, \bibinfo{number}{6} (\bibinfo{year}{2016}), \bibinfo{pages}{1137--1149}.
\newblock


\bibitem[Rezatofighi et~al\mbox{.}(2021)]%
        {rezatofighi2021learn}
\bibfield{author}{\bibinfo{person}{Hamid Rezatofighi}, \bibinfo{person}{Tianyu Zhu}, \bibinfo{person}{Roman Kaskman}, \bibinfo{person}{Farbod~T Motlagh}, \bibinfo{person}{Javen~Qinfeng Shi}, \bibinfo{person}{Anton Milan}, \bibinfo{person}{Daniel Cremers}, \bibinfo{person}{Laura Leal-Taix{\'e}}, {and} \bibinfo{person}{Ian Reid}.} \bibinfo{year}{2021}\natexlab{}.
\newblock \showarticletitle{Learn to predict sets using feed-forward neural networks}.
\newblock \bibinfo{journal}{\emph{IEEE Transactions on Pattern Analysis and Machine Intelligence}} \bibinfo{volume}{44}, \bibinfo{number}{12} (\bibinfo{year}{2021}), \bibinfo{pages}{9011--9025}.
\newblock


\bibitem[Rezatofighi et~al\mbox{.}(2017)]%
        {rezatofighi2017deepsetnet}
\bibfield{author}{\bibinfo{person}{S~Hamid Rezatofighi}, \bibinfo{person}{Vijay~Kumar Bg}, \bibinfo{person}{Anton Milan}, \bibinfo{person}{Ehsan Abbasnejad}, \bibinfo{person}{Anthony Dick}, {and} \bibinfo{person}{Ian Reid}.} \bibinfo{year}{2017}\natexlab{}.
\newblock \showarticletitle{Deepsetnet: Predicting sets with deep neural networks}. In \bibinfo{booktitle}{\emph{2017 IEEE International Conference on Computer Vision (ICCV)}}. IEEE, \bibinfo{pages}{5257--5266}.
\newblock


\bibitem[Ridnik et~al\mbox{.}(2021)]%
        {ridnik2021asymmetric}
\bibfield{author}{\bibinfo{person}{Tal Ridnik}, \bibinfo{person}{Emanuel Ben-Baruch}, \bibinfo{person}{Nadav Zamir}, \bibinfo{person}{Asaf Noy}, \bibinfo{person}{Itamar Friedman}, \bibinfo{person}{Matan Protter}, {and} \bibinfo{person}{Lihi Zelnik-Manor}.} \bibinfo{year}{2021}\natexlab{}.
\newblock \showarticletitle{Asymmetric loss for multi-label classification}. In \bibinfo{booktitle}{\emph{Proceedings of the IEEE/CVF international conference on computer vision}}. \bibinfo{pages}{82--91}.
\newblock


\bibitem[Ronneberger et~al\mbox{.}(2015)]%
        {ronneberger2015u}
\bibfield{author}{\bibinfo{person}{Olaf Ronneberger}, \bibinfo{person}{Philipp Fischer}, {and} \bibinfo{person}{Thomas Brox}.} \bibinfo{year}{2015}\natexlab{}.
\newblock \showarticletitle{U-net: Convolutional networks for biomedical image segmentation}. In \bibinfo{booktitle}{\emph{Medical image computing and computer-assisted intervention--MICCAI 2015: 18th international conference, Munich, Germany, October 5-9, 2015, proceedings, part III 18}}. Springer, \bibinfo{pages}{234--241}.
\newblock


\bibitem[Salas-Moreno et~al\mbox{.}(2013)]%
        {salas2013slam++}
\bibfield{author}{\bibinfo{person}{Renato~F Salas-Moreno}, \bibinfo{person}{Richard~A Newcombe}, \bibinfo{person}{Hauke Strasdat}, \bibinfo{person}{Paul~HJ Kelly}, {and} \bibinfo{person}{Andrew~J Davison}.} \bibinfo{year}{2013}\natexlab{}.
\newblock \showarticletitle{Slam++: Simultaneous localisation and mapping at the level of objects}. In \bibinfo{booktitle}{\emph{Proceedings of the IEEE conference on computer vision and pattern recognition}}. \bibinfo{pages}{1352--1359}.
\newblock


\bibitem[Sbrana et~al\mbox{.}(2020)]%
        {sbrana2020n}
\bibfield{author}{\bibinfo{person}{Attilio Sbrana}, \bibinfo{person}{Andr{\'e} Luis~Debiaso Rossi}, {and} \bibinfo{person}{Murilo~Coelho Naldi}.} \bibinfo{year}{2020}\natexlab{}.
\newblock \showarticletitle{N-BEATS-RNN: deep learning for time series forecasting}. In \bibinfo{booktitle}{\emph{2020 19th IEEE International Conference on Machine Learning and Applications (ICMLA)}}. IEEE, \bibinfo{pages}{765--768}.
\newblock


\bibitem[Scheibler et~al\mbox{.}(2015)]%
        {scheibler2015fast}
\bibfield{author}{\bibinfo{person}{Robin Scheibler}, \bibinfo{person}{Saeid Haghighatshoar}, {and} \bibinfo{person}{Martin Vetterli}.} \bibinfo{year}{2015}\natexlab{}.
\newblock \showarticletitle{A fast Hadamard transform for signals with sublinear sparsity in the transform domain}.
\newblock \bibinfo{journal}{\emph{IEEE Transactions on Information Theory}} \bibinfo{volume}{61}, \bibinfo{number}{4} (\bibinfo{year}{2015}), \bibinfo{pages}{2115--2132}.
\newblock


\bibitem[Sedhain et~al\mbox{.}(2015)]%
        {sedhain2015autorec}
\bibfield{author}{\bibinfo{person}{Suvash Sedhain}, \bibinfo{person}{Aditya~Krishna Menon}, \bibinfo{person}{Scott Sanner}, {and} \bibinfo{person}{Lexing Xie}.} \bibinfo{year}{2015}\natexlab{}.
\newblock \showarticletitle{Autorec: Autoencoders meet collaborative filtering}. In \bibinfo{booktitle}{\emph{Proceedings of the 24th international conference on World Wide Web}}. \bibinfo{pages}{111--112}.
\newblock


\bibitem[Sharma et~al\mbox{.}(2020)]%
        {sharma2020adaptation}
\bibfield{author}{\bibinfo{person}{Eva Sharma}, \bibinfo{person}{Guoli Ye}, \bibinfo{person}{Wenning Wei}, \bibinfo{person}{Rui Zhao}, \bibinfo{person}{Yao Tian}, \bibinfo{person}{Jian Wu}, \bibinfo{person}{Lei He}, \bibinfo{person}{Ed Lin}, {and} \bibinfo{person}{Yifan Gong}.} \bibinfo{year}{2020}\natexlab{}.
\newblock \showarticletitle{Adaptation of rnn transducer with text-to-speech technology for keyword spotting}. In \bibinfo{booktitle}{\emph{ICASSP 2020-2020 IEEE International Conference on Acoustics, Speech and Signal Processing (ICASSP)}}. IEEE, \bibinfo{pages}{7484--7488}.
\newblock


\bibitem[Shi et~al\mbox{.}(2020)]%
        {shi2020deep}
\bibfield{author}{\bibinfo{person}{Yifeng Shi}, \bibinfo{person}{Junier Oliva}, {and} \bibinfo{person}{Marc Niethammer}.} \bibinfo{year}{2020}\natexlab{}.
\newblock \showarticletitle{Deep message passing on sets}. In \bibinfo{booktitle}{\emph{Proceedings of the AAAI Conference on Artificial Intelligence}}, Vol.~\bibinfo{volume}{34}. \bibinfo{pages}{5750--5757}.
\newblock


\bibitem[Skianis et~al\mbox{.}(2020)]%
        {skianis2020rep}
\bibfield{author}{\bibinfo{person}{Konstantinos Skianis}, \bibinfo{person}{Giannis Nikolentzos}, \bibinfo{person}{Stratis Limnios}, {and} \bibinfo{person}{Michalis Vazirgiannis}.} \bibinfo{year}{2020}\natexlab{}.
\newblock \showarticletitle{Rep the set: Neural networks for learning set representations}. In \bibinfo{booktitle}{\emph{International conference on artificial intelligence and statistics}}. PMLR, \bibinfo{pages}{1410--1420}.
\newblock


\bibitem[Sun et~al\mbox{.}(2014)]%
        {sun2014deep}
\bibfield{author}{\bibinfo{person}{Yi Sun}, \bibinfo{person}{Xiaogang Wang}, {and} \bibinfo{person}{Xiaoou Tang}.} \bibinfo{year}{2014}\natexlab{}.
\newblock \showarticletitle{Deep learning face representation from predicting 10,000 classes}. In \bibinfo{booktitle}{\emph{Proceedings of the IEEE conference on computer vision and pattern recognition}}. \bibinfo{pages}{1891--1898}.
\newblock


\bibitem[Sutskever et~al\mbox{.}(2014)]%
        {sutskever2014sequence}
\bibfield{author}{\bibinfo{person}{Ilya Sutskever}, \bibinfo{person}{Oriol Vinyals}, {and} \bibinfo{person}{Quoc~V Le}.} \bibinfo{year}{2014}\natexlab{}.
\newblock \showarticletitle{Sequence to sequence learning with neural networks}.
\newblock \bibinfo{journal}{\emph{Advances in neural information processing systems}}  \bibinfo{volume}{27} (\bibinfo{year}{2014}).
\newblock


\bibitem[Tan and Le(2019)]%
        {tan2019efficientnet}
\bibfield{author}{\bibinfo{person}{Mingxing Tan} {and} \bibinfo{person}{Quoc Le}.} \bibinfo{year}{2019}\natexlab{}.
\newblock \showarticletitle{Efficientnet: Rethinking model scaling for convolutional neural networks}. In \bibinfo{booktitle}{\emph{International conference on machine learning}}. PMLR, \bibinfo{pages}{6105--6114}.
\newblock


\bibitem[Tan et~al\mbox{.}(2024)]%
        {tan2024naturalspeech}
\bibfield{author}{\bibinfo{person}{Xu Tan}, \bibinfo{person}{Jiawei Chen}, \bibinfo{person}{Haohe Liu}, \bibinfo{person}{Jian Cong}, \bibinfo{person}{Chen Zhang}, \bibinfo{person}{Yanqing Liu}, \bibinfo{person}{Xi Wang}, \bibinfo{person}{Yichong Leng}, \bibinfo{person}{Yuanhao Yi}, \bibinfo{person}{Lei He}, {et~al\mbox{.}}} \bibinfo{year}{2024}\natexlab{}.
\newblock \showarticletitle{Naturalspeech: End-to-end text-to-speech synthesis with human-level quality}.
\newblock \bibinfo{journal}{\emph{IEEE Transactions on Pattern Analysis and Machine Intelligence}} (\bibinfo{year}{2024}).
\newblock


\bibitem[Thomas et~al\mbox{.}(2022)]%
        {thomas2022integrating}
\bibfield{author}{\bibinfo{person}{Samuel Thomas}, \bibinfo{person}{Brian Kingsbury}, \bibinfo{person}{George Saon}, {and} \bibinfo{person}{Hong-Kwang~J Kuo}.} \bibinfo{year}{2022}\natexlab{}.
\newblock \showarticletitle{Integrating text inputs for training and adapting rnn transducer asr models}. In \bibinfo{booktitle}{\emph{ICASSP 2022-2022 IEEE International Conference on Acoustics, Speech and Signal Processing (ICASSP)}}. IEEE, \bibinfo{pages}{8127--8131}.
\newblock


\bibitem[Van Den~Oord et~al\mbox{.}(2016)]%
        {van2016pixel}
\bibfield{author}{\bibinfo{person}{A{\"a}ron Van Den~Oord}, \bibinfo{person}{Nal Kalchbrenner}, {and} \bibinfo{person}{Koray Kavukcuoglu}.} \bibinfo{year}{2016}\natexlab{}.
\newblock \showarticletitle{Pixel recurrent neural networks}. In \bibinfo{booktitle}{\emph{International conference on machine learning}}. PMLR, \bibinfo{pages}{1747--1756}.
\newblock


\bibitem[Vaswani et~al\mbox{.}(2017)]%
        {vaswani2017attention}
\bibfield{author}{\bibinfo{person}{Ashish Vaswani}, \bibinfo{person}{Noam Shazeer}, \bibinfo{person}{Niki Parmar}, \bibinfo{person}{Jakob Uszkoreit}, \bibinfo{person}{Llion Jones}, \bibinfo{person}{Aidan~N Gomez}, \bibinfo{person}{{\L}ukasz Kaiser}, {and} \bibinfo{person}{Illia Polosukhin}.} \bibinfo{year}{2017}\natexlab{}.
\newblock \showarticletitle{Attention is all you need}.
\newblock \bibinfo{journal}{\emph{Advances in neural information processing systems}}  \bibinfo{volume}{30} (\bibinfo{year}{2017}).
\newblock


\bibitem[Vinyals et~al\mbox{.}(2016)]%
        {vinyals2015order}
\bibfield{author}{\bibinfo{person}{Oriol Vinyals}, \bibinfo{person}{Samy Bengio}, {and} \bibinfo{person}{Manjunath Kudlur}.} \bibinfo{year}{2016}\natexlab{}.
\newblock \showarticletitle{Order matters: Sequence to sequence for sets}.
\newblock  (\bibinfo{year}{2016}).
\newblock


\bibitem[Wagstaff et~al\mbox{.}(2019)]%
        {wagstaff2019limitations}
\bibfield{author}{\bibinfo{person}{Edward Wagstaff}, \bibinfo{person}{Fabian Fuchs}, \bibinfo{person}{Martin Engelcke}, \bibinfo{person}{Ingmar Posner}, {and} \bibinfo{person}{Michael~A Osborne}.} \bibinfo{year}{2019}\natexlab{}.
\newblock \showarticletitle{On the limitations of representing functions on sets}. In \bibinfo{booktitle}{\emph{International Conference on Machine Learning}}. PMLR, \bibinfo{pages}{6487--6494}.
\newblock


\bibitem[Wagstaff et~al\mbox{.}(2022)]%
        {wagstaff2022universal}
\bibfield{author}{\bibinfo{person}{Edward Wagstaff}, \bibinfo{person}{Fabian~B Fuchs}, \bibinfo{person}{Martin Engelcke}, \bibinfo{person}{Michael~A Osborne}, {and} \bibinfo{person}{Ingmar Posner}.} \bibinfo{year}{2022}\natexlab{}.
\newblock \showarticletitle{Universal approximation of functions on sets}.
\newblock \bibinfo{journal}{\emph{Journal of Machine Learning Research}} \bibinfo{volume}{23}, \bibinfo{number}{151} (\bibinfo{year}{2022}), \bibinfo{pages}{1--56}.
\newblock


\bibitem[Wang et~al\mbox{.}(2023a)]%
        {wang2023deep}
\bibfield{author}{\bibinfo{person}{Dingrong Wang}, \bibinfo{person}{Deep~Shankar Pandey}, \bibinfo{person}{Krishna~Prasad Neupane}, \bibinfo{person}{Zhiwei Yu}, \bibinfo{person}{Ervine Zheng}, \bibinfo{person}{Zhi Zheng}, {and} \bibinfo{person}{Qi Yu}.} \bibinfo{year}{2023}\natexlab{a}.
\newblock \showarticletitle{Deep temporal sets with evidential reinforced attentions for unique behavioral pattern discovery}. In \bibinfo{booktitle}{\emph{International Conference on Machine Learning}}. PMLR, \bibinfo{pages}{36205--36223}.
\newblock


\bibitem[Wang et~al\mbox{.}(2023b)]%
        {wang2023polynomial}
\bibfield{author}{\bibinfo{person}{Peihao Wang}, \bibinfo{person}{Shenghao Yang}, \bibinfo{person}{Shu Li}, \bibinfo{person}{Zhangyang Wang}, {and} \bibinfo{person}{Pan Li}.} \bibinfo{year}{2023}\natexlab{b}.
\newblock \showarticletitle{Polynomial Width is Sufficient for Set Representation with High-dimensional Features}. In \bibinfo{booktitle}{\emph{The Twelfth International Conference on Learning Representations}}.
\newblock


\bibitem[Wang et~al\mbox{.}(2020)]%
        {wang2020amortized}
\bibfield{author}{\bibinfo{person}{Yueqi Wang}, \bibinfo{person}{Yoonho Lee}, \bibinfo{person}{Pallab Basu}, \bibinfo{person}{Juho Lee}, \bibinfo{person}{Yee~Whye Teh}, \bibinfo{person}{Liam Paninski}, {and} \bibinfo{person}{Ari Pakman}.} \bibinfo{year}{2020}\natexlab{}.
\newblock \showarticletitle{Amortized probabilistic detection of communities in graphs}.
\newblock \bibinfo{journal}{\emph{arXiv preprint arXiv:2010.15727}} (\bibinfo{year}{2020}).
\newblock


\bibitem[Wang et~al\mbox{.}(2019)]%
        {wang2019dynamic}
\bibfield{author}{\bibinfo{person}{Yue Wang}, \bibinfo{person}{Yongbin Sun}, \bibinfo{person}{Ziwei Liu}, \bibinfo{person}{Sanjay~E Sarma}, \bibinfo{person}{Michael~M Bronstein}, {and} \bibinfo{person}{Justin~M Solomon}.} \bibinfo{year}{2019}\natexlab{}.
\newblock \showarticletitle{Dynamic graph cnn for learning on point clouds}.
\newblock \bibinfo{journal}{\emph{ACM Transactions on Graphics (tog)}} \bibinfo{volume}{38}, \bibinfo{number}{5} (\bibinfo{year}{2019}), \bibinfo{pages}{1--12}.
\newblock


\bibitem[Wei et~al\mbox{.}(2014)]%
        {wei2014unsupervised}
\bibfield{author}{\bibinfo{person}{Kai Wei}, \bibinfo{person}{Yuzong Liu}, \bibinfo{person}{Katrin Kirchhoff}, {and} \bibinfo{person}{Jeff Bilmes}.} \bibinfo{year}{2014}\natexlab{}.
\newblock \showarticletitle{Unsupervised submodular subset selection for speech data}. In \bibinfo{booktitle}{\emph{2014 IEEE International Conference on Acoustics, Speech and Signal Processing (ICASSP)}}. IEEE, \bibinfo{pages}{4107--4111}.
\newblock


\bibitem[Wendler et~al\mbox{.}(2021)]%
        {wendler2021learning}
\bibfield{author}{\bibinfo{person}{Chris Wendler}, \bibinfo{person}{Andisheh Amrollahi}, \bibinfo{person}{Bastian Seifert}, \bibinfo{person}{Andreas Krause}, {and} \bibinfo{person}{Markus P{\"u}schel}.} \bibinfo{year}{2021}\natexlab{}.
\newblock \showarticletitle{Learning set functions that are sparse in non-orthogonal Fourier bases}. In \bibinfo{booktitle}{\emph{Proceedings of the AAAI Conference on Artificial Intelligence}}, Vol.~\bibinfo{volume}{35}. \bibinfo{pages}{10283--10292}.
\newblock


\bibitem[Wendler et~al\mbox{.}(2019)]%
        {wendler2019powerset}
\bibfield{author}{\bibinfo{person}{Chris Wendler}, \bibinfo{person}{Markus P{\"u}schel}, {and} \bibinfo{person}{Dan Alistarh}.} \bibinfo{year}{2019}\natexlab{}.
\newblock \showarticletitle{Powerset convolutional neural networks}.
\newblock \bibinfo{journal}{\emph{Advances in Neural Information Processing Systems}}  \bibinfo{volume}{32} (\bibinfo{year}{2019}).
\newblock


\bibitem[Willette et~al\mbox{.}(2023)]%
        {willette2023scalable}
\bibfield{author}{\bibinfo{person}{Jeffrey Willette}, \bibinfo{person}{Seanie Lee}, \bibinfo{person}{Bruno Andreis}, \bibinfo{person}{Kenji Kawaguchi}, \bibinfo{person}{Juho Lee}, {and} \bibinfo{person}{Sung~Ju Hwang}.} \bibinfo{year}{2023}\natexlab{}.
\newblock \showarticletitle{Scalable set encoding with universal mini-batch consistency and unbiased full set gradient approximation}. In \bibinfo{booktitle}{\emph{International Conference on Machine Learning}}. PMLR, \bibinfo{pages}{37008--37041}.
\newblock


\bibitem[Wu et~al\mbox{.}(2015)]%
        {wu20153d}
\bibfield{author}{\bibinfo{person}{Zhirong Wu}, \bibinfo{person}{Shuran Song}, \bibinfo{person}{Aditya Khosla}, \bibinfo{person}{Fisher Yu}, \bibinfo{person}{Linguang Zhang}, \bibinfo{person}{Xiaoou Tang}, {and} \bibinfo{person}{Jianxiong Xiao}.} \bibinfo{year}{2015}\natexlab{}.
\newblock \showarticletitle{3d shapenets: A deep representation for volumetric shapes}. In \bibinfo{booktitle}{\emph{Proceedings of the IEEE conference on computer vision and pattern recognition}}. \bibinfo{pages}{1912--1920}.
\newblock


\bibitem[Xu et~al\mbox{.}(2018)]%
        {xu2018spidercnn}
\bibfield{author}{\bibinfo{person}{Yifan Xu}, \bibinfo{person}{Tianqi Fan}, \bibinfo{person}{Mingye Xu}, \bibinfo{person}{Long Zeng}, {and} \bibinfo{person}{Yu Qiao}.} \bibinfo{year}{2018}\natexlab{}.
\newblock \showarticletitle{Spidercnn: Deep learning on point sets with parameterized convolutional filters}. In \bibinfo{booktitle}{\emph{Proceedings of the European conference on computer vision (ECCV)}}. \bibinfo{pages}{87--102}.
\newblock


\bibitem[Yang et~al\mbox{.}(2018)]%
        {yang2018pixor}
\bibfield{author}{\bibinfo{person}{Bin Yang}, \bibinfo{person}{Wenjie Luo}, {and} \bibinfo{person}{Raquel Urtasun}.} \bibinfo{year}{2018}\natexlab{}.
\newblock \showarticletitle{Pixor: Real-time 3d object detection from point clouds}. In \bibinfo{booktitle}{\emph{Proceedings of the IEEE conference on Computer Vision and Pattern Recognition}}. \bibinfo{pages}{7652--7660}.
\newblock


\bibitem[Yazici et~al\mbox{.}(2020)]%
        {yazici2020orderless}
\bibfield{author}{\bibinfo{person}{Vacit~Oguz Yazici}, \bibinfo{person}{Abel Gonzalez-Garcia}, \bibinfo{person}{Arnau Ramisa}, \bibinfo{person}{Bartlomiej Twardowski}, {and} \bibinfo{person}{Joost van~de Weijer}.} \bibinfo{year}{2020}\natexlab{}.
\newblock \showarticletitle{Orderless recurrent models for multi-label classification}. In \bibinfo{booktitle}{\emph{Proceedings of the IEEE/CVF Conference on Computer Vision and Pattern Recognition}}. \bibinfo{pages}{13440--13449}.
\newblock


\bibitem[Yeh et~al\mbox{.}(2017)]%
        {yeh2017learning}
\bibfield{author}{\bibinfo{person}{Chih-Kuan Yeh}, \bibinfo{person}{Wei-Chieh Wu}, \bibinfo{person}{Wei-Jen Ko}, {and} \bibinfo{person}{Yu-Chiang~Frank Wang}.} \bibinfo{year}{2017}\natexlab{}.
\newblock \showarticletitle{Learning deep latent space for multi-label classification}. In \bibinfo{booktitle}{\emph{Proceedings of the AAAI conference on artificial intelligence}}, Vol.~\bibinfo{volume}{31}.
\newblock


\bibitem[Yi and Stanley(2021)]%
        {yi2021cytoset}
\bibfield{author}{\bibinfo{person}{Haidong Yi} {and} \bibinfo{person}{Natalie Stanley}.} \bibinfo{year}{2021}\natexlab{}.
\newblock \showarticletitle{CytoSet: Predicting clinical outcomes via set-modeling of cytometry data}. In \bibinfo{booktitle}{\emph{Proceedings of the 12th ACM Conference on Bioinformatics, Computational Biology, and Health Informatics}}. \bibinfo{pages}{1--8}.
\newblock


\bibitem[Yi et~al\mbox{.}(2016)]%
        {yi2016scalable}
\bibfield{author}{\bibinfo{person}{Li Yi}, \bibinfo{person}{Vladimir~G Kim}, \bibinfo{person}{Duygu Ceylan}, \bibinfo{person}{I-Chao Shen}, \bibinfo{person}{Mengyan Yan}, \bibinfo{person}{Hao Su}, \bibinfo{person}{Cewu Lu}, \bibinfo{person}{Qixing Huang}, \bibinfo{person}{Alla Sheffer}, {and} \bibinfo{person}{Leonidas Guibas}.} \bibinfo{year}{2016}\natexlab{}.
\newblock \showarticletitle{A scalable active framework for region annotation in 3d shape collections}.
\newblock \bibinfo{journal}{\emph{ACM Transactions on Graphics (ToG)}} \bibinfo{volume}{35}, \bibinfo{number}{6} (\bibinfo{year}{2016}), \bibinfo{pages}{1--12}.
\newblock


\bibitem[Yu et~al\mbox{.}(2023)]%
        {yu2023predicting}
\bibfield{author}{\bibinfo{person}{Le Yu}, \bibinfo{person}{Zihang Liu}, \bibinfo{person}{Tongyu Zhu}, \bibinfo{person}{Leilei Sun}, \bibinfo{person}{Bowen Du}, {and} \bibinfo{person}{Weifeng Lv}.} \bibinfo{year}{2023}\natexlab{}.
\newblock \showarticletitle{Predicting temporal sets with simplified fully connected networks}. In \bibinfo{booktitle}{\emph{Proceedings of the AAAI Conference on Artificial Intelligence}}, Vol.~\bibinfo{volume}{37}. \bibinfo{pages}{4835--4844}.
\newblock


\bibitem[Zaheer et~al\mbox{.}(2017)]%
        {zaheer2017deep}
\bibfield{author}{\bibinfo{person}{Manzil Zaheer}, \bibinfo{person}{Satwik Kottur}, \bibinfo{person}{Siamak Ravanbakhsh}, \bibinfo{person}{Barnabas Poczos}, \bibinfo{person}{Russ~R Salakhutdinov}, {and} \bibinfo{person}{Alexander~J Smola}.} \bibinfo{year}{2017}\natexlab{}.
\newblock \showarticletitle{Deep sets}.
\newblock \bibinfo{journal}{\emph{Advances in neural information processing systems}}  \bibinfo{volume}{30} (\bibinfo{year}{2017}).
\newblock


\bibitem[Zhang et~al\mbox{.}(2020)]%
        {zhang2020set}
\bibfield{author}{\bibinfo{person}{David~W Zhang}, \bibinfo{person}{Gertjan~J Burghouts}, {and} \bibinfo{person}{Cees~GM Snoek}.} \bibinfo{year}{2020}\natexlab{}.
\newblock \showarticletitle{Set Prediction without Imposing Structure as Conditional Density Estimation}. In \bibinfo{booktitle}{\emph{International Conference on Learning Representations}}.
\newblock


\bibitem[Zhang et~al\mbox{.}(2022b)]%
        {zhang2022relational}
\bibfield{author}{\bibinfo{person}{Fengzhuo Zhang}, \bibinfo{person}{Boyi Liu}, \bibinfo{person}{Kaixin Wang}, \bibinfo{person}{Vincent Tan}, \bibinfo{person}{Zhuoran Yang}, {and} \bibinfo{person}{Zhaoran Wang}.} \bibinfo{year}{2022}\natexlab{b}.
\newblock \showarticletitle{Relational reasoning via set transformers: Provable efficiency and applications to MARL}.
\newblock \bibinfo{journal}{\emph{Advances in Neural Information Processing Systems}}  \bibinfo{volume}{35} (\bibinfo{year}{2022}), \bibinfo{pages}{35825--35838}.
\newblock


\bibitem[Zhang et~al\mbox{.}(2022a)]%
        {zhang2022composition}
\bibfield{author}{\bibinfo{person}{Jie Zhang}, \bibinfo{person}{Chen Cai}, \bibinfo{person}{George Kim}, \bibinfo{person}{Yusu Wang}, {and} \bibinfo{person}{Wei Chen}.} \bibinfo{year}{2022}\natexlab{a}.
\newblock \showarticletitle{Composition design of high-entropy alloys with deep sets learning}.
\newblock \bibinfo{journal}{\emph{npj Computational Materials}} \bibinfo{volume}{8}, \bibinfo{number}{1} (\bibinfo{year}{2022}), \bibinfo{pages}{89}.
\newblock


\bibitem[Zhang et~al\mbox{.}(2022c)]%
        {zhang2022set}
\bibfield{author}{\bibinfo{person}{Lily Zhang}, \bibinfo{person}{Veronica Tozzo}, \bibinfo{person}{John Higgins}, {and} \bibinfo{person}{Rajesh Ranganath}.} \bibinfo{year}{2022}\natexlab{c}.
\newblock \showarticletitle{Set norm and equivariant skip connections: Putting the deep in deep sets}. In \bibinfo{booktitle}{\emph{International Conference on Machine Learning}}. PMLR, \bibinfo{pages}{26559--26574}.
\newblock


\bibitem[Zhang et~al\mbox{.}(2019a)]%
        {zhang2019deep}
\bibfield{author}{\bibinfo{person}{Yan Zhang}, \bibinfo{person}{Jonathon Hare}, {and} \bibinfo{person}{Adam Prugel-Bennett}.} \bibinfo{year}{2019}\natexlab{a}.
\newblock \showarticletitle{Deep set prediction networks}.
\newblock \bibinfo{journal}{\emph{Advances in Neural Information Processing Systems}}  \bibinfo{volume}{32} (\bibinfo{year}{2019}).
\newblock


\bibitem[Zhang et~al\mbox{.}(2019b)]%
        {zhang2019fspool}
\bibfield{author}{\bibinfo{person}{Yan Zhang}, \bibinfo{person}{Jonathon Hare}, {and} \bibinfo{person}{Adam Pr{\"u}gel-Bennett}.} \bibinfo{year}{2019}\natexlab{b}.
\newblock \showarticletitle{FSPool: Learning Set Representations with Featurewise Sort Pooling}. In \bibinfo{booktitle}{\emph{International Conference on Learning Representations}}.
\newblock


\bibitem[Zhang et~al\mbox{.}(2019c)]%
        {zhang2018learning}
\bibfield{author}{\bibinfo{person}{Yan Zhang}, \bibinfo{person}{Jonathon Hare}, {and} \bibinfo{person}{Adam Prügel-Bennett}.} \bibinfo{year}{2019}\natexlab{c}.
\newblock \showarticletitle{Learning Representations of Sets through Optimized Permutations}. In \bibinfo{booktitle}{\emph{International Conference on Learning Representations}}.
\newblock


\bibitem[Zhang et~al\mbox{.}(2021b)]%
        {zhang2021multiset}
\bibfield{author}{\bibinfo{person}{Yan Zhang}, \bibinfo{person}{David~W Zhang}, \bibinfo{person}{Simon Lacoste-Julien}, \bibinfo{person}{Gertjan~J Burghouts}, {and} \bibinfo{person}{Cees~GM Snoek}.} \bibinfo{year}{2021}\natexlab{b}.
\newblock \showarticletitle{Multiset-Equivariant Set Prediction with Approximate Implicit Differentiation}. In \bibinfo{booktitle}{\emph{International Conference on Learning Representations}}.
\newblock


\bibitem[Zhang et~al\mbox{.}(2021a)]%
        {zhang2021sbo}
\bibfield{author}{\bibinfo{person}{Ziming Zhang}, \bibinfo{person}{Yun Yue}, \bibinfo{person}{Guojun Wu}, \bibinfo{person}{Yanhua Li}, {and} \bibinfo{person}{Haichong Zhang}.} \bibinfo{year}{2021}\natexlab{a}.
\newblock \showarticletitle{SBO-RNN: reformulating recurrent neural networks via stochastic bilevel optimization}.
\newblock \bibinfo{journal}{\emph{Advances in Neural Information Processing Systems}}  \bibinfo{volume}{34} (\bibinfo{year}{2021}), \bibinfo{pages}{25839--25851}.
\newblock


\bibitem[Zhao et~al\mbox{.}(2021)]%
        {zhao2021point}
\bibfield{author}{\bibinfo{person}{Hengshuang Zhao}, \bibinfo{person}{Li Jiang}, \bibinfo{person}{Jiaya Jia}, \bibinfo{person}{Philip~HS Torr}, {and} \bibinfo{person}{Vladlen Koltun}.} \bibinfo{year}{2021}\natexlab{}.
\newblock \showarticletitle{Point transformer}. In \bibinfo{booktitle}{\emph{Proceedings of the IEEE/CVF international conference on computer vision}}. \bibinfo{pages}{16259--16268}.
\newblock


\bibitem[Zhong et~al\mbox{.}(2018)]%
        {zhong2018compact}
\bibfield{author}{\bibinfo{person}{Yujie Zhong}, \bibinfo{person}{Relja Arandjelovic}, {and} \bibinfo{person}{Andrew Zisserman}.} \bibinfo{year}{2018}\natexlab{}.
\newblock \showarticletitle{Compact deep aggregation for set retrieval}. In \bibinfo{booktitle}{\emph{Proceedings of the European conference on computer vision (ECCV) workshops}}. \bibinfo{pages}{0--0}.
\newblock


\bibitem[Zhou and Tuzel(2018)]%
        {zhou2018voxelnet}
\bibfield{author}{\bibinfo{person}{Yin Zhou} {and} \bibinfo{person}{Oncel Tuzel}.} \bibinfo{year}{2018}\natexlab{}.
\newblock \showarticletitle{Voxelnet: End-to-end learning for point cloud based 3d object detection}. In \bibinfo{booktitle}{\emph{Proceedings of the IEEE conference on computer vision and pattern recognition}}. \bibinfo{pages}{4490--4499}.
\newblock


\bibitem[Zweig and Bruna(2022)]%
        {zweig2022exponential}
\bibfield{author}{\bibinfo{person}{Aaron Zweig} {and} \bibinfo{person}{Joan Bruna}.} \bibinfo{year}{2022}\natexlab{}.
\newblock \showarticletitle{Exponential separations in symmetric neural networks}.
\newblock \bibinfo{journal}{\emph{Advances in Neural Information Processing Systems}}  \bibinfo{volume}{35} (\bibinfo{year}{2022}), \bibinfo{pages}{33134--33145}.
\newblock


\end{thebibliography}

\end{document}